\documentclass[10pt]{article}

\usepackage[a4paper]{geometry}
\usepackage[english]{babel}
\usepackage{amsmath}
\usepackage{amssymb}
\usepackage{array}
\usepackage{authblk}
\usepackage{booktabs}
\usepackage{tabularx}
\usepackage{caption}
\usepackage{csquotes}
\usepackage{enumitem}
\usepackage{graphicx}
\usepackage{microtype}
\usepackage{multirow}
\usepackage{parskip}
\usepackage{placeins}
\usepackage{lineno}

\usepackage{rotating}
\usepackage{xcolor}
\usepackage[style=numeric,backend=biber,sorting=none,maxcitenames=2]{biblatex}
\usepackage[
  colorlinks=true,
  linkcolor=black,
  citecolor=black,
  urlcolor=black,
  pdfborder={0 0 0},
  bookmarksnumbered=true,
  pdftitle={Distributed Lag Neural Additive Models},
  pdfauthor={Calle Helmersson, Shivang Pandey, Leonardo Olivetti, Elena Raffetti},
  pdfsubject={Statistical Methodology for Distributed Lag Neural Additive Models},
  pdfkeywords={Distributed Lag Neural Additive Models; Distributed Lag Non-linear Models; Neural Additive Models; Deep Learning},
  pdfdisplaydoctitle=true
]{hyperref}

\renewcommand{\arraystretch}{1.2}
\title{Distributed Lag Neural Additive Models}

\author[1,2]{Calle Helmersson}
\author[2,3,*]{Shivang Pandey}
\author[3,4,5]{Leonardo Olivetti}
\author[2,3,4,6,7]{Elena Raffetti}

\affil[1]{Department of Mathematics, KTH Royal Institute of Technology, Stockholm, Sweden}
\affil[2]{Department of Global Public Health, Karolinska Institutet, Stockholm, Sweden}
\affil[3]{Swedish Centre for Impacts of Climate Extremes (climes), Uppsala University, Uppsala, Sweden}
\affil[4]{Department of Earth Sciences, Uppsala University, Uppsala, Sweden}
\affil[5]{Centre of Natural Hazards and Disaster Science (CNDS), Uppsala University, Uppsala, Sweden}
\affil[6]{British Heart Foundation Cardiovascular Epidemiology Unit, Department of Public Health and Primary Care, University of Cambridge, Cambridge, UK}
\affil[7]{Victor Phillip Dahdaleh Heart and Lung Research Institute, University of Cambridge, Cambridge, UK}
\affil[*]{Corresponding author: Shivang Pandey, \href{mailto:shivang.pandey@ki.se}{\texttt{shivang.pandey@ki.se}}}

\date{}

\begin{document}

\maketitle

\pdfbookmark[1]{Abstract}{abstract}
\begin{abstract}
  We introduce Distributed Lag Neural Additive Models (DLNAMs), neural-additive analogues of Distributed Lag Non-linear Models (DLNMs) for learning nonlinear effects distributed over lags. DLNAMs replace a prespecified spline cross-basis with neural components that learn exposure--lag response surfaces, avoiding choices of basis family, dimension, and knot placement while preserving additive interpretability and familiar distributed-lag summaries. Exp-centered input layers, smooth activations, and learned subnetwork mixtures produce smooth, locally adaptive representations; pointwise uncertainty combines a conditional last-layer Laplace approximation with between-member ensemble variation. In simulations, DLNAMs generally outperformed DLNM comparators, including penalized and treed variants, in recovering known response functions, with lower bias, stronger boundary recovery, and better-calibrated cumulative intervals; gains were largest for more demanding functions. The architecture performed consistently across sample sizes, outcome families, lag horizons, and jointly fitted multi-exposure settings, retaining recovery performance as exposures were added; fit-specific changes were largely confined to optimization, and applications recovered established empirical patterns.

\end{abstract}
\section{Introduction}
\label{sec:introduction}

When an outcome depends on a history of past exposure rather than on its current value alone, the scientific object is a response function over both exposure level and elapsed time. Such distributed-lag relationships arise whenever a system responds gradually, transiently, or with delay. Formalized early in econometrics \cite{almon1965}, they are widely studied in environmental epidemiology, where the timing of temperature, pollution, and other exposures can matter alongside their magnitude \cite{gasparrini2015lancet,bhaskaran2013timeseries}. Distributed Lag Non-linear Models (DLNMs) provide a widely used framework for modeling nonlinear effects distributed over past exposures within an additive, interpretable structure \cite{gasparrini2010dlnm,gasparrini2014elr,gasparrini2011dlnmR}.

The DLNM framework represents exposure--lag relationships through smooth additive functions, most commonly implemented in environmental epidemiology using spline cross-bases \cite{gasparrini2010dlnm,gasparrini2014elr,gasparrini2011dlnmR}. These combine marginal basis functions over exposure and lag to construct a smooth response surface. Their flexibility is therefore shaped by choices such as basis family, degrees of freedom, and knot placement. These choices may be guided by substantive expertise or selected from candidate specifications using quasi-likelihood information criteria (QAIC or QBIC). P-DLNM reduces sensitivity to smoothness specification through penalization, but local variation is still expressed within a prescribed spline representation \cite{gasparrini2017penalised}. T-DLNM instead learns local partitions of the exposure--lag domain, gaining substantially greater local adaptivity but at the cost of a piecewise-constant response \cite{mork2022tdlnm}. Each approach addresses a different facet of the representation problem, yet smoothness, local adaptivity, and analysis-specific representation remain in tension. Neither combines a smooth, locally adaptive response with a representation whose internal features are learned during estimation.

We introduce Distributed Lag Neural Additive Models (DLNAMs), in which each lagged predictor contributes a learned neural component within an additive predictor. DLNAMs replace prespecified spline representations with learned smooth functions and retain the principal scientific objects through which distributed-lag models are interpreted: exposure--lag surfaces, cumulative response functions, and covariate effects. Additional lagged predictors enter as further learned components, with the core architecture generally reusable as the model expands rather than requiring the response representation to be redesigned for each added predictor. Prediction-oriented work has used distributed-lag exposure histories as inputs to generic artificial neural networks \cite{guo2021ann}; DLNAMs instead estimate a separate exposure--lag response surface for each exposure, from which the usual distributed-lag summaries are obtained. The organizing principle is that additivity and functional representation are separable: the additive predictor makes each component separately reportable, irrespective of the function class used to represent it.

This separation has an additive-model precedent. Generalized Linear Models (GLMs) use linear components \cite{mccullagh1989glm}, Generalized Additive Models (GAMs) replace them with smooth basis-expanded functions \cite{hastie1990gam}, and Neural Additive Models (NAMs) learn the component functions with neural networks \cite{agarwal2021neural}. In the cross-sectional setting, NAMs achieve accuracy comparable to state-of-the-art GAMs while retaining additive interpretability and the architectural flexibility of neural networks \cite{agarwal2021neural}. The distributed-lag lineage runs in parallel, from Distributed Lag Models (DLMs) to DLNMs \cite{almon1965,gasparrini2010dlnm}; DLNAMs complete it with the neural-component analogue.
\[
  \begin{array}{l@{\qquad}ccccc}
    \text{Cross-sectional:} & \mathrm{GLM} & \longrightarrow & \mathrm{GAM}  & \longrightarrow & \mathrm{NAM},\\
    \text{Distributed lag:} & \mathrm{DLM} & \longrightarrow & \mathrm{DLNM} & \longrightarrow & \mathrm{DLNAM}.
  \end{array}
\]
Each step replaces the functional representation of the components while leaving the additive structure intact.

Constructing that analogue is not straightforward. A generic neural substitution is insufficient: a useful component must reconcile local adaptivity with smoothness, stable estimation, additive reporting, and uncertainty quantification. The DLNAM architecture combines exp-centered unit (ExU) input layers \cite{agarwal2021neural} with smooth activations \cite{misra2020mish,elfwing2018silu} and a learned mixture of subnetworks. Because exposure and lag enter a surface component jointly, the scalar ExU construction of NAMs is extended to multiple input coordinates; alternative multivariate constructions are compared directly. Complete fitted models are ensembled across independent initializations \cite{lakshminarayanan2017ensembles}. Pointwise uncertainty combines conditional last-layer Laplace approximations with between-member ensemble variation \cite{daxberger2021laplace,bouchiat2024bayesian}. Together, these design choices make DLNAMs structured neural estimators rather than generic feed-forward substitutes for splines.

We evaluate DLNAMs by how well they recover the exposure--lag response functions of scientific interest. Simulation studies assess recovery of cumulative responses and full exposure--lag surfaces, interval calibration, architectural ablations, and additive component recovery under correlated concurrent exposures. Applications to Chicago temperature--mortality data and a large multi-country malaria analysis test whether the same component architecture can be reused across markedly different sample sizes, outcome families, and lag structures. We additionally quantify recurring representation-selection burden and fixed-budget computational scaling as concurrent lagged components are added. Taken together, these experiments test whether representation learning can augment distributed-lag modeling without surrendering the scientific structure that makes its effects interpretable.

\section{Results}
\label{sec:results}

The principal simulation comprised one separable data-generating process (DGP) and three structurally demanding, nonseparable DGPs with localized or heterogeneous exposure--lag structure (Fig.~\ref{fig:supp_dgp}). We compare DLNAM with four DLNM-family estimators: DLNM (QAIC), DLNM (QBIC), P-DLNM, and T-DLNM (Section~\ref{sec:estimators_fitting}). Across Monte Carlo replicates, we evaluated recovery of the cumulative response and full exposure--lag surface, interval calibration, and performance in sparsely supported exposure tails (Section~\ref{sec:metrics}). Here, recovery denotes agreement between an estimated target and its known data-generating counterpart.

\subsection{Recovery}

\begin{figure}[!tbp]
\centering
\includegraphics[width=\textwidth]{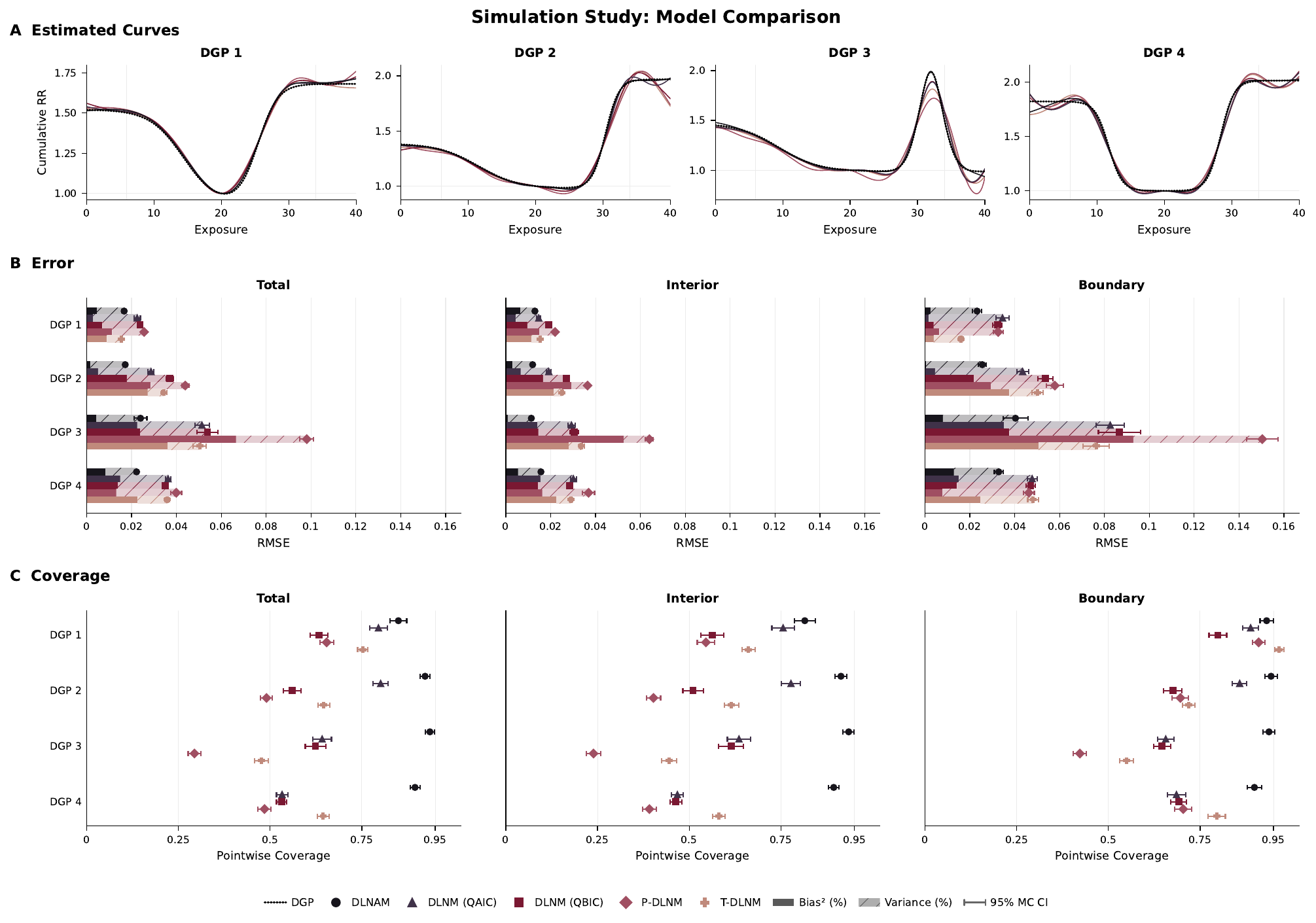}
\caption{\textbf{Simulation Study: Model Comparison.} Monte Carlo simulation of four DGPs, with \(R=200\) replicates per DGP. (\textbf{A}) Estimated cumulative relative-risk functions and data-generating functions, shown as pointwise Monte Carlo means across fitted replicates. (\textbf{B}) Cumulative log-relative-risk RMSE by method and exposure region, partitioned into squared bias and variance. (\textbf{C}) Empirical coverage of pointwise \(95\%\) intervals. Error bars denote 95\% Monte Carlo confidence intervals. Boundary denotes exposure values below the 5th or above the 95th percentile of the reference exposure distribution; interior denotes its complement. Associated numerical results are reported in Tables~\ref{tab:supp_mc} and~\ref{tab:supp_mc_bv}.}
\label{fig:mc_cum}
\end{figure}

Across DGPs 2--4, DLNAM attains the lowest error of the five estimators over the whole exposure grid, the interior, and the boundary region (Fig.~\ref{fig:mc_cum}B; Table~\ref{tab:supp_mc}). On the separable DGP 1, T-DLNM attains the lowest whole-grid and boundary RMSE, while DLNAM is lowest in the interior; T-DLNM's advantage is driven by lower variance rather than lower bias. DLNAM also has the smallest across-DGP range in whole-grid RMSE. Its gains concentrate on DGPs 2--4, consistent with representation learning being most useful for demanding local or heterogeneous structure, although the design does not isolate nonseparability, localization, or any single dimension of complexity as the cause.

The broad pattern persists when recovery is scored on the full exposure--lag surface---a target emphasized in both comparator studies \cite{gasparrini2017penalised,mork2022tdlnm}---with the treed fit remaining competitive at the boundary (Fig.~\ref{fig:supp_mc_surf}; Table~\ref{tab:supp_mc_surf}). The cumulative gains therefore do not arise from lag-specific errors cancelling in the sum.

\subsection{Bias and Variance}

The bias--variance decomposition exposes distinct failure modes (Fig.~\ref{fig:mc_cum}B; Table~\ref{tab:supp_mc_bv}). DLNAM has the lowest whole-grid squared bias in DGPs 2--4, while T-DLNM has the lowest whole-grid variance in DGPs 1--2. On DGPs 2--4, rising comparator error is driven chiefly by squared bias; DLNAM's residual cumulative error is predominantly variance on every DGP. The same contrast appears in the sparsely supported exposure tails: fitted cross-bases can oscillate there, whereas the learned components tend to flatten. Oscillation can displace the fitted response; flattening instead tends to attenuate a true tail effect.

\subsection{Calibration}

No estimator attains nominal coverage of pointwise \(95\%\) intervals throughout the design (Fig.~\ref{fig:mc_cum}C). On the cumulative curve, the DLNAM is closest to nominal coverage in 11 of the 12 DGP-by-region cells; the exception is the DGP 1 boundary, where T-DLNM is closer to nominal (Table~\ref{tab:supp_mc}).

Coverage and interval width must be read jointly because coverage can be raised mechanically by widening intervals (Table~\ref{tab:supp_mc_rel}). On the full exposure--lag surface, every estimator with higher coverage than DLNAM also has wider intervals; on the cumulative curve the same holds apart from two DGP-by-region cells. T-DLNM makes the trade-off most visible on the surface target, where it combines the highest coverage with the widest intervals and larger error. DLNAM's calibration advantage therefore does not come from uniformly wider intervals. Adding between-member ensemble variation to the last-layer Laplace variance raises DLNAM coverage in every DGP for both cumulative and full-surface targets, with the Laplace-only result shown in Table~\ref{tab:supp_cov_decomp}. Fit-to-fit instability thus carries material information for interval calibration.

\FloatBarrier
\subsection{Architecture Ablation}

\begin{figure}[!tbp]
\centering
\includegraphics[width=\textwidth]{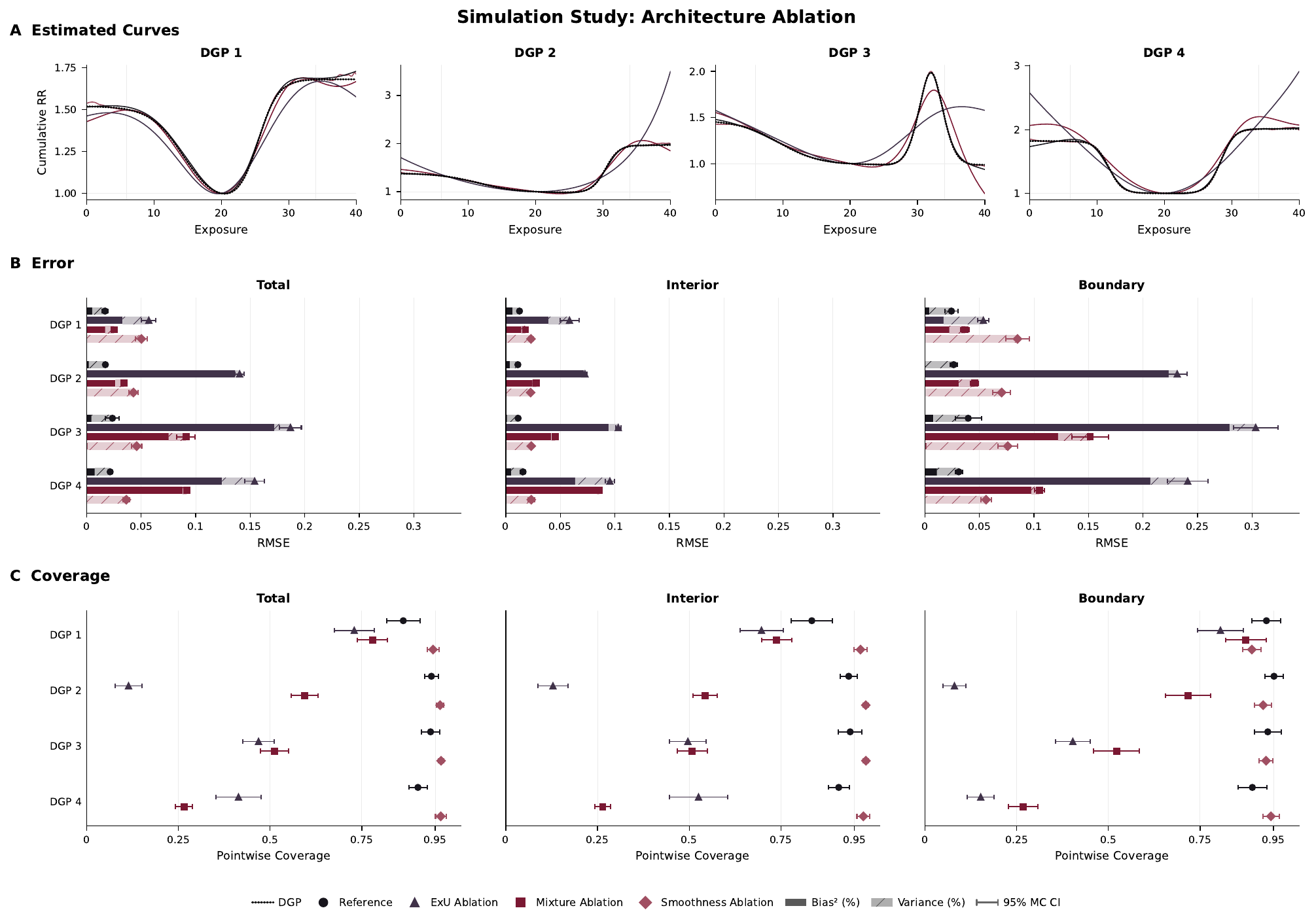}
\caption{\textbf{Simulation Study: Architecture Ablation.} Monte Carlo simulation of four DGPs, with \(R=50\) replicates per DGP. (\textbf{A}) Estimated cumulative relative-risk functions and data-generating functions, shown as pointwise Monte Carlo means across fitted replicates. (\textbf{B}) Cumulative log-relative-risk RMSE by method and exposure region, partitioned into squared bias and variance. (\textbf{C}) Empirical coverage of pointwise \(95\%\) intervals. Error bars denote 95\% Monte Carlo confidence intervals. Boundary denotes exposure values below the 5th or above the 95th percentile of the reference exposure distribution; interior denotes its complement. Ablations are applied one at a time and are not factorial. Associated numerical results are reported in Table~\ref{tab:supp_abl}.}
\label{fig:abl}
\end{figure}

The ablation isolates which architectural elements matter for recovery (Fig.~\ref{fig:abl}; Table~\ref{tab:supp_abl}). The reference specification denotes the complete DLNAM architecture used in the principal analysis. Replacing the ExU input layer with a linear input layer produces the largest increase in error across all four DGPs and sharply reduces coverage, with the largest degradation on DGPs 2--4. The learned input scales therefore play a role analogous to marginal resolution in a cross-basis, except that they are fitted parameters rather than a dimension declared in advance.

The remaining ablations expose different failure modes. Replacing the mixture with the standard single-network formulation increases error and depresses coverage across all four DGPs; replacing the smooth activations increases error but leaves coverage at or above the full model's. The input-layer and mixture ablations are most costly on DGPs 2--4, whereas the smooth-activation ablation costs most when the generating surface itself varies smoothly. These patterns indicate that recovery is not explained by generic neural-network capacity alone. ExU supplies local sensitivity, the mixture composes learned functions, and the smooth hidden activations govern how that flexibility is expressed. Because the ablations are one-at-a-time, they do not support factorial attribution of independent effects.

\FloatBarrier
\subsection{Joint-Exposure Recovery}

\begin{figure}[!tbp]
\centering
\includegraphics[width=\textwidth]{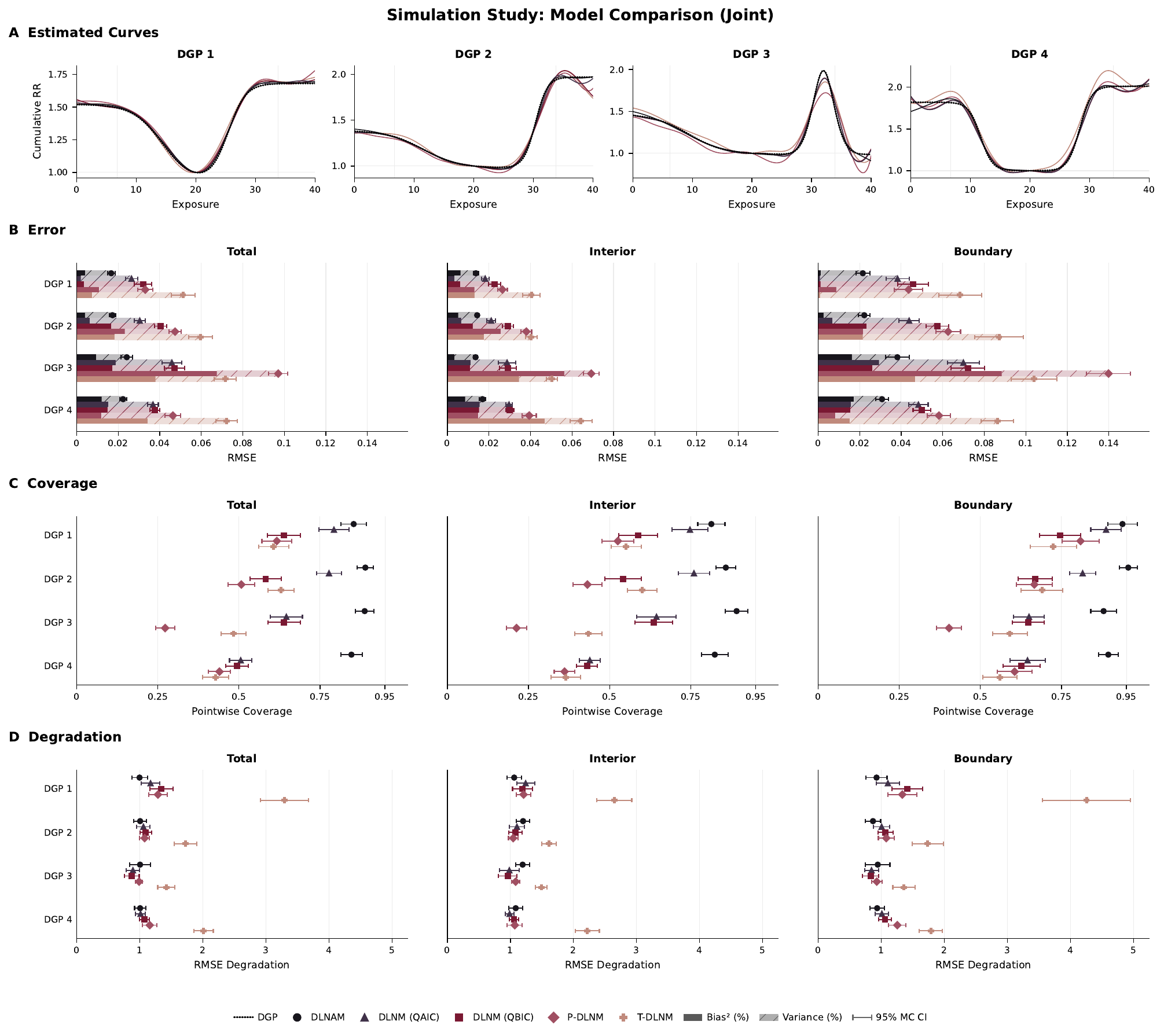}
\caption{\textbf{Simulation Study: Model Comparison (Joint).} Monte Carlo simulation jointly fitting the four DGPs as concurrent exposure--lag components, with \(R=50\) replicates, alongside one correlated null exposure. (\textbf{A}) Estimated cumulative relative-risk functions and data-generating functions for the four active components, shown as pointwise Monte Carlo means across fitted replicates. (\textbf{B}) Cumulative log-relative-risk RMSE by method and exposure region, partitioned into squared bias and variance. (\textbf{C}) Empirical coverage of pointwise \(95\%\) intervals. (\textbf{D}) RMSE degradation relative to the corresponding single-exposure fits. Error bars denote 95\% Monte Carlo confidence intervals. Boundary denotes exposure values below the 5th or above the 95th percentile of the reference exposure distribution; interior denotes its complement. Associated numerical results, including the correlated null exposure, are reported in Table~\ref{tab:supp_joint}.}
\label{fig:mc_joint}
\end{figure}

With four active exposure--lag components fitted concurrently alongside a fifth correlated null exposure, DLNAM achieved the lowest cumulative error of the five estimators for every active component over the whole exposure grid, the interior, and the boundary region (Fig.~\ref{fig:mc_joint}A,~B). Whole-grid degradation ratios remained near 1 across all four active components relative to the corresponding single-exposure experiments (Fig.~\ref{fig:mc_joint}D). Region-specific ratios were somewhat higher in the interior and lower at the boundary, indicating redistribution of error across the exposure domain rather than a uniform joint-fitting penalty. Because T-DLNM is target-specific in this experiment, its larger degradation pertains to the practical multi-exposure formulation used here, not to a hypothetical fully joint treed model.

Coverage of the active components remained below nominal, but the DLNAM was closest to nominal in all 12 active component-by-region cells (Fig.~\ref{fig:mc_joint}C; Table~\ref{tab:supp_joint}). For the correlated null exposure, DLNAM and DLNM (QBIC) showed the least leakage by whole-grid RMSE (0.0124 and 0.0123, respectively), while DLNAM had coverage closest to nominal (0.955).

Together, these results support stable recovery of additive cumulative components under the specified correlation structure without establishing robustness to arbitrary collinearity or exposure--exposure interactions.

\FloatBarrier
\subsection{Chicago NMMAPS Temperature and Mortality}

\begin{figure}[!ht]
\centering
\includegraphics[width=\textwidth]{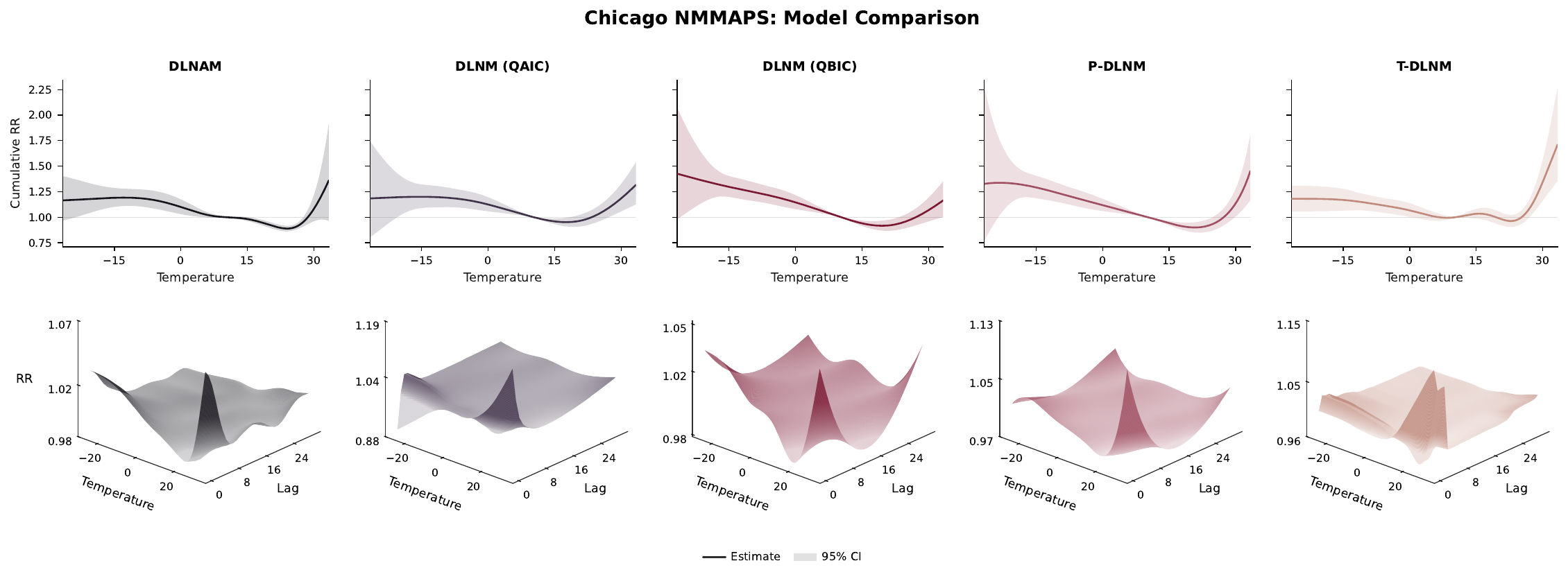}
\caption{\textbf{Chicago NMMAPS: Model Comparison.} Cumulative relative-risk functions and corresponding temperature--lag response surfaces are shown for the fitted methods. Shaded bands around the cumulative functions denote pointwise \(95\%\) intervals. All methods use the same adjustment variables. Analysis specifications are reported in Table~\ref{tab:supp_comparators}.}
\label{fig:chicago}
\end{figure}

Using 4638 complete days from the Chicago component of the National Morbidity, Mortality, and Air Pollution Study (NMMAPS) and a 30-day lag window, the DLNAM recovers the familiar temperature--mortality pattern \cite{gasparrini2011dlnmR} (Fig.~\ref{fig:chicago}): a minimum-mortality temperature with risk rising toward both extremes, heat acting within days, and cold accumulating over one to two weeks. All five estimators agree on the broad cumulative shape but place the minimum across a range of several degrees, showing the fitted optimum's sensitivity to specification. They nonetheless differ in how the heat association is distributed across lags, with the DLNAM surface spreading a lower elevation across more of the early lags.

The Chicago NMMAPS data are the worked example distributed with the \texttt{dlnm} package \cite{gasparrini2011dlnmR}, and the published analysis supplies an empirical reference under a different representation and adjustment set. DLNAM produced a similar overall shape, minimum location, and heat-response magnitude without explicit cross-basis selection. In the sparse cold tail, its curve remains within the envelope of the likelihood-matched fits while its intervals are the narrowest among them. This observational, in-sample application tests whether the architecture can recover a familiar empirical pattern; it is not intended to establish causal, predictive, or benchmark superiority.

\FloatBarrier
\subsection{Multi-Country DHS/MIS Malaria}

\begin{figure}[!tbp]
\centering
\includegraphics[width=\textwidth]{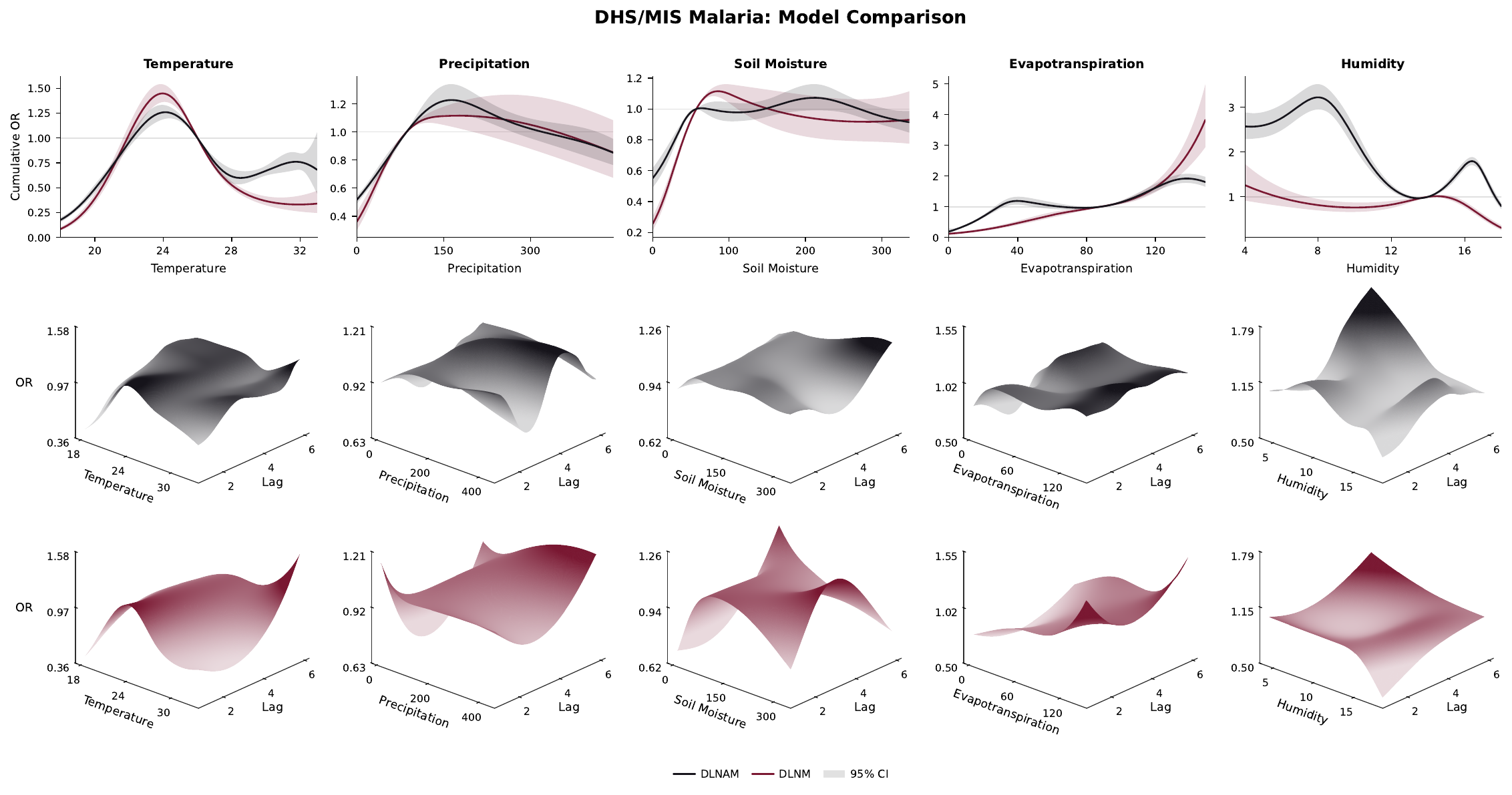}
\caption{\textbf{DHS/MIS Malaria: Model Comparison.} Cumulative odds-ratio functions and corresponding DLNAM and DLNM exposure--lag response surfaces are shown for five environmental exposures. Shaded bands around the cumulative functions denote pointwise \(95\%\) intervals. Each column reports the target-specific fit for that exposure; the analyses differ in adjustment representation and hierarchy. Analysis specifications are reported in Table~\ref{tab:supp_comparators}.}
\label{fig:malaria}
\end{figure}

DLNAM was applied to 348{,}565 childhood malaria outcomes from Demographic and Health Surveys (DHS) and Malaria Indicator Surveys (MIS) across 26 sub-Saharan African countries, under a logit link over 6 monthly lags (Fig.~\ref{fig:malaria}). One model was fitted per reported exposure using the adjustment set specified for that target in the source analysis \cite{martellini2026malaria}. Within each fit, the adjusting exposures retain full exposure--lag surfaces; the specific-humidity model therefore contains 5 concurrently lagged surfaces.

A DLNM is fitted under the same adjustment sets, with adjusting exposures reduced to scalar lag means. The two analyses broadly agree on the response shapes and temperature optimum, with the largest divergence for specific humidity. We do not interpret that divergence as evidence in favor of either model because the analyses also differ in adjustment representation and treatment of the survey hierarchy: weight-decayed level-specific effects in the DLNAM and variance-component random intercepts in the DLNM fit. The application demonstrates that several fully lagged exposures can be represented together at this scale. At the lag-specific level, the fitted DLNAM temperature surface also concentrates the elevated association around the shortest lags, consistent with the source analysis.

\FloatBarrier
\subsection{Computational Scaling}

\begin{figure}[!ht]
  \centering
  \includegraphics[width=0.7\linewidth]{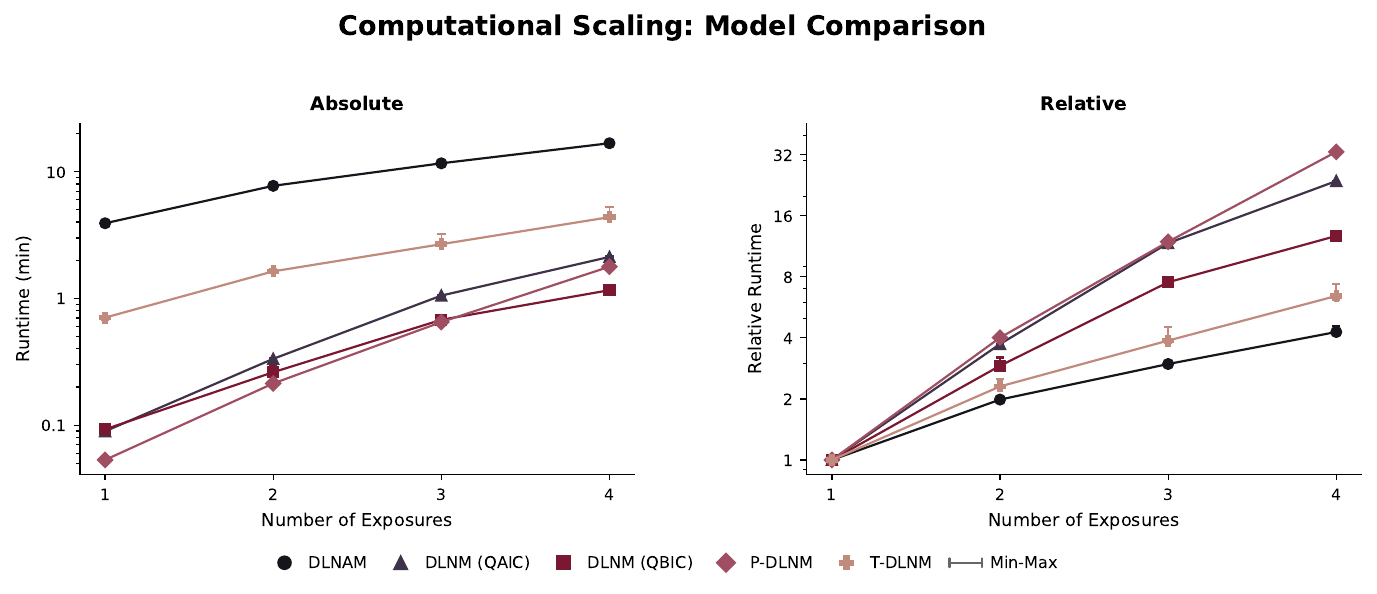}
  \caption{\textbf{Computational Scaling: Model Comparison.} Absolute end-to-end fit time and runtime relative to each method's one-exposure fit are shown across one to four exposures, with \(R=3\) repeats per method and exposure count. Markers denote the median across repeats; bars denote the observed minimum--maximum range. Benchmark specifications are reported in Table~\ref{tab:supp_comparators}.}
  \label{fig:runtime}
\end{figure}

A fixed-budget benchmark measured absolute fitting time and scaling as concurrently lagged exposures were added (Fig.~\ref{fig:runtime}; Table~\ref{tab:supp_comparators}). On the reported workstation, DLNAM was slower in absolute wall-clock time than the comparators for one exposure. At a fixed 2500-epoch DLNAM budget, runtime increased 4.3-fold from one to four exposures, compared with 6.2-fold for T-DLNM, 12.5-fold for DLNM (QBIC), 23.7-fold for DLNM (QAIC), and 33.5-fold for P-DLNM. The production joint DLNAM used 5000 epochs; under approximately linear scaling with training budget, this would imply an approximately 8.6-fold increase over the one-exposure benchmark. This extrapolation exceeds the observed T-DLNM increase, although its multi-exposure implementation uses target-specific treed fits and represents remaining exposures by spline adjustment terms rather than fitting a single joint treed model. The benchmark therefore compares implemented procedures rather than establishing an unconditional efficiency ranking; absolute timings depend on hardware and implementation.

\FloatBarrier
\section{Discussion}
\label{sec:discussion}

The central methodological result is the model class itself: DLNAMs learn smooth exposure--lag response surfaces as neural-additive components while preserving the additive structure and familiar scientific summaries of DLNMs. On the more structurally demanding surfaces, comparator error rises chiefly through squared bias, whereas DLNAM's remaining cumulative error is predominantly variance. Representation bias systematically displaces the reported curve when the fitted function class cannot express the target; sampling variability, by contrast, can recede with additional information. The same contrast is especially visible in sparsely supported exposure tails. On the more structurally demanding DGPs, DLNAM retained lower boundary error and its learned components tended to flatten rather than display the oscillatory edge behavior seen in some spline fits. This suggests a practical advantage where support is weakest, without implying that spline estimators are generically unstable at boundaries. Some remaining variability reflects random initialization and can be reduced by enlarging the ensemble, at additional computational cost.

Recovery alone is insufficient: nominal interval formulas do not guarantee nominal coverage, so we treat calibration as an empirical property, as do the comparator method studies \cite{gasparrini2017penalised,mork2022tdlnm}. In the principal simulation, DLNAM is closest to nominal cumulative coverage in 11 of 12 DGP-by-region cells; in the joint experiment it is closest in all 12 active component-by-region cells. On the full-surface target, every estimator with higher coverage than DLNAM also reports wider intervals, with only two exceptions to the analogous pattern on the cumulative target. The advantage is therefore one of calibration relative to interval width, not coverage in isolation. The interval construction is deliberately pragmatic: combining a conditional last-layer Laplace approximation with between-member ensemble variation avoids repeated bootstrap refitting and the posterior-sampling burden of Markov chain Monte Carlo. Conditional on the learned representation, final-layer inference reduces to a low-dimensional generalized linear model calculation, giving a neural-additive analogue of pointwise DLNM inference \cite{gasparrini2011dlnmR}. These are not simultaneous bands or prediction intervals and do not propagate all uncertainty in the learned representation; the ensemble term captures the part expressed as disagreement across independently initialized fits.

A neural estimator exposes a large nominal hyperparameter space, but nominal choice is not recurring choice. Operationally, the relevant question is which decisions must be reopened from one analysis to the next. DLNMs require the analyst to prescribe or select a cross-basis representation, whether through substantive judgment or a selection procedure. P-DLNM learns effective smoothness within a penalized representation, but the marginal smoother and nominal basis dimension remain modeling choices. T-DLNM instead learns local partition structure under a specified tree-ensemble, prior, and sampler configuration \cite{gasparrini2017penalised,mork2022tdlnm}. A spline cross-basis fixes the coordinates of the function space before estimation, after which fitting estimates their coefficients; DLNAM learns its internal features jointly with the response. Across the analyses here, the same architecture remained effective over two outcome families, three lag horizons, and one to five simultaneous exposures; fit-specific changes were concentrated primarily in optimization---training budget, learning rate, or batching---when the fitting problem changed materially (Section~\ref{sec:estimators_fitting}). This does not establish universal tuning insensitivity; it does show that a large nominal neural design space need not translate into a comparably large recurring specification burden.

The ablation shows that learning the representation is nontrivial: recovery depends on how the neural component is structured, not merely on replacing a cross-basis with a larger feed-forward network. Exposure and lag enter jointly through a multivariate ExU input layer, several jointly trained subnetworks compose the component, and smooth hidden activations regulate how local sensitivity propagates through it. Removing any of these elements degrades recovery with a distinct error or coverage signature. The contribution is therefore architectural as well as representational: local sensitivity, compositional flexibility, smoothness, and stable fitting must coexist inside an additive component. Independent complete fits are then ensembled across initializations to temper fit-to-fit instability. The ExU comparison and subnetwork decomposition are reported in Supplementary Texts~\ref{supp:exu} and~\ref{supp:mixture}.

The additive form is especially important when several lagged inputs are modeled together. A cross-basis analysis can also accommodate multiple lagged predictors, but every additional surface introduces another representation to specify. In DLNAMs, additional lagged predictors enter as further learned components. The joint simulation tests component recovery under correlated exposures; the malaria analysis shows that the construction remains feasible at application scale. The substantive scientific decisions---which exposures to include, what to adjust for, and which contrasts to report---remain. What need not be redesigned for every added exposure is the learned response representation. The malaria analysis still fits one model per reported target exposure, but within each fit every adjusting exposure retains its full distributed-lag surface rather than being reduced to the scalar lag-mean adjustment used in the comparison.

These advantages do not imply that DLNAMs are preferable in every setting. When modeling a single exposure with a comparatively simple smooth surface adequately represented by a low-dimensional cross-basis, a DLNM is simpler, competitive, and cheaper to fit in this benchmark. Abrupt critical-window structure can likewise favor a treed representation over a smooth estimator \cite{mork2022tdlnm}. Learning the representation entails greater computational demand in the reported setup. Under a fixed training budget, however, DLNAM runtime increased less steeply as concurrent exposures were added than the comparator implementations. The larger training budget used for the production joint fit cautions against treating that benchmark as an unconditional efficiency ranking. Comparison with T-DLNM is further qualified by its target-specific rather than fully joint multi-exposure implementation. Absolute runtime also depends on hardware and software implementation, so the reported timings should not be interpreted as a fixed computational property of the model class.

The evidence is necessarily finite in scope: the principal comparison uses \(R=200\) replicates and the architecture and joint-exposure studies \(R=50\). Monte Carlo uncertainty remains relevant for closely spaced quantities, although in the joint study Monte Carlo standard errors are small relative to the RMSE separations among active components, so the principal recovery ranking remains well resolved. The joint experiment evaluates a single correlation structure and therefore does not establish how recovery changes across a broader range of between-exposure dependence. The simulations span several distinct surface geometries but do not define a universal threshold at which a learned neural representation becomes preferable to a simpler smoother. The applications target recovery of response structure rather than prediction; out-of-sample evaluation is secondary to that aim and can be nontrivial with dependent lag histories.

The model remains additive across exposures. Selected pairwise terms extend readily to additive models, including NAMs \cite{lou2013accurate,agarwal2021neural}, but admit non-equivalent distributed-lag formulations; \textcite{chen2019interaction} provide one structured alternative, whose integration with DLNAMs remains future work. Two further compositional extensions follow directly. Outcome-specific mixing weights over shared subnetworks define a Multitask DLNAM in the manner of multitask NAMs~\cite{agarwal2021neural}. For one outcome it reduces to the Mixture DLNAM, of which the standard DLNAM is a special case. Shared subnetworks induce cross-outcome regularization, potentially improving recovery for related outcomes, but confine outcome-specific flexibility to the mixing weights. A Hierarchical DLNAM adds partially pooled group-specific deviations to global components, reducing to the Mixture DLNAM when these vanish and paralleling hierarchical DLNMs~\cite{economou2024hierarchical}. Hierarchical DLNMs provide a single-stage alternative to the two-stage pooling common in multicountry analyses~\cite{gasparrini2015lancet}; the malaria analysis motivates the corresponding DLNAM because country and survey cluster currently enter only through level-specific intercepts. Neither extension was implemented or evaluated here; formal definitions are given in Supplementary Text~\ref{supp:extensions}.

Beyond these extensions, the reported specification was not exhaustively optimized; the ablations show that changes to individual modules can materially alter recovery, while its modular construction allows modules to evolve without redesigning the model. Component-specific shrinkage \cite{bouchiat2024bayesian} and fuller propagation of uncertainty in the learned representation \cite{daxberger2021laplace} are particularly immediate directions suggested by the present results. The reported architecture is therefore an effective first realization rather than a performance ceiling. At the application level, fitted DLNAMs could provide the temperature--mortality response component in operational frameworks that translate weather forecasts into real-time health forecasts \cite{mistry2024realtime}, using trajectories from AI weather models such as Pangu-Weather and FourCastNet \cite{bi2023pangu,pathak2022fourcastnet}, reflecting the growing role of deep learning in weather forecasting \cite{eisenstein2026technologies}.

The present results therefore identify a specific role for DLNAMs: settings in which the response structure is sufficiently demanding that a prescribed representation may become restrictive, several lagged predictors must be modeled together, or repeated cross-basis specification is itself a substantial part of the analysis. Simpler established estimators remain attractive when a modest smooth representation is adequate, particularly where computation is limiting. More broadly, DLNAMs provide a route for translating advances in deep-learning methodology into distributed-lag models while preserving additive interpretability and familiar distributed-lag summaries.

\section{Methods}
\label{sec:methods}

\subsection{Distributed Lag Neural Additive Model}
\label{sec:dlnam_formulation}

Let \(\{Y_t\}_{t=1}^{n}\), \(\{x_t\}_{t=1}^{n}\), and \(\{u_t\}_{t=1}^{n}\) denote outcomes, exposures, and additional covariates, respectively, with \(Y_t\in\mathcal Y\), \(x_t\in\mathbb{R}^{M}\) and \(u_t\in\mathbb{R}^{K}\). Writing \(\mathcal L=(\mathcal L_1,\ldots,\mathcal L_M)\) for the collection of finite integer-lag sets, define
\[
x_t^{\mathcal L}
=
\bigl((x_{t-\ell,m})_{\ell\in\mathcal L_m}\bigr)_{m=1}^{M},
\qquad
\mathcal H_t=(x_t^{\mathcal L},u_t),
\]
and let \(\mu_t=\mathbb{E}(Y_t\mid\mathcal H_t)\) denote the conditional mean outcome given the exposure history and additional covariates. For an invertible link function \(g\), the DLNAM predictor is
\begin{equation}
g(\mu_t)
=
\alpha
+
\sum_{m=1}^{M}\sum_{\ell\in\mathcal L_m}
f_m^{\theta_m^x}(x_{t-\ell,m},\ell)
+
\sum_{k=1}^{K}h_k^{\theta_k^u}(u_{tk}).
\label{eq:dlnam_predictor_general}
\end{equation}

Recurring notation is summarized in Supplementary Table~\ref{tab:supp_notation}. The exposure--lag response surfaces \(f_m^{\theta_m^x}\) and covariate shape functions \(h_k^{\theta_k^u}\) are, for continuous inputs, represented by feed-forward neural networks, with \(\theta=(\alpha,\theta^x,\theta^u)\) denoting the complete parameter vector.

In DLNMs, the exposure--lag function is instead defined through a prescribed cross-basis combining marginal basis expansions over exposure and lag, with the resulting contribution evaluated and summed over the lagged exposure history \cite{gasparrini2010dlnm,gasparrini2014elr,gasparrini2011dlnmR}. Written in terms of the response surface implied by that cross-basis, DLNMs share the predictor form of Equation~\eqref{eq:dlnam_predictor_general}, with the defining distinction lying in the representation of \(f_m^{\theta_m^x}\): DLNAMs learn this function directly as a neural component, while terms corresponding to \(h_k^{\theta_k^u}\) may enter parametrically or through separate smooths.

For identifiability and interpretation, \(x^\star\in\mathbb{R}^{M}\) and \(u^\star\in\mathbb{R}^{K}\) define the reference point, and fitted components are expressed as contrasts satisfying
\[
f_m^{\theta_m^x}(x_m^\star,\ell)=0,
\quad \ell\in\mathcal L_m,
\qquad
h_k^{\theta_k^u}(u_k^\star)=0,
\]
with the resulting constants absorbed into \(\alpha\), so that \(\alpha\) is the linear predictor at the reference values and each component is interpreted as a contrast on the link scale.

Motivated by the compositional subnetwork architecture of multitask NAMs \cite{agarwal2021neural}, our primary specification, the Mixture DLNAM, generalizes the standard DLNAM by replacing the direct parameterization of each component with a learned linear combination of jointly trained subnetworks,
\begin{equation}
f_m^{\theta_m^x}(\cdot,\cdot)
=
\sum_{s=1}^{S_m^x}
\omega_{ms}^x
\tilde f_{ms}^{\vartheta_{ms}^x}(\cdot,\cdot),
\qquad
h_k^{\theta_k^u}(\cdot)
=
\sum_{s=1}^{S_k^u}
\omega_{ks}^u
\tilde h_{ks}^{\vartheta_{ks}^u}(\cdot).
\label{eq:dlnam_mixture}
\end{equation}

Under this parameterization, \(\vartheta=(\vartheta^x,\vartheta^u)\) collects the subnetwork parameters and \(\omega=(\omega^x,\omega^u)\) their mixing weights, so that \(\theta=(\alpha,\vartheta,\omega)\). The mixing weights are unconstrained rather than restricted to form a convex combination, allowing subnetworks to reinforce or partially offset one another.

The standard DLNAM is nested within Equation~\eqref{eq:dlnam_mixture} as the special case in which each component contains a single subnetwork and the mixing weights form a convex combination. Allowing the sole mixing weight to remain unconstrained leaves the resulting function class unchanged, since that scalar can be absorbed into the output layer, but changes the optimization parameterization: its independent initialization rescales the component and can temper its early contribution and gradients. The distinction between the standard DLNAM and a one-subnetwork Mixture DLNAM is therefore computational rather than functional.

To reduce variability attributable to random initialization, we ensemble \(B\) independently initialized DLNAMs. Mixture composition and ensembling operate at different levels: subnetworks are optimized jointly within each component of a fitted model, whereas the ensemble averages complete fitted models. Supplementary Text~\ref{supp:mixture} illustrates how weighted subnetworks combine through reinforcement and partial cancellation within one ensemble member. The ensemble is combined on the link scale,
\begin{equation}
  \eta_t
  =
  \frac{1}{B}\sum_{b=1}^{B}
  \eta_t\!\left(\theta^{(b)}\right),
  \qquad
  \eta_t\!\left(\theta^{(b)}\right)
  =
  g\!\left(\mu_t\!\left(\theta^{(b)}\right)\right).
  \label{eq:dlnam_ensemble}
\end{equation}

By additivity, the same averaging applies componentwise to centered exposure--lag functions, covariate effects, and their derived cumulative contrasts. Reported summaries are obtained only after this averaging by applying the appropriate transformation from the link scale; Section~\ref{sec:last_layer_laplace} extends the ensemble construction to pointwise uncertainty.

\subsection{Neural Component Architecture}

For continuous components that must resolve local nonlinear structure, we use exp-centered hidden units (ExU) \cite{agarwal2021neural} in the input layer; simpler covariate components may instead use linear inputs. Let \(z\in\mathbb{R}^{p}\) denote the input to a network component. A standard input layer maps \(z\) affinely and then applies an elementwise activation \(\phi\), \(h:\mathbb{R}^{p}\to\mathbb{R}^{d}\),
\[
  h(z)=\phi(Wz+b),
  \qquad
  W\in\mathbb{R}^{d\times p},\quad b\in\mathbb{R}^{d},
\]
where \(W\) is the weight matrix and \(b\) the bias vector. ExU preserves this map-then-activation structure but reparameterizes the input map through log-scale weights and unit-specific locations. The scalar ExU used in NAMs acts on a univariate input \(z_j\). For \(h_j:\mathbb{R}\to\mathbb{R}^{d_j}\) and \(w_j,b_j\in\mathbb{R}^{d_j}\),
\begin{equation}
  h_j(z_j)
  =
  \phi\left(e^{w_j}\odot(z_j\mathbf{1}_{d_j}-b_j)\right),
  \label{eq:dlnam_exu}
\end{equation}
where \(\odot\) denotes the Hadamard product, \(w_j\) are log-scale weights, and \(b_j\) are location biases. Exponential scaling gives each unit locally adjustable sensitivity around \(b_j\), allowing the input map to resolve thresholds and rapid transitions. For multivariate inputs, we extend the scalar ExU coordinate-wise by concatenation, with \(d=\sum_{j=1}^{p}d_j\):
\begin{equation}
  h(z)
  =
  \bigl(h_1(z_1)^\top,\ldots,h_p(z_p)^\top\bigr)^\top
  \in\mathbb{R}^{d}.
  \label{eq:dlnam_concat}
\end{equation}
Alternative multivariate generalizations are compared in Supplementary Text~\ref{supp:exu}. For an exposure--lag surface, \(h(x,\ell)=(h_x(x)^\top,h_\ell(\ell)^\top)^\top\), with the layer width split as evenly as possible between the two coordinates, \(d_x=\lceil d/2\rceil\) and \(d_\ell=\lfloor d/2\rfloor\).

The architecture pairs ExU with smooth activations rather than the capped ReLU-\(n\) used in the original NAM construction \cite{agarwal2021neural}, where capping localizes individual ExU units to a restricted input range. We consider two closely related smooth activation functions, Mish \cite{misra2020mish} and SiLU \cite{elfwing2018silu}, both of which provide smooth ReLU-like nonlinearities. Mish is particularly attractive for its smooth profile and because its original study reported performance matching or improving upon SiLU (Swish-1) across several deep-network benchmarks. Their activation functions are defined as
\[
  \phi_{\mathrm{Mish}}(z)
  =
  z\,\tanh\!\left(\operatorname{softplus}(z)\right),
  \qquad
  \phi_{\mathrm{SiLU}}(z)
  =
  z\,\sigma(z),
\]
where \(\operatorname{softplus}(z)=\log(1+e^z)\) is the Softplus function and \(\sigma(z)=(1+e^{-z})^{-1}\) is the logistic sigmoid. Their continuous derivatives preserve a smooth reported response. The reported fits use Mish; SiLU is a close alternative from the same family and was not separately evaluated. The division of labor is deliberate: ExU controls the location and scale of sensitivity, while the hidden activation controls how smoothly that sensitivity propagates through the component. This architectural distinction is evaluated in the ablation study (Fig.~\ref{fig:abl}).

\subsection{Optimization}
\label{sec:optimization}

Parameters are estimated by minimizing the average negative log-likelihood,
\begin{equation}
  \mathcal{L}(\theta)
  =
  -\frac{1}{n}\sum_{t=1}^{n}
  \log p\{Y_t\mid \mu_t(\theta)\},
  \label{eq:dlnam_loss}
\end{equation}
where \(p\) denotes the outcome likelihood or working likelihood as appropriate, \(\eta_t(\theta)\) denotes the predictor in Equation~\eqref{eq:dlnam_predictor_general}, and \(\mu_t(\theta)=g^{-1}(\eta_t(\theta))\). Optimization uses AdamW \cite{loshchilov2019adamw} under cosine annealing \cite{loshchilov2017sgdr}, with gradient-norm clipping and full-batch updates unless otherwise stated. In the quasi-Poisson application, optimization proceeds under a Poisson working likelihood \cite{wedderburn1974quasi}, with overdispersion estimated post hoc from the Pearson dispersion and carried into the quasi-likelihood uncertainty calculation of Section~\ref{sec:last_layer_laplace}.

The reported fits use weight decay, with coefficient \(\lambda_w\), as their principal explicit regularizer, together with a surface-roughness penalty in the Chicago analysis. Standard dropout, subnetwork dropout, and an output penalty were also implemented and examined in preliminary simulation work, but did not yield consistent improvements in recovery and were therefore inactive in the reported analyses. The Chicago roughness penalty is applied because the unpenalized fit is sensitive to the training budget beyond roughly 3000 epochs, whereas under the penalty the fitted surface is stable across budgets from 2500 to 10000. Beyond explicit regularization, the architecture and fitting procedure shape the learned representation in several distinct ways: smooth activations favor smooth component functions, mixture composition represents each component through jointly trained subnetworks rather than a single network, and ensemble averaging reduces variation attributable to random initialization. Optimizer, scheduler, batching, gradient-clipping, and regularization settings are reported in Table~\ref{tab:supp_fitting}.

Input scaling is matched to the parameterization of the first layer and is treated as part of the fitted model. Inputs to ExU layers are min--max scaled to \([0,1]\), aligning the observed domain with the initialization range of the location biases, whereas linear inputs are standardized by their analyzed-sample mean and standard deviation. Exposure scaling pools the raw values supplied across a term's observed lag columns, giving each exposure a common scale across its lag history; lag coordinates are scaled using the fixed lag range used for fitting. Data-dependent scaling constants are estimated once within each fitted data set, and all transformations are then held fixed for training and effect extraction, so evaluation-grid values are transformed using the fitted scaling constants rather than rescaled to the evaluation grid.

Initialization likewise reflects the role of each parameter in the architecture. ExU log-weights are drawn from \(\mathcal{N}(\mu_w,\sigma_w^2)\) and location biases from \(\mathcal{U}(0,1)\), so that \(\mu_w\) controls the initial sensitivity scale through \(e^w\) while \(\sigma_w\) disperses that sensitivity across units. Ordinary linear layers use Kaiming uniform initialization \cite{he2015kaiming}, and mixing weights are initialized independently from \(\mathcal{N}(0,\sigma_\omega^2)\) across components, subnetworks, and ensemble members.

\subsection{Last-Layer Laplace Confidence Intervals}
\label{sec:last_layer_laplace}

Uncertainty is quantified with a target-specific last-layer Laplace approximation applied separately to each ensemble member, related to the Laplace treatment of NAMs developed by \textcite{bouchiat2024bayesian}. The approximation conditions on the learned representation and, in the mixture formulation, the mixing weights; the resulting member-level conditional uncertainty is combined with between-member variation. Supplementary Text~\ref{supp:laplace} gives the full derivation and regularity conditions, and frequentist calibration is evaluated directly in simulation.

For a given inferential target, partition the parameter vector as \(\theta=(\psi,\xi)\), where \(\xi\in\mathbb{R}^{q}\) contains the global intercept and the final-layer parameters of the corresponding component, while \(\psi\) contains all remaining parameters. Conditional on the fitted representation \(\hat\psi\), the predictor is affine in \(\xi\), reducing the calculation to a generalized linear model in a low-dimensional last-layer parameterization. The mixing weights remain conditioned because allowing them to vary jointly with the corresponding output layers would introduce a bilinear scale non-identifiability.

Let \(\eta(\xi)=(\eta_1(\xi),\ldots,\eta_n(\xi))^\top\) denote the vector of linear predictors across the \(n\) observations, with each \(\eta_t\) given by Equation~\eqref{eq:dlnam_predictor_general}, and define the last-layer Jacobian
\begin{equation}
  \Phi
  =
  \frac{\partial\eta(\xi)}{\partial\xi^\top}
  \in\mathbb{R}^{n\times q},
  \label{eq:main_phi}
\end{equation}
which plays the same design-matrix role as the cross-basis in a DLNM. Let \(W=\operatorname{diag}(w_1,\ldots,w_n)\), where \(w_t=(\mathrm{d}\mu_t/\mathrm{d}\eta_t)^2/V(\mu_t)\) are the generalized-linear-model working weights evaluated at the fitted means and \(V\) is the variance function. Let \(\hat\varphi\) denote the dispersion or working-dispersion factor appropriate to the fitted outcome model, taking its family-specified value when fixed and being estimated where required. The dispersion-adjusted last-layer information is
\begin{equation}
  H
  =
  \Phi^\top W\Phi/\hat\varphi .
  \label{eq:main_H}
\end{equation}
Because \(\eta(\xi)\) is affine in \(\xi\), \(\Phi\) is exact and constant and second derivatives of the predictor with respect to \(\xi\) vanish, so no predictor-curvature term enters the Hessian. For canonical likelihood fits, \(H\) coincides with the observed likelihood information; for non-canonical links, it is the corresponding Fisher information, while for working-likelihood fits it is interpreted as dispersion-scaled working information.

For the post-hoc Laplace approximation, a zero-mean Gaussian prior with precision \(\lambda\) is placed on the final-layer parameters but not on the intercept, a split encoded by the diagonal selector \(P_\xi\). The precision \(\hat\lambda\) is selected by a MacKay evidence fixed-point approximation with the unpenalized intercept profiled out \cite{mackay1992interpolation,mackay1992practical}. Because \(\hat\lambda\) is selected after training, the fitted \(\hat\xi\) is not re-optimized under the Gaussian prior; it is retained as the expansion center and treated as an approximate posterior mode. The conditional last-layer covariance is
\begin{equation}
  \widehat\Sigma_\xi
  =
  \bigl(H+\hat\lambda P_\xi+\epsilon I\bigr)^{-1},
  \label{eq:main_covariance}
\end{equation}
where \(\epsilon>0\) is a numerical ridge.

The inferential targets are the centered components of Equation~\eqref{eq:dlnam_predictor_general}: the exposure--lag surface \(f_m^{\hat\psi,\xi}\), its cumulative
\begin{equation}
  F_m^{\hat\psi,\xi}(\cdot)
  =
  \sum_{\ell\in\mathcal L_m}
  f_m^{\hat\psi,\xi}(\cdot,\ell),
  \label{eq:main_cumulative}
\end{equation}
and the covariate shape function \(h_k^{\hat\psi,\xi}\). For
\[
  \tau^{\hat\psi,\xi}
  \in
  \left\{
    f_m^{\hat\psi,\xi},
    F_m^{\hat\psi,\xi},
    h_k^{\hat\psi,\xi}
  \right\}
\]
evaluated at point \(\zeta\), let
\[
  J_\zeta
  =
  \frac{\partial\tau^{\hat\psi,\xi}(\zeta)}
       {\partial\xi^\top}
\]
denote its gradient. Conditional on \(\hat\psi\), these targets are affine in \(\xi\), so \(J_\zeta\) is constant and covariance propagation is exact, with no delta-method linearization error. The pointwise conditional standard deviation on the link scale is
\begin{equation}
  \operatorname{sd}\{\tau^{\hat\psi,\xi}(\zeta)\}
  =
  \bigl\{
    J_\zeta\widehat\Sigma_\xi J_\zeta^\top
  \bigr\}^{1/2}.
  \label{eq:main_sd}
\end{equation}
Pointwise \(1-\alpha\) intervals are first formed on the link-contrast scale and, where the reported estimand is a monotone transformation of that contrast, their endpoints are transformed accordingly. When an absolute linear predictor is reported on the mean scale, the transformation is \(g^{-1}\).

For the ensemble, write
\[
  \hat\psi=(\hat\psi_1,\ldots,\hat\psi_B),
  \qquad
  \hat\xi=(\hat\xi_1,\ldots,\hat\xi_B)
\]
for the collections of fitted member-specific conditioned parameters and last-layer expansion centers, respectively. Applying the ensemble construction of Equation~\eqref{eq:dlnam_ensemble} to a reported target, the interval is centered at the ensemble mean and its variance combines average member-level conditional uncertainty with between-member variation,
\begin{equation}
  \hat\tau^{\hat\psi,\hat\xi}(\zeta)
  =
  \frac{1}{B}\sum_{b=1}^{B}
  \hat\tau_b^{\hat\psi_b,\hat\xi_b}(\zeta),
  \qquad
  \widehat{\mathcal V}(\zeta)
  =
  \underbrace{\frac{1}{B}\sum_{b=1}^{B}
  J_{b\zeta}\widehat\Sigma_{\xi_b}J_{b\zeta}^{\top}}_{\text{within-member}}
  +
  \underbrace{\frac{1}{B}\sum_{b=1}^{B}
  \left\{
  \hat\tau_b^{\hat\psi_b,\hat\xi_b}(\zeta)
  -
  \hat\tau^{\hat\psi,\hat\xi}(\zeta)
  \right\}^{2}}_{\text{between-member}} .
  \label{eq:main_ensemble}
\end{equation}
Combining the conditional last-layer Laplace uncertainty with between-member variation, the reported pointwise \(1-\alpha\) confidence interval is
\begin{equation}
  \hat\tau^{\hat\psi,\hat\xi}(\zeta)
  \pm
  z_{1-\alpha/2}\widehat{\mathcal V}(\zeta)^{1/2},
  \label{eq:main_interval}
\end{equation}
with the same transformation to the reported scale as above. The first term in Equation~\eqref{eq:main_ensemble} averages member-specific conditional last-layer variances without further division by \(B\), while the second records between-member variation. Consequently, \(\hat\tau^{\hat\psi,\hat\xi}(\zeta)\) and \(\widehat{\mathcal V}(\zeta)\) are the mean and variance of the equally weighted mixture of member-level Gaussian Laplace approximations. The reported interval is a Gaussian moment-matched summary of this generally non-Gaussian mixture; its variance is therefore a constructed uncertainty measure, not the sampling variance of the ensemble mean or full posterior uncertainty in the learned representation.

\subsection{Simulation Design}
\label{sec:mc_design}

For each DGP, we simulated a single-exposure distributed lag Poisson model with lag horizon \(L=14\), \(n=5000\) daily observations, and reference exposure \(x^\star=20\):
\[
  Y_t\mid\mathcal H_t
  \sim
  \operatorname{Poisson}(\mu_t),
  \qquad
  \log\mu_t
  =
  \alpha
  +
  \sum_{\ell=0}^{L}
  f(x_{t-\ell},\ell),
\]
with \(\alpha=\log 50\). Exposure followed a seasonal autoregressive process,
\[
  x_t
  =
  20
  +
  12\sin(2\pi t/365)
  +
  a_t,
  \qquad
  a_t
  =
  0.8\,a_{t-1}
  +
  \varepsilon_t,
  \qquad
  \varepsilon_t
  \sim
  \mathcal N(0,2^2),
\]
with \(a_1\) drawn from its stationary distribution. Each replicate included an additional \(L\)-observation exposure prehistory, leaving \(n=5000\) analyzed observations with complete lag histories. Effects were evaluated on the equally spaced grid
\[
  \mathcal G
  =
  \{0,0.2,\ldots,40\},
\]
which includes the reference \(x^\star\).

The four generating surfaces \(f_1,\ldots,f_4\) were constructed from sigmoidal exposure responses, Gaussian bumps, and normalized lag kernels; Supplementary Text~\ref{supp:dgp} gives their exact definitions. DGP 1 is separable, with a smooth exposure response modulated by a single exponentially decaying lag kernel across the exposure domain. DGPs 2--4 are nonseparable and differ in how lag structure varies across the exposure domain: DGP 2 combines early and late effects with opposing sign at intermediate lags, DGP 3 combines a narrow high-exposure response with a broader low-exposure response on a slower kernel, and DGP 4 places sharp responses at both extremes of the exposure range with distinct dominant lag timing. All four data-generating surfaces are smooth and continuous, so the spline-based comparators are evaluated on targets compatible with their intended function class rather than on discontinuous or otherwise deliberately unfavorable alternatives.

\subsection{Evaluation Metrics}
\label{sec:metrics}

The likelihood depends on each exposure history through the aggregate contribution \(\sum_{\ell=0}^{L}f(x_{t-\ell},\ell)\). Under the log link, we take the cumulative effect
\[
  F(x)
  =
  \sum_{\ell=0}^{L}f(x,\ell)
\]
as the primary estimand, corresponding to the cumulative log-relative-risk summary conventionally reported in distributed lag nonlinear model analyses \cite{gasparrini2011dlnmR,gasparrini2017penalised,mork2022tdlnm}. This contrast holds exposure fixed across lags and sums the resulting surface values. We additionally evaluate \(f\) as a complementary diagnostic of lag-specific allocation. Because individual surface locations are not observed by the likelihood in isolation, surface scoring evaluates a finer decomposition of the fitted effect rather than a more directly fitted or intrinsically superior target.

Let \(\tau\in\{F,f\}\) denote the evaluation target, with target-specific grids
\[
  \mathcal G_F
  =
  \mathcal G,
  \qquad
  \mathcal G_f
  =
  \mathcal G\times\{0,\ldots,L\}.
\]
Across \(R\) Monte Carlo replicates, define
\[
  d_r^{\tau}(\zeta)
  =
  \hat\tau_r(\zeta)-\tau(\zeta),
  \qquad
  \bar d^{\tau}(\zeta)
  =
  \frac{1}{R}\sum_{r=1}^{R}d_r^{\tau}(\zeta),
  \qquad
  \zeta\in\mathcal G_\tau.
\]
For each target, summaries are reported over the full grid, the boundary region below the \(5\)th or above the \(95\)th empirical percentile of a fixed reference exposure series, and its interior complement. The reference series is generated once from the simulation exposure process at the nominal sample size and held fixed across replicates and estimators, ensuring identical evaluation regions between fits. Surface regions are defined by the exposure coordinate and include every lag. Writing \(\mathcal A_\tau\subseteq\mathcal G_\tau\) for any such region,
\[
  \operatorname{RMSE}_\tau(\mathcal A_\tau)
  =
  \sqrt{
    \frac{1}{R}
    \sum_{r=1}^{R}
    \frac{1}{|\mathcal A_\tau|}
    \sum_{\zeta\in\mathcal A_\tau}
    \{d_r^\tau(\zeta)\}^2
  },
  \qquad
  \operatorname{Bias}_\tau^2(\mathcal A_\tau)
  =
  \frac{1}{|\mathcal A_\tau|}
  \sum_{\zeta\in\mathcal A_\tau}
  \{\bar d^\tau(\zeta)\}^{2},
\]
and
\[
  \operatorname{Var}_\tau(\mathcal A_\tau)
  =
  \frac{1}{R}
  \sum_{r=1}^{R}
  \frac{1}{|\mathcal A_\tau|}
  \sum_{\zeta\in\mathcal A_\tau}
  \{d_r^\tau(\zeta)-\bar d^\tau(\zeta)\}^2 .
\]
Under these definitions,
\[
  \operatorname{RMSE}_\tau^2(\mathcal A_\tau)
  =
  \operatorname{Bias}_\tau^2(\mathcal A_\tau)
  +
  \operatorname{Var}_\tau(\mathcal A_\tau),
\]
so squared bias and variance partition the mean squared error exactly; the reported bias and variance shares are their corresponding proportions of \(\operatorname{RMSE}_\tau^2(\mathcal A_\tau)\).

If \(I_r^\tau(\zeta)\) denotes the pointwise \(95\%\) interval for target \(\tau\) in replicate \(r\), empirical coverage is
\[
  \operatorname{Cov}_\tau(\mathcal A_\tau)
  =
  \frac{1}{R}
  \sum_{r=1}^{R}
  \frac{1}{|\mathcal A_\tau|}
  \sum_{\zeta\in\mathcal A_\tau}
  \mathbf 1
  \bigl\{
    \tau(\zeta)\in I_r^\tau(\zeta)
  \bigr\}.
\]
For DLNAM, \(I_r^\tau(\zeta)\) is the combined last-layer Laplace and ensemble interval of Equation~\eqref{eq:main_interval}. This evaluates pointwise coverage of the fixed generating function, not simultaneous coverage over \(\mathcal G_\tau\).

Monte Carlo uncertainty is calculated from replicate-level summaries, thereby accounting for dependence among evaluation-grid points within each replicate rather than treating them as independent Monte Carlo replicates. For any reported summary \(\widehat S\) with Monte Carlo standard error \(\operatorname{MCSE}(\widehat S)\), plotted \(95\%\) Monte Carlo confidence intervals use the normal approximation
\[
  \widehat S
  \pm
  1.96\,\operatorname{MCSE}(\widehat S).
\]
Coverage MCSEs use replicate-level variability directly; RMSE uncertainty is obtained by applying the delta method to replicate-wise mean squared error, while regional squared-bias and variance uncertainties use delete-one-replicate jackknife estimates. The complete replicate-level, delta-method, and jackknife calculations are given in Supplementary Text~\ref{supp:mcse}.

\subsection{Model Specifications and Fitting}
\label{sec:estimators_fitting}

All estimators were fitted to the same replicate data sets, evaluated on the same grids, and centered at the same reference value. DLNAM and the spline-based comparators used the log link. DLNAM minimized the negative average Poisson log-likelihood in Equation~\eqref{eq:dlnam_loss}; DLNM (QAIC), DLNM (QBIC), and P-DLNM were fitted as quasi-Poisson regressions, with the Poisson log-likelihood at the fitted means entering the QAIC and QBIC calculations. The likelihood difference for T-DLNM is described below. Comparator configurations and, where applicable, tuning ranges were anchored in the originating methodological papers and their simulation studies rather than optimized for the present DGPs; basis- or outcome-specific adaptations and departures required by the application or multi-exposure setting are stated explicitly below. The comparisons should therefore be read as benchmarks against established, literature-grounded implementations in the settings they were designed to address, rather than against deliberately weak baselines.

\paragraph{DLNAM.}
The reported DLNAM used 3 subnetworks per component and 3 independently initialized complete fits, with uncertainty constructed as in Section~\ref{sec:last_layer_laplace}. Each surface subnetwork comprised a 128-unit ExU input layer, one 128-unit dense hidden layer with a Mish activation, and a linear output. ExU log-weight initialization used \(\mu_x=1.5\) for exposure and \(\mu_\ell=2.5\) for lag, with \(\sigma_w=0.5\) for both. The larger \(\mu_\ell\) gives greater initial sensitivity along the lag coordinate, reflecting the expectation of more rapid variation at short lags that also motivates logarithmic lag-knot placement in conventional DLNMs \cite{gasparrini2011dlnmR}; the ExU scales themselves remain learned parameters. This surface-component architecture was fixed before the comparison and carried across the reported analyses, while optimization budgets were allowed to vary where required by problem size. Developed in preliminary simulation work, the configuration is treated as an empirically supported specification rather than as a globally optimal design.

The principal simulation and Chicago analysis used 5000 configured full-batch epochs, with no early stopping. Study-specific optimization departures for the joint-exposure simulation and malaria analysis are described in their respective sections. Exact parameter counts and remaining fitting settings are given in Table~\ref{tab:supp_fitting}.

\paragraph{DLNM.}
The two criterion-selected variants, DLNM (QAIC) and DLNM (QBIC), follow the conventional DLNM framework of \textcite{gasparrini2010dlnm} and the \texttt{dlnm} implementation~\cite{gasparrini2011dlnmR}: quasi-Poisson log-linear fitting with an exposure--lag cross-basis, natural cubic spline marginal bases, equally spaced exposure knots, logarithmically spaced lag knots, and information-criterion selection of basis complexity. Criterion-based selection of cross-basis complexity has also been evaluated in simulation studies \cite{gasparrini2014elr}. The candidate range was anchored in prior methodological simulation studies: \textcite{gasparrini2017penalised} selected the complexity of an unpenalized DLNM over combinations spanning 1--10 degrees of freedom in each marginal dimension, a range subsequently used for the corresponding spline comparator in the T-DLNM simulation study of \textcite{mork2022tdlnm}. We use the corresponding range \(\{2,\ldots,10\}\) in each margin, excluding one degree of freedom because under the present natural-spline construction it contains no interior knot and reduces the margin to a linear term. Thus, in simulation, both criteria are evaluated over \(\{2,\ldots,10\}^{2}\), giving 81 candidate cross-bases per replicate. Chicago uses the narrower range \(\{2,\ldots,5\}^{2}\). Sensitivity analysis of this model structure reports that increasing the number of exposure knots produces a visibly less smooth exposure--response curve attributable to overfitting, while the lag dimension is comparatively insensitive to basis choice \cite{gasparrini2010dlnm}; the applied range is restricted accordingly. On this series both criteria select the largest available exposure dimension, so the upper limit acts as a specification choice rather than as a search boundary, and widening it reproduces that instability, with QAIC placing minimum mortality at \(-19.7\,^{\circ}\mathrm{C}\).

Pointwise Wald intervals are formed from the fitted cross-basis contrasts and coefficient covariance. QAIC and QBIC differ in their complexity penalty, with QBIC penalizing larger cross-bases more strongly at the present sample sizes; exact criterion formulas are reported in Supplementary Text~\ref{supp:fitting}. For these quasi-Poisson fits, the estimated dispersion enters both the coefficient covariance and the criterion calculations.

\paragraph{Penalized DLNM.}
P-DLNM follows the penalized framework of \textcite{gasparrini2017penalised}, implemented using the \texttt{dlnm} cross-basis construction and \texttt{mgcv} penalized-GAM machinery \cite{gasparrini2011dlnmR,wood2017gam}. Marginal cubic P-spline bases are coupled through the framework's cross-basis penalty construction, with second-order difference penalties and smoothing parameters estimated by restricted maximum likelihood. In simulation, we use the rank-10 marginal specification used in the P-DLNM simulation study, giving \(10\times10\); the same rank-10, second-order-penalty, REML specification was subsequently used as the penalized spline comparator in the T-DLNM simulation study \cite{mork2022tdlnm}. In simulation, realized effective degrees of freedom reached at most 57\% of the available rank. On the Chicago series the penalty does not play the same role: realized effective degrees of freedom track the imposed rank closely, reaching 78\% of the basis at the specification used, and both fitted complexity and the roughness of the cumulative curve increase with rank rather than stabilizing. Marginal rank is therefore a specification choice on this series rather than an upper bound resolved by the penalty. Chicago accordingly uses a fixed \(5\times5\) specification, matching the dimension available to the criterion-selected variants, so that the spline comparators differ in how complexity is controlled rather than in the dimension available to them. Pointwise Wald intervals use the fitted quasi-Poisson coefficient covariance, including its estimated dispersion.

\paragraph{Treed DLNM.}
T-DLNM uses the nonlinear Treed Distributed Lag Model of \textcite{mork2022tdlnm}, implemented in the authors' \texttt{dlmtree} package~\cite{im2025dlmtree}. We retain the tree priors and sampler configuration used in the authors' simulation study rather than retuning T-DLNM to the present DGPs; exact tree and Markov chain settings are reported in Table~\ref{tab:supp_comparators}. Because the \texttt{dlmtree} version used here does not accept a count likelihood in its nonlinear implementation, T-DLNM was fitted with a Gaussian likelihood for \(\log(1+Y_t)\). It therefore serves as a benchmark for the flexible treed method class rather than as a likelihood-matched quasi-Poisson comparator. Its exponentiated contrasts in \(\mathbb E[\log(1+Y)]\) are used as a comparison-scale proxy for relative risk; they are not algebraically identical to contrasts in \(\log\{\mathbb E(Y)\}\). Posterior credible intervals were extracted on the common evaluation grid and reference. Implementation-specific fixed-effect and retry details are given in Supplementary Text~\ref{supp:fitting}.

\subsection{Joint-Exposure Simulation}
\label{sec:joint_mc_design}

The multi-exposure recovery study reuses the four data-generating surfaces of Section~\ref{sec:mc_design} as concurrent additive effects. Five exposure series share the marginal seasonal AR(1) structure of the single-exposure simulation; the four active series have seasonal phase shifts \(0,\pi/6,\pi/3,\pi/2\), while the null series has phase \(0\). All five innovation processes are equicorrelated at \(0.5\), and the null exposure has an identically zero true surface. The Poisson log mean is the intercept plus the four active distributed-lag contributions. The final study comprises \(R=50\) independent data sets of \(n=5000\) observations with the same lag horizon, reference value, and evaluation grid as the principal simulation.

The joint DLNAM contains one exposure--lag component per exposure and otherwise retains the specification of the principal simulation. For the larger five-component model, the initial learning rate was reduced from \(8\cdot10^{-4}\) to \(7\cdot10^{-4}\), cosine annealing still reached \(1\cdot10^{-4}\), and training was extended to \(5000\) full-batch epochs. Preliminary stability checks at the baseline optimization setting showed occasional high-loss convergence and material under-training after \(2500\) epochs.

The comparator implementations reflect the feasible multi-exposure form of each method. DLNM (QAIC) and DLNM (QBIC) use one sequential coordinate-wise sweep, initialized at \(4\times4\) marginal dimensions for all five components and updated in the order DGP 1, DGP 2, DGP 3, DGP 4, then the null exposure. For each exposure, its marginal dimensions are selected over \(\{2,\ldots,10\}^{2}\) while the current dimensions of the other exposures are held fixed, followed by one joint quasi-Poisson fit. This practical strategy is potentially order-dependent and is not an exhaustive combinatorial search: with 81 candidate cross-bases per exposure, the coordinate-wise sweep requires \(5\times81\) fits, whereas an exhaustive joint search over the five components would require \(81^{5}\approx3.5\times10^{9}\). P-DLNM fits all penalized cross-bases jointly with REML. T-DLNM is fitted separately for each target exposure, with the other concurrent exposures represented by fixed \(4\times4\) natural-spline cross-bases; its target component retains the Gaussian likelihood on \(\log(1+Y)\) used in the single-exposure comparison.

Recovery is scored for each active component on the cumulative target using the full-grid, interior, and boundary regions and pointwise coverage definition of Section~\ref{sec:metrics}. The null component is scored against zero as a leakage diagnostic under the specified correlation structure. For each active exposure and evaluation region, degradation is defined as the ratio of joint-fit RMSE to the corresponding single-exposure RMSE, with values above one denoting worse recovery in the joint setting. When uncertainty is attached to this ratio, Monte Carlo uncertainty from both RMSE estimates is propagated using the delta-method approximation defined in Supplementary Text~\ref{supp:mcse}.

\subsection{Chicago Temperature and Mortality Analysis}
\label{sec:chicago_design}

We analyzed the Chicago mortality series distributed with the R \texttt{dlnm} package using the adjustment specification of \textcite{gasparrini2010dlnm}, with \(\mathrm{PM}_{10}\) in place of carbon monoxide, which is not included in the distributed Chicago series. The analysis is not intended to reproduce a published model: all estimators use the same adjustment variables, while their functional representations differ. Daily all-cause mortality is modeled as a function of temperature over lags \(0\)--\(30\), centered at the observed median of \(10.3\,^\circ\mathrm{C}\). Shared adjustment variables are date, day of week, lag \(0\)--\(1\) mean dew point temperature, ozone, and \(\mathrm{PM}_{10}\). Dropping days with any missing model variable leaves 4638 complete days.

DLNM (QAIC), DLNM (QBIC), and P-DLNM use quasi-Poisson regression. DLNAM uses a Poisson working likelihood with post-hoc Pearson dispersion \(\hat\varphi=1.03\) for the reported Laplace intervals, while T-DLNM retains its Gaussian model for \(\log(1+Y_t)\). DLNAM retains the exposure--lag surface architecture used in simulation, while the adjustment components use architectures tailored to their roles. Date is represented by a deeper ExU trend network, with an additional 128-unit hidden layer and a larger ExU log-weight mean, \(\mu_w=4.5\), to accommodate variation across multiple temporal scales; the remaining continuous covariates use separate 32-unit Mish networks with linear inputs after standardization. This allocation follows the relative flexibility of the adjustment specification itself, which assigns 98 degrees of freedom to the long-term trend and at most three to the remaining continuous covariates. The spline and treed comparators follow Section~\ref{sec:estimators_fitting}, including its Chicago-specific dimensions. Full architecture, adjustment, and comparator specifications are reported in Tables~\ref{tab:supp_fitting} and~\ref{tab:supp_comparators}.

The analysis is observational and in-sample. It assesses whether the estimators yield a familiar empirical temperature--mortality pattern on a canonical DLNM data set and therefore serves as an empirical plausibility analysis rather than an evaluation of causal effects or forecasting performance.

\subsection{Multi-Exposure Malaria Analysis}
\label{sec:malaria_design}

The second application uses 348{,}565 individual childhood malaria outcomes from a large multi-country survey \cite{martellini2026malaria}. Five environmental exposures are modeled: mean temperature, precipitation, soil moisture, actual evapotranspiration, and specific humidity. DLNAM uses a Bernoulli likelihood with a logit link and 6 monthly lags spanning months 1--6 before testing, with effects centered at each exposure's median.

One DLNAM is fitted per reported exposure using the target-specific adjustment set of the source study. Exposures treated there as confounders enter alongside the target, whereas mediators are omitted. The adjustment set therefore varies by target, ranging from one additional exposure for temperature and precipitation to four for specific humidity. Table~\ref{tab:supp_comparators} gives the exact target-to-adjustment mapping; each reported curve comes from the fit in which that exposure is the target. In the DLNAM fits, every adjusting exposure retains a full exposure--lag surface, so a target-specific fit contains up to 5 concurrently lagged surfaces. Month and survey year enter through 32-unit Mish covariate networks with standardized linear inputs, while survey cluster and country enter as weight-decayed level-specific effects, permitting joint minibatch optimization; variance-component random intercepts would require a hierarchical extension not implemented here. Fits were trained for 50 configured epochs using minibatches containing 1\% of the data rather than the full-batch updates used in the smaller analyses. For the reported malaria intervals, the Laplace approximation includes the target exposure's final-layer parameters together with the global intercept; all remaining fitted parameters, including the adjusting-exposure networks, month/year covariate networks, and cluster/country effects, are conditioned upon.

The DLNM uses the same target-specific environmental adjustment sets and is likewise fitted one target at a time. The target exposure uses a natural-spline cross-basis with 3 degrees of freedom in each exposure and lag margin over months 1--6, while adjusting exposures enter as scalar lag means. Month is represented by a 4-degree-of-freedom spline, survey year enters linearly, and cluster and country enter as random intercepts estimated by \texttt{glmmTMB} \cite{brooks2017glmmtmb}. Thus, the two analyses use the same environmental control sets but differ in the functional representation of those controls and in their treatment of the survey hierarchy: cluster and country effects are shrunk by a fixed weight-decay penalty in the DLNAM and by an estimated variance component in the DLNM, so both are pooled but only the latter estimates the degree of pooling. The comparison therefore contrasts two complete analysis specifications rather than isolating the effect of estimator choice. The application assesses feasibility and interpretation in a large, multi-exposure observational setting rather than comparative recovery. Full architecture, optimization, adjustment, and sampling specifications are reported in Table~\ref{tab:supp_fitting}.

\section*{Declarations}

\paragraph{Acknowledgments.}
We are grateful to Manuel Martellini O Nocentini, whose suggestions helped shape the experimental design, comparator selection, and evaluation criteria of the simulation study.

\paragraph{Funding.}
This research was supported by Formas, the Swedish Research Council for
Sustainable Development (grant nos. 2023-01774 and 2022-01845);
Forte, the Swedish Research Council for Health, Working Life and Welfare
(grant no. 2024-00833); and the Swedish Research Council
(grant nos. 2025-02804 and 2022-06599).

\paragraph{Computational resources.}
The computational work presented in this study was supported by
high-performance computing resources provided by the National Academic
Infrastructure for Supercomputing in Sweden (NAISS), funded by the Swedish
Research Council. These resources were used for model development, training,
simulation experiments, and computational testing.

\paragraph{Author contributions.}
Conceptualization: C.H., S.P., and L.O.; Data curation: C.H. and S.P.; Formal analysis: C.H.; Investigation: C.H. and S.P.; Methodology: C.H., S.P., and L.O.; Project administration: C.H., S.P., L.O., and E.R.; Resources: S.P. and L.O.; Software: C.H. and S.P.; Supervision: S.P., L.O., and E.R.; Validation: C.H., S.P., L.O., and E.R.; Visualization: C.H.; Writing – original draft: C.H.; Writing – review and editing: C.H., S.P., L.O., and E.R.

All authors reviewed and approved the final version of the manuscript.

\paragraph{Competing interests.}
The authors declare no competing interests.

\paragraph{Data availability.}
The simulated data sets generated for this study can be reproduced using the simulation code (see Code availability) and are also available from the corresponding author upon reasonable request. The Chicago NMMAPS data set is publicly available through the \texttt{dlnm} R package. The malaria data are subject to the data-access conditions of the original study and may be available upon request from the authors of that study.

\paragraph{Code availability.}
The source code implementing DLNAM, together with scripts for the simulation
studies and analyses presented in this manuscript, is publicly available on
GitHub at \url{https://github.com/shivangpandey31/DLNAM}, and is provided
without the underlying restricted-access data for the malaria analysis. The
version corresponding to this manuscript is archived on Zenodo (DOI: \url{https://doi.org/10.5281/zenodo.22288964}).

\paragraph{Ethics statement.}
This methodological study involved no new collection of individual-level participant data. The Chicago NMMAPS data set is publicly available. The malaria analysis used previously collected data, for which the relevant ethical approvals and informed-consent procedures are described in the original study and corresponding data sources.

\paragraph{AI use.}
Generative artificial intelligence tools were used during the development of this work to assist with coding and to improve sentence structure, grammar, and language clarity. All AI-assisted code and text were reviewed and verified by the authors, who take full responsibility for the final manuscript, analyses, and interpretation of the results.

\paragraph{Prior dissemination.}
This article is based on and extends work first presented in C.H.'s master's
thesis at KTH Royal Institute of Technology, carried out at Karolinska
Institutet and supervised by S.P., L.O., and E.R.

\pdfbookmark[1]{References}{references}

\printbibliography

\clearpage
\setcounter{section}{0}
\setcounter{figure}{0}
\setcounter{table}{0}
\setcounter{equation}{0}
\renewcommand{\thesection}{S\arabic{section}}
\renewcommand{\thefigure}{S\arabic{figure}}
\renewcommand{\thetable}{S\arabic{table}}
\renewcommand{\theequation}{S\arabic{equation}}
\renewcommand{\theHsection}{supp.section.\arabic{section}}
\renewcommand{\theHfigure}{supp.figure.\arabic{figure}}
\renewcommand{\theHtable}{supp.table.\arabic{table}}
\renewcommand{\theHequation}{supp.equation.\arabic{equation}}

\pdfbookmark[1]{Supplementary Materials}{supplementary-materials}
\begin{center}
  {\Large\bfseries Supplementary Materials\par}
\end{center}
\vspace{0.8em}
\makeatletter
\def\toclevel@section{2}
\def\toclevel@subsection{3}
\makeatother
\section{Notation}
\label{supp:notation}

\begin{table}[!ht]
\centering
\footnotesize
\renewcommand{\arraystretch}{1.04}
\setlength{\tabcolsep}{4.5pt}
\begin{tabularx}{\textwidth}{>{\raggedright\arraybackslash}p{2.5cm} >{\raggedright\arraybackslash}X}
\toprule
\textbf{Symbol} & \textbf{Meaning} \\
\midrule
\multicolumn{2}{l}{\textbf{A. Data and Model}} \\
\addlinespace[0.2em]
\(Y_t\)                              & Outcome at Time \(t\) \\
\(x_t\), \(u_t\)                     & Exposure Vector; Covariate Vector \\
\(x^\star\), \(u^\star\)             & Exposure and Covariate Reference Values \\
\(\mathcal L_m\), \(\mathcal L\)     & Exposure-Specific Lag Set; Collection of Lag Sets \\
\(x_t^{\mathcal L}\)                 & Lagged Exposure History \\
\(\mathcal H_t\)                     & Conditioning History \\
\(\mu_t\), \(\eta_t\)                & Conditional Mean; Linear Predictor \\
\(g\)                                & Invertible Link Function \\
\(n\), \(M\), \(K\)                  & Number of Observations; Number of Exposures; Number of Covariates \\
\addlinespace[0.35em]
\multicolumn{2}{l}{\textbf{B. Additive Components}} \\
\addlinespace[0.2em]
\(\alpha\)                           & Model Intercept; Reference-Value Linear Predictor \\
\(f_m\), \(h_k\)                     & Exposure--Lag Response Surface; Covariate Shape Function \\
\(F_m\)                              & Cumulative Exposure--Response Component \\
\(\tilde f_{ms}\), \(\tilde h_{ks}\) & Exposure--Lag Subnetwork; Covariate Subnetwork \\
\(S_m^x\), \(S_k^u\)                 & Numbers of Exposure and Covariate Subnetworks \\
\addlinespace[0.35em]
\multicolumn{2}{l}{\textbf{C. Parameters, Mixture, and Ensemble}} \\
\addlinespace[0.2em]
\(\theta\), \(\vartheta\)            & Complete Parameter Vector; Subnetwork Parameters \\
\(\omega\)                           & Mixing Weights \\
\(\lambda_w\)                        & Weight-Decay Coefficient \\
\(B\)                                & Number of Ensemble Members \\
\(\psi\)                             & Conditioned Parameters \\
\(\xi\)                              & Target-Specific Last-Layer Parameters \\
\addlinespace[0.35em]
\multicolumn{2}{l}{\textbf{D. Optimization}} \\
\addlinespace[0.2em]
\(E\)                                & Number of Training Epochs \\
\(\eta\), \(\eta_{\min}\)            & Learning Rate; Minimum Learning Rate \\
\addlinespace[0.35em]
\multicolumn{2}{l}{\textbf{E. Last-Layer Inference}} \\
\addlinespace[0.2em]
\(\zeta\)                            & Target Evaluation Point \\
\(\tau\)                             & Inferential Target \\
\(\Phi\)                             & Last-Layer Predictor Jacobian \\
\(W\)                                & GLM Working-Weight Matrix \\
\(\hat\varphi\)                      & Dispersion or Working-Dispersion Factor \\
\(H\)                                & Dispersion-Adjusted Last-Layer Information \\
\(\lambda\), \(P_\xi\)               & Gaussian Prior Precision; Penalization Selector \\
\(\widehat\Sigma_\xi\)               & Conditional Last-Layer Laplace Covariance \\
\(J_\zeta\)                          & Target Jacobian \\
\(\widehat{\mathcal V}\)             & Moment-Matched Ensemble Variance \\
\addlinespace[0.35em]
\multicolumn{2}{l}{\textbf{F. Simulation Evaluation}} \\
\addlinespace[0.2em]
\(\mathcal G\)                       & Evaluation Grid \\
\(R\)                                & Number of Monte Carlo Replicates \\
\bottomrule
\end{tabularx}
\caption{\textbf{Principal Notation.} Recurring notation in the main and supplementary methods; symbols local to individual derivations are defined where they occur.}
\label{tab:supp_notation}
\end{table}

\section{Last-Layer Laplace Confidence Intervals}
\label{supp:laplace}

Notation follows the main text. This section gives the full derivation of the target-specific last-layer Laplace construction, its propagation to the reported inferential targets, and its combination across ensemble members.

\subsection{Setup, Partition, and Scope}
\label{supp:laplace_setup}

For a given inferential target, partition the parameter vector as
\[
  \theta=(\psi,\xi),
  \qquad
  \xi\in\mathbb{R}^{q},
\]
where \(\xi\) contains the global intercept and the final-layer parameters of the corresponding component, while \(\psi\) contains all remaining parameters: the ExU input layers, hidden layers, other model components and, in the mixture formulation, the mixing weights \(\omega\). The conditional calculation fixes \(\psi\) at its fitted value \(\hat\psi\), leaving \(\xi\) free; \(\hat\xi\) denotes its fitted value and serves as the expansion center below.

The construction follows the principle of Laplace-approximated Neural Additive Models (LA-NAMs) \cite{bouchiat2024bayesian}: form a Gaussian approximation in parameter space and propagate it through the additive components to obtain function-space uncertainty. LA-NAM develops both linearized and last-layer variants for cross-sectional NAM feature networks; DLNAM uses the last-layer construction for distributed-lag components, conditioning on the learned representation and, through \(\hat\psi\), the fitted mixing weights.

The partition is chosen so that the conditional model becomes a generalized linear model in the last-layer parameters. Given \(\hat\psi\), each subnetwork contributing to the selected component maps its input to a fixed hidden-feature vector, while all components outside the selected target are fixed. The selected component is therefore a linear combination of fixed features, with the conditioned mixing weights acting only as fixed multipliers, and the predictor is affine in \(\xi\). Conditional curvature and function-space propagation consequently reduce to generalized linear model calculations with a Gaussian prior, avoiding high-dimensional full-network curvature and restricting the local Gaussian approximation to those last-layer parameters \cite{daxberger2021laplace,kristiadi2020being}.

Conditioning on the mixing weights is also required for an affine, identified last-layer parameterization. In the mixture formulation,
\[
  f_m^{\theta_m^x}(\cdot,\cdot)
  =
  \sum_{s=1}^{S_m^x}
  \omega_{ms}^x
  \tilde f_{ms}^{\vartheta_{ms}^x}(\cdot,\cdot),
\]
so a mixing weight and the corresponding subnetwork's final linear layer enter the predictor only through their product. Allowing both to vary within \(\xi\) would make the predictor bilinear and scale-non-identifiable: rescaling
\[
  \omega_{ms}^x\mapsto c\,\omega_{ms}^x
\]
while multiplying the corresponding final-layer parameters by \(c^{-1}\) leaves the fitted function, and hence the likelihood or working likelihood, unchanged. The curvature is therefore singular along this direction. Including \(\omega\) in \(\psi\) and conditioning on \(\hat\psi\) removes the indeterminacy at the cost of omitting within-member mixing-weight uncertainty; Section~\ref{supp:laplace_scope} returns to this limitation.

Write the vector of linear predictors across the \(n\) observations as
\[
  \eta(\xi)
  =
  \bigl(\eta_1(\xi),\ldots,\eta_n(\xi)\bigr)^\top,
\]
where
\[
  \eta_t(\xi)
  =
  \alpha
  +
  \sum_{m=1}^{M}\sum_{\ell\in\mathcal L_m}
  f_m^{\hat\psi,\xi}(x_{t-\ell,m},\ell)
  +
  \sum_{k=1}^{K}
  h_k^{\hat\psi,\xi}(u_{tk}),
\]
with \(\eta_t(\xi)=g\{\mu_t(\xi)\}\) and fitted means
\[
  \hat\mu_t
  =
  g^{-1}\{\eta_t(\hat\xi)\}.
\]
Define the last-layer Jacobian
\begin{equation}
  \Phi
  =
  \frac{\partial\eta(\xi)}{\partial\xi^\top}
  \in\mathbb{R}^{n\times q},
  \label{eq:supp_phi}
\end{equation}
which maps perturbations of the last-layer parameters to changes in the linear predictors and plays the same design-matrix role as the cross-basis in a DLNM. Because \(\eta(\xi)\) is affine in \(\xi\), \(\Phi\) is exact, constant, and independent of the point at which it is evaluated.

Throughout, \(\overset{c}{=}\) denotes equality, and \(\overset{c}{\approx}\) approximation, up to an additive constant that does not depend on the free variable of the expression: \(\xi\) in Section~\ref{supp:laplace_posterior} and \(\lambda\) in Section~\ref{supp:laplace_evidence}.

\subsection{Prior and Laplace Approximation}
\label{supp:laplace_posterior}

Partition the last-layer parameters as
\[
  \xi=(\xi_0,\xi_P),
\]
where \(\xi_0\) is the global intercept and \(\xi_P\) contains the remaining final-layer parameters. Because the intercept is left unpenalized, define
\[
  P_\xi
  =
  \operatorname{diag}(0,1,\ldots,1)
  \in\mathbb{R}^{q\times q}.
\]
The number of penalized parameters is therefore
\[
  q_P
  =
  \operatorname{tr}(P_\xi)
  =
  q-1,
  \qquad
  \xi_P\in\mathbb{R}^{q_P},
\]
and
\[
  \xi^\top P_\xi\xi
  =
  \|\xi_P\|_2^2.
\]
For the post-hoc Laplace construction, place an isotropic zero-mean Gaussian prior with precision \(\lambda\) on \(\xi_P\) and a flat, improper prior on \(\xi_0\),
\begin{equation}
  \xi_P\mid\lambda
  \sim
  \mathcal{N}_{q_P}\!\bigl(0,\lambda^{-1}I_{q_P}\bigr),
  \qquad
  p(\xi_0)\propto1.
  \label{eq:supp_prior}
\end{equation}
Thus
\[
  -\log p(\xi\mid\lambda)
  \;\overset{c}{=}\;
  \frac{\lambda}{2}\xi^\top P_\xi\xi
  -
  \frac{q_P}{2}\log\lambda .
\]
Leaving the intercept unpenalized lets the data determine the baseline response. Its improper normalization is independent of \(\lambda\) and therefore does not affect its selection in Section~\ref{supp:laplace_evidence}. The term \(-q_P\log(\lambda)/2\) affects neither the mode nor the curvature but must be retained in the evidence because it depends on \(\lambda\).

For a likelihood fit, write
\[
  \ell(\xi)
  =
  \log p(Y\mid\xi,\hat\psi)
\]
for the conditional log-likelihood. By Bayes' rule,
\[
  p(\xi\mid Y,\hat\psi,\lambda)
  \propto
  p(Y\mid\xi,\hat\psi)\,
  p(\xi\mid\lambda),
\]
so the negative log-posterior is
\begin{equation}
  \Psi(\xi)
  \;\overset{c}{=}\;
  -\ell(\xi)-\log p(\xi\mid\lambda).
  \label{eq:supp_psi}
\end{equation}
For a working-likelihood fit, the same expression is used with \(\ell\) interpreted as the corresponding conditional working criterion; the resulting construction is therefore a Gaussian working analogue rather than a literal posterior from a normalized likelihood.

The precision \(\hat\lambda\) is selected after network training by the evidence-style procedure of Section~\ref{supp:laplace_evidence}. The fitted \(\hat\xi\) is therefore retained rather than re-optimized under the Gaussian prior and is treated as an approximate posterior mode for likelihood fits, or as the corresponding working-mode analogue for working-likelihood fits. Thus,
\[
  \nabla_\xi\Psi(\hat\xi)\approx0,
\]
with accuracy determined by how closely the trained solution satisfies the post-hoc penalized score condition at \(\hat\lambda\). Unlike the alternating parameter--hyperparameter optimization of LA-NAM \cite{bouchiat2024bayesian}, the present construction holds \(\hat\xi\) and \(H\) fixed while selecting \(\lambda\), leaving the reported point estimate unchanged. Expanding about \(\hat\xi\),
\begin{equation}
  \Psi(\xi)
  =
  \Psi(\hat\xi)
  +
  \nabla_\xi\Psi(\hat\xi)^\top(\xi-\hat\xi)
  +
  \frac{1}{2}
  (\xi-\hat\xi)^\top
  \mathcal I_\xi
  (\xi-\hat\xi)
  +
  \mathcal O\bigl(\|\xi-\hat\xi\|^3\bigr),
  \qquad
  \mathcal I_\xi
  =
  \nabla_\xi^2\Psi(\hat\xi).
  \label{eq:supp_taylor}
\end{equation}
The remainder is of third order under the usual smoothness of the conditional likelihood or working criterion as a function of the linear predictor. Once \(\hat\psi\) is fixed and \(\eta\) is affine in \(\xi\), this condition places no further smoothness requirement on the network representation itself.

Under the approximate-mode assumption, the linear term is neglected and the quadratic expansion defines the member-level Gaussian Laplace approximation
\[
  \xi\mid Y,\hat\psi
  \ \approx\
  \mathcal N\!\bigl(\hat\xi,\widehat\Sigma_\xi\bigr),
  \qquad
  \widehat\Sigma_\xi
  =
  \mathcal I_\xi^{-1},
\]
for likelihood fits \cite{mackay1992practical,daxberger2021laplace}; for working-likelihood fits, the same expression is interpreted as its Gaussian working analogue.

\subsection{Last-Layer Curvature}
\label{supp:laplace_curvature}

Because the likelihood or working criterion depends on \(\xi\) only through \(\eta(\xi)\), the chain rule gives
\[
  \nabla_\xi(-\ell)
  =
  \Phi^\top\nabla_\eta(-\ell),
\]
and
\[
  \nabla_\xi^2(-\ell)
  =
  \Phi^\top
  \bigl[\nabla_\eta^2(-\ell)\bigr]
  \Phi
  +
  \sum_{t=1}^{n}
  \frac{\partial(-\ell)}{\partial\eta_t}
  \nabla_\xi^2\eta_t .
\]
Because \(\eta(\xi)\) is affine in \(\xi\),
\[
  \nabla_\xi^2\eta_t=0
\]
for every \(t\), so no predictor-curvature term enters the Hessian. The second term above therefore vanishes identically, and no Gauss--Newton approximation is required for the parameterization used here. Had \(\xi\) entered nonlinearly, as it would if the mixing weights were allowed to vary jointly with their corresponding output layers, this term would generally be nonzero; discarding it on the basis of its vanishing expectation would then constitute a Gauss--Newton or Fisher approximation \cite{immer2021linearization}.

For a generalized linear model \cite{mccullagh1989glm}, let
\[
  W
  =
  \operatorname{diag}(w_1,\ldots,w_n),
  \qquad
  w_t
  =
  \frac{(\mathrm d\mu_t/\mathrm d\eta_t)^2}{V(\mu_t)},
\]
with the weights evaluated at the fitted means and \(V\) denoting the variance function. To see how these weights enter the curvature, for an individual observation the likelihood or quasi-likelihood score with respect to the linear predictor has the form
\[
  \frac{\partial\ell_t}{\partial\eta_t}
  =
  \frac{Y_t-\mu_t}
       {\varphi V(\mu_t)}
  \frac{\mathrm d\mu_t}{\mathrm d\eta_t}.
\]
Differentiating once more gives a term
\[
  -\frac{1}{\varphi}
  \frac{(\mathrm d\mu_t/\mathrm d\eta_t)^2}
       {V(\mu_t)}
\]
together with terms proportional to \(Y_t-\mu_t\). Because
\[
  \mathbb E(Y_t-\mu_t\mid\mu_t)=0,
\]
the latter vanish in expectation, yielding
\[
  -\mathbb E\left[
    \frac{\partial^2\ell_t}{\partial\eta_t^2}
    \,\middle|\,
    \mu_t
  \right]
  =
  \frac{w_t}{\varphi},
\]
or, across observations,
\[
  -\mathbb E\bigl[
    \nabla_\eta^2\ell
    \mid\mu
  \bigr]
  =
  W/\varphi.
\]
Thus \(W/\varphi\) is the expected information on the linear-predictor scale, and \(\Phi\) pulls this information back to the last-layer parameterization. For likelihood fits with canonical links, the residual-dependent terms in the observed second derivative vanish and the observed and expected information coincide. For non-canonical likelihood fits, \(W/\varphi\) gives the corresponding Fisher information; for working-likelihood fits it gives the analogous working information.

Let \(\hat\varphi\) denote the dispersion or working-dispersion factor appropriate to the fitted outcome model, taking its family-specified value when fixed and being estimated where required. The dispersion-adjusted last-layer information is therefore
\begin{equation}
  H
  =
  \Phi^\top W\Phi/\hat\varphi .
  \label{eq:supp_H}
\end{equation}

When dispersion is fixed by the fitted family, its family-specified value is used. When it is estimated in the reported analyses, we use a Pearson-residual estimate. For each ensemble member \(b=1,\ldots,B\), with fitted means \(\hat\mu_{bt}\),
\begin{equation}
  \hat\varphi_b
  =
  \frac{1}{\nu}
  \sum_{t=1}^{n}
  \frac{(Y_t-\hat\mu_{bt})^2}
       {V(\hat\mu_{bt})},
  \qquad
  \hat\varphi
  =
  \max\left\{
    1,\,
    \frac{1}{B}\sum_{b=1}^{B}\hat\varphi_b
  \right\}.
  \label{eq:supp_dispersion}
\end{equation}
The lower bound \(\hat\varphi\ge1\) prevents apparent underdispersion from understating uncertainty. We use the pragmatic large-sample approximation \(\nu=n-1\). This does not treat the conditioned representation as parameter-free: the effective last-layer dimension \(\gamma\) derived below quantifies the penalized last-layer directions supported by the data, not the residual degrees of freedom of the complete trained network, and is therefore not substituted for \(\nu\). An exact full-network correction is unavailable, and the distinction is negligible at the sample sizes considered here.

Differentiating the Gaussian prior twice gives
\[
  \nabla_\xi[-\log p(\xi\mid\lambda)]
  =
  \lambda P_\xi\xi,
  \qquad
  \nabla_\xi^2[-\log p(\xi\mid\lambda)]
  =
  \lambda P_\xi.
\]
A Gaussian prior therefore contributes its precision additively to the conditional curvature. At the selected precision \(\hat\lambda\), and with a small numerical ridge \(\epsilon>0\),
\begin{equation}
  \mathcal I_\xi
  =
  H+\hat\lambda P_\xi+\epsilon I,
  \qquad
  \widehat\Sigma_\xi
  =
  \mathcal I_\xi^{-1}.
  \label{eq:supp_covariance}
\end{equation}
This is the conditional last-layer covariance given \(\hat\psi\), with \(\Phi\) playing the role of the design matrix.

In implementation, the computation is organized on the unscaled information as
\[
  \widehat\Sigma_\xi
  =
  \hat\varphi
  \bigl(
    \Phi^\top W\Phi
    +
    \hat\varphi\hat\lambda P_\xi
    +
    \epsilon I
  \bigr)^{-1},
\]
which is equivalent to Equation~\eqref{eq:supp_covariance} up to the scaling convention for the numerical ridge. Factoring out the dispersion rescales the prior term to \(\hat\varphi\hat\lambda P_\xi\) in this implementation form, while \(\epsilon\) serves only as a numerical safeguard. Although \(\Phi^\top W\Phi\) may be rank-deficient because constant columns induced by final-layer biases are collinear with the global intercept, the prior supplies curvature in penalized directions and the ridge additionally protects against numerical ill-conditioning, including in the unpenalized intercept direction.

\subsection{Evidence-Style Prior Precision}
\label{supp:laplace_evidence}

For likelihood fits, \(\lambda\) is selected by a post-hoc approximation to type-II maximum likelihood, or empirical Bayes. The target criterion is the evidence for \(\lambda\), with the last-layer parameters integrated out,
\[
  p(Y\mid\lambda)
  \;\propto_\lambda\;
  \int
  p(Y\mid\xi,\hat\psi)
  p(\xi\mid\lambda)
  \,\mathrm d\xi,
\]
rather than the prior density evaluated at \(\hat\xi\), whose maximization over \(\lambda\) would make no reference to the data. Here \(\propto_\lambda\) denotes equality up to a multiplicative constant independent of \(\lambda\), including the arbitrary normalization of the flat intercept prior.

In the implemented post-hoc calculation, the trained expansion center \(\hat\xi\) and working information \(H\) are held fixed as \(\lambda\) is updated. The procedure is therefore a MacKay evidence fixed-point approximation rather than literal maximization of a fully re-optimized marginal likelihood \cite{mackay1992interpolation,mackay1992practical}. For working-likelihood fits, where no normalized likelihood and hence no literal marginal likelihood is available, the same fixed-point calculation is applied to the working information and interpreted as the corresponding evidence-style analogue.

Using
\[
  \mathcal I_\xi(\lambda)
  =
  H+\lambda P_\xi,
\]
with the numerical ridge omitted for selection of \(\lambda\), the \(\lambda\)-dependent terms of the Laplace log evidence under the fixed-\(\hat\xi\), fixed-\(H\) approximation are
\[
  \log p(Y\mid\lambda)
  \;\overset{c}{\approx}\;
  \ell(\hat\xi)
  -
  \frac{\lambda}{2}\|\hat\xi_P\|_2^2
  +
  \frac{q_P}{2}\log\lambda
  -
  \frac{1}{2}
  \log|\mathcal I_\xi(\lambda)|.
\]
The three \(\lambda\)-dependent terms balance the Gaussian-prior normalization, the compatibility of the prior scale with the fitted final-layer parameters, and the log-determinant complexity term arising from the Gaussian integral.

Differentiating with respect to \(\lambda\) and using
\[
  \frac{\partial}{\partial\lambda}
  \log|\mathcal I_\xi(\lambda)|
  =
  \operatorname{tr}
  \bigl[
    \mathcal I_\xi(\lambda)^{-1}P_\xi
  \bigr]
\]
gives
\[
  -\frac{1}{2}\|\hat\xi_P\|_2^2
  +
  \frac{q_P}{2\lambda}
  -
  \frac{1}{2}
  \operatorname{tr}
  \bigl[
    \mathcal I_\xi(\lambda)^{-1}P_\xi
  \bigr]
  =
  0.
\]
The dependence of the expansion center and working information on \(\lambda\) is omitted by construction. Multiplying by \(2\lambda\) gives
\[
  \lambda\|\hat\xi_P\|_2^2
  =
  \gamma(\lambda),
\]
where
\begin{equation}
  \gamma(\lambda)
  =
  q_P
  -
  \lambda
  \operatorname{tr}
  \bigl[
    (H+\lambda P_\xi)^{-1}P_\xi
  \bigr]
  =
  \sum_{i=1}^{q_P}
  \frac{e_i}{e_i+\lambda}
  \in[0,q_P),
  \label{eq:supp_gamma}
\end{equation}
and hence the MacKay fixed point is
\begin{equation}
  \hat\lambda
  =
  \frac{\gamma(\hat\lambda)}
       {\|\hat\xi_P\|_2^2}.
  \label{eq:supp_fixedpoint}
\end{equation}

Here \(e_1,\ldots,e_{q_P}\) are the eigenvalues of the penalized information after profiling out the unpenalized intercept. Partitioning \(H\) into intercept (\(0\)) and penalized (\(P\)) components gives
\[
  H+\lambda P_\xi
  =
  \begin{pmatrix}
    H_{00} & H_{0P}\\
    H_{P0} & H_{PP}+\lambda I
  \end{pmatrix}.
\]
The raw block \(H_{PP}\) does not by itself represent the information available for the penalized parameters when the intercept is allowed to adjust, because changes in the penalized coefficients may be partially offset by changes in the intercept. Profiling out the intercept removes this shared direction, leaving the Schur complement
\begin{equation}
  H_P
  =
  H_{PP}
  -
  H_{P0}H_{00}^{-1}H_{0P}.
  \label{eq:supp_schur}
\end{equation}
Thus \(H_P\) is the information for the penalized parameters after accounting for their dependence on the unpenalized intercept. By the block-inverse identity, the penalized block of \((H+\lambda P_\xi)^{-1}\) is
\[
  (H_P+\lambda I)^{-1},
\]
and consequently
\[
  \operatorname{tr}
  \bigl[
    (H+\lambda P_\xi)^{-1}P_\xi
  \bigr]
  =
  \operatorname{tr}
  \bigl[
    (H_P+\lambda I)^{-1}
  \bigr].
\]
The raw submatrix \(H_{PP}\) would suffice only if the intercept were orthogonal to the penalized directions, which reference centering does not ensure.

Equation~\eqref{eq:supp_gamma} identifies \(\gamma\) as the effective number of penalized directions constrained by the data: each eigendirection contributes \(e_i/(e_i+\lambda)\), approaching one when the data dominate the prior and zero when the prior dominates. If the log-determinant term were omitted, the fixed point would reduce to
\[
  \lambda
  =
  q_P/\|\hat\xi_P\|_2^2,
\]
treating all \(q_P\) penalized directions as constrained. Retaining the determinant replaces \(q_P\) by the effective count \(\gamma\). Equivalently,
\[
  \hat\lambda^{-1}
  =
  \frac{\|\hat\xi_P\|_2^2}{\gamma}
\]
estimates the prior variance as the mean squared fitted weight per effective penalized direction, structurally analogous to degrees-of-freedom corrections in linear regression and to marginal-likelihood selection of smoothing parameters in penalized spline models.

Under the fixed expansion-center and fixed-information approximation, the \(e_i\) are computed once from \(H_P\) and do not vary with \(\lambda\). To characterize the solution, define
\[
  F(\lambda)
  =
  \lambda\|\hat\xi_P\|_2^2
  -
  \gamma(\lambda).
\]
In the nondegenerate case with \(\|\hat\xi_P\|_2>0\) and information in at least one penalized direction,
\[
  F'(\lambda)
  =
  \|\hat\xi_P\|_2^2
  +
  \sum_{i=1}^{q_P}
  \frac{e_i}{(e_i+\lambda)^2}
  >0,
\]
while \(F(\lambda)\) is negative as \(\lambda\downarrow0\) and diverges to \(+\infty\) as \(\lambda\to\infty\). The stationarity equation therefore has a unique positive solution. Equation~\eqref{eq:supp_fixedpoint} is solved by the iteration
\[
  \lambda^{(r+1)}
  =
  \frac{\gamma(\lambda^{(r)})}
       {\|\hat\xi_P\|_2^2},
  \qquad
  r=0,1,\ldots,
\]
from an initial \(\lambda^{(0)}>0\) until numerical convergence. For an ensemble, \(\gamma\) and \(\|\hat\xi_P\|_2^2\) are summed across members before the update, yielding one shared pooled fixed-point rule across representations fitted to the same data. This pooling is not a product of independent marginal likelihood contributions.

\subsection{Inferential Targets and Propagation}
\label{supp:laplace_delta}

The inferential targets are the centered components of Equation~\eqref{eq:dlnam_predictor_general}: the exposure--lag surface \(f_m^{\hat\psi,\xi}\), its cumulative
\[
  F_m^{\hat\psi,\xi}(\cdot)
  =
  \sum_{\ell\in\mathcal L_m}
  f_m^{\hat\psi,\xi}(\cdot,\ell),
\]
and the covariate shape function \(h_k^{\hat\psi,\xi}\). Under the reference-centering convention,
\[
  f_m^{\hat\psi,\xi}(x_m^\star,\ell)=0
  \quad\text{for all }\ell\in\mathcal L_m,
  \qquad
  h_k^{\hat\psi,\xi}(u_k^\star)=0,
\]
so these functions are contrasts on the link scale.

For
\[
  \tau^{\hat\psi,\xi}
  \in
  \left\{
    f_m^{\hat\psi,\xi},
    F_m^{\hat\psi,\xi},
    h_k^{\hat\psi,\xi}
  \right\},
\]
let \(\tau_{\mathcal G}^{\hat\psi,\xi}\) denote its evaluations on a finite grid \(\mathcal G\), and define
\[
  J_{\mathcal G}
  =
  \frac{\partial
  \tau_{\mathcal G}^{\hat\psi,\xi}}
  {\partial\xi^\top},
  \qquad
  J_\zeta
  =
  \frac{\partial
  \tau^{\hat\psi,\xi}(\zeta)}
  {\partial\xi^\top},
  \quad
  \zeta\in\mathcal G.
\]
Conditional on \(\hat\psi\), every target is affine in \(\xi\), so
\[
  \tau_{\mathcal G}^{\hat\psi,\xi}
  =
  \tau_{\mathcal G}^{\hat\psi,\hat\xi}
  +
  J_{\mathcal G}(\xi-\hat\xi)
\]
holds exactly and \(J_{\mathcal G}\) is constant. An affine transformation of a Gaussian is Gaussian; hence
\begin{equation}
  \widehat\Sigma_{\mathcal G}
  =
  J_{\mathcal G}
  \widehat\Sigma_\xi
  J_{\mathcal G}^\top,
  \qquad
  \operatorname{sd}
  \{\tau^{\hat\psi,\xi}(\zeta)\}
  =
  \bigl\{
    J_\zeta
    \widehat\Sigma_\xi
    J_\zeta^\top
  \bigr\}^{1/2},
  \quad
  \zeta\in\mathcal G.
  \label{eq:supp_delta}
\end{equation}
Thus covariance propagation is exact conditional on \(\hat\psi\), with no delta-method linearization error. Because the inferential targets are reference-centered contrasts, the intercept column of \(J_{\mathcal G}\) is zero, although the intercept remains in \(\mathcal I_\xi\).

Evaluating the conditional target at the fitted expansion center gives
\(\hat\tau^{\hat\psi,\hat\xi}(\zeta)
\equiv
\left.\tau^{\hat\psi,\xi}(\zeta)\right|_{\xi=\hat\xi}\).
The member-level pointwise \(1-\alpha\) interval on the link-contrast scale is therefore
\[
  \hat\tau^{\hat\psi,\hat\xi}(\zeta)
  \pm
  z_{1-\alpha/2}
  \operatorname{sd}
  \{\tau^{\hat\psi,\xi}(\zeta)\}.
\]
Where the reported estimand is a monotone transformation of the link-scale contrast, the interval endpoints are transformed accordingly. When an absolute linear predictor is reported on the mean scale, the transformation is \(g^{-1}\). In the reported analyses, exponentiation therefore maps log-link contrasts to relative risks and logit-link contrasts to odds ratios, while the identity link requires no transformation.

\subsection{Ensembles}
\label{supp:laplace_ensemble}

For an ensemble of \(B\) independently initialized members \cite{lakshminarayanan2017ensembles}, write
\[
  \hat\psi
  =
  (\hat\psi_1,\ldots,\hat\psi_B),
  \qquad
  \hat\xi
  =
  (\hat\xi_1,\ldots,\hat\xi_B)
\]
for the collections of fitted member-specific conditioned parameters and last-layer expansion centers, respectively. Applying the ensemble construction of Equation~\eqref{eq:dlnam_ensemble} to a reported inferential target gives
\begin{equation}
  \hat\tau^{\hat\psi,\hat\xi}(\zeta)
  =
  \frac{1}{B}\sum_{b=1}^{B}
  \hat\tau_b^{\hat\psi_b,\hat\xi_b}(\zeta),
  \qquad
  \widehat{\mathcal V}(\zeta)
  =
  \underbrace{\frac{1}{B}\sum_{b=1}^{B}
  J_{b\zeta}\widehat\Sigma_{\xi_b}J_{b\zeta}^{\top}}_{\text{within-member}}
  +
  \underbrace{\frac{1}{B}\sum_{b=1}^{B}
  \left\{
  \hat\tau_b^{\hat\psi_b,\hat\xi_b}(\zeta)
  -
  \hat\tau^{\hat\psi,\hat\xi}(\zeta)
  \right\}^{2}}_{\text{between-member}} .
  \label{eq:supp_ensemble}
\end{equation}

Combining conditional last-layer Laplace uncertainty with between-member variation, the reported pointwise \(1-\alpha\) confidence interval is
\[
  \hat\tau^{\hat\psi,\hat\xi}(\zeta)
  \pm
  z_{1-\alpha/2}
  \widehat{\mathcal V}(\zeta)^{1/2},
\]
with the same transformation to the reported scale as above.

The first term in Equation~\eqref{eq:supp_ensemble} averages member-specific conditional last-layer variances without further division by \(B\). Dividing it by \(B\) would instead estimate the variance of an arithmetic mean under an independence interpretation and would mechanically narrow the interval as more initializations were fitted, whereas the present construction represents an equally weighted mixture of member-level conditional approximations. The second term records between-member variation; its \(1/B\) denominator is the variance convention of that finite mixture, not an unbiased estimate of a hypothetical population variance over random initializations.

Consequently, \(\hat\tau^{\hat\psi,\hat\xi}(\zeta)\) and \(\widehat{\mathcal V}(\zeta)\) are the mean and variance of the equally weighted mixture of member-level Gaussian Laplace approximations. The reported interval is a Gaussian moment-matched summary of this generally non-Gaussian mixture. Because the members are fitted to the same data, \(\widehat{\mathcal V}(\zeta)\) is a constructed uncertainty measure, not the sampling variance of the ensemble mean or full posterior uncertainty in the learned representation. Its between-member term captures only fit instability expressed through disagreement among independently initialized representations and last-layer centers.

\subsection{Regularity Conditions and Scope}
\label{supp:laplace_scope}

The procedure as a whole is approximate, but conditional on \(\hat\psi\) several parts of the calculation are exact. Because the predictor is affine in \(\xi\), the Jacobian \(\Phi\) is exact and constant and the predictor-curvature term in the Hessian vanishes identically, so no Gauss--Newton approximation is required. The inferential targets are likewise affine in \(\xi\), making the propagation in Equation~\eqref{eq:supp_delta} exact with no delta-method linearization error. For likelihood fits with canonical links, the likelihood contribution to the last-layer curvature is also exact.

The remaining assumptions and approximations are: (i) the trained \(\hat\xi\), which is not re-optimized after selection of \(\hat\lambda\), is treated as an approximate posterior mode for likelihood fits, or as the corresponding working-mode analogue for working-likelihood fits, so that \(\nabla_\xi\Psi(\hat\xi)\approx0\); (ii) \(\mathcal I_\xi\) is positive definite, secured numerically by the prior together with the ridge; (iii) the conditional likelihood or working criterion is three times continuously differentiable in the linear predictor over the relevant domain, ensuring the third-order expansion in Equation~\eqref{eq:supp_taylor}; (iv) for non-canonical likelihood fits, the Fisher information adequately represents the observed information; (v) for working-likelihood fits, the dispersion-scaled working information provides an adequate curvature surrogate; and (vi) at the ensemble level, the equally weighted member mixture is adequately summarized by the Gaussian interval obtained by matching its mean and variance.

The member-level Laplace approximations are conditional on \(\hat\psi_b\), including the fitted mixing weights in the mixture formulation, and therefore do not constitute a posterior over the complete parameter vector. Relaxing this conditioning would require the linearized Laplace variant, which linearizes the complete network about the trained parameters and so propagates uncertainty in \(\hat\psi\) as well \cite{immer2021linearization,bouchiat2024bayesian}. For distributed-lag components this would require Jacobians with respect to all network parameters at every evaluated exposure--lag point, and a correspondingly larger and denser information matrix; the last-layer construction is retained here for that reason. The reported ensemble confidence intervals additionally incorporate between-member variation, which captures fit instability not represented by the member-level conditional variance, including that arising from differences in learned representations and fitted last-layer centers, but only insofar as it appears as disagreement among independently initialized fits of the same data. Their frequentist adequacy is consequently an empirical question and is evaluated by coverage against the generating functions in the simulation studies.

The reported intervals are pointwise confidence intervals for the inferential targets \(f_m\), \(F_m\), and \(h_k\), not simultaneous confidence bands; statements about an entire curve or surface would require an additional multiplicity adjustment. They quantify uncertainty in the fitted response functions rather than prediction of future outcomes. For a fully specified outcome distribution, prediction intervals could instead be obtained from quantiles of the predictive distribution, additionally integrating over uncertainty in the fitted linear predictor. A mean--variance specification alone, as in quasi-likelihood models, does not determine predictive quantiles and would require an additional distributional assumption.

\section{Data-Generating Surfaces}
\label{supp:dgp}

The outcome model, exposure process, reference value, and evaluation grid are defined in Section~\ref{sec:mc_design}. Let
\[
  \sigma(z)
  =
  \{1+\exp(-z)\}^{-1},
  \qquad
  \rho_{c,\delta}(z)
  =
  \exp\left\{
    -\frac{1}{2}
    \left(\frac{z-c}{\delta}\right)^2
  \right\},
\]
denote the logistic sigmoid and Gaussian bump, respectively, and define the normalized lag kernels
\[
  D_\kappa(\ell)
  =
  \frac{\exp(-\ell/\kappa)}
       {\sum_{\ell'=0}^{L}\exp(-\ell'/\kappa)},
  \qquad
  Q_{c,\delta}(\ell)
  =
  \frac{\rho_{c,\delta}(\ell)}
       {\sum_{\ell'=0}^{L}\rho_{c,\delta}(\ell')}.
\]
The four data-generating surfaces were
\[
\begin{aligned}
  f_1(x,\ell)
  &=
  \bigl\{
    0.50\,\sigma((15-x)/2.5)
    +
    0.60\,\sigma((x-25)/1.5)
  \bigr\}
  D_3(\ell),
  \\
  f_2(x,\ell)
  &=
  0.70\,\sigma(x-30)\,Q_{1,1.5}(\ell)
  +
  \sigma((12-x)/3)
  \bigl\{
    0.70\,Q_{9,2}(\ell)
    -
    0.35\,Q_{5,1.5}(\ell)
  \bigr\},
  \\
  f_3(x,\ell)
  &=
  0.70\,\rho_{32,2}(x)\,Q_{1,1.5}(\ell)
  +
  0.40\,\sigma((10-x)/3)\,D_4(\ell),
  \\
  f_4(x,\ell)
  &=
  0.70\,\sigma(x-28)\,Q_{1,1}(\ell)
  +
  0.60\,\sigma(12-x)\,Q_{9,1.5}(\ell).
\end{aligned}
\]

As throughout, the generating surfaces are reported and scored after centering at \(x^\star=20\).

\begin{figure}[!ht]
  \centering
  \includegraphics[width=\linewidth]{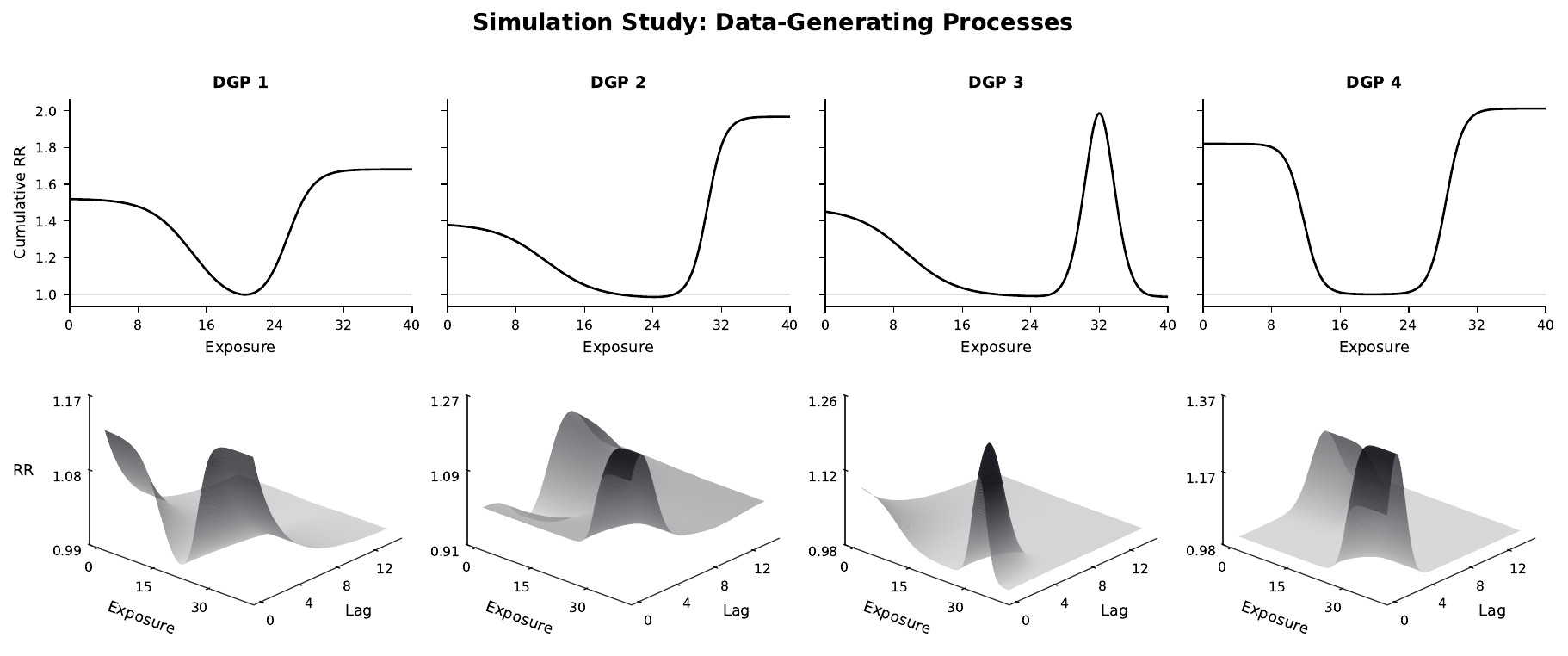}
  \caption{\textbf{Simulation Study: Data-Generating Processes.} Cumulative relative-risk functions and corresponding exposure--lag relative-risk surfaces for the four DGPs.}
  \label{fig:supp_dgp}
\end{figure}

DGP 1 is separable: a smooth exposure response is multiplied by a single exponentially decaying lag kernel. DGP 2 combines a sharp high-exposure response acting at early lags with a low-exposure response acting later and an opposing-sign contribution at intermediate lags. DGP 3 combines a narrow high-exposure response with a broader low-exposure arm on a slower lag kernel. DGP 4 places sharp responses at both extremes of the exposure range, with different dominant lag timing. Figure~\ref{fig:supp_dgp} shows the resulting exposure--lag surfaces and cumulative curves, the latter being the primary estimand of the simulation.

\section{Monte Carlo Uncertainty}
\label{supp:mcse}

Monte Carlo uncertainty is calculated from replicate-level summaries, thereby accounting for dependence among evaluation-grid points within each replicate rather than treating them as independent Monte Carlo replicates. For any reported summary that is an average of replicate-wise quantities,
\[
  \widehat S
  =
  \bar S
  =
  \frac{1}{R}\sum_{r=1}^{R}S_r,
\]
the Monte Carlo standard error is
\[
  \operatorname{MCSE}(\widehat S)
  =
  \frac{1}{\sqrt{R}}
  \sqrt{
    \frac{1}{R-1}
    \sum_{r=1}^{R}
    (S_r-\bar S)^2
  }.
\]
The plotted \(95\%\) Monte Carlo confidence interval is then
\[
  \widehat S
  \pm
  1.96\,\operatorname{MCSE}(\widehat S).
\]

For pointwise coverage of target \(\tau\) over evaluation region
\(\mathcal A_\tau\), define the replicate-level coverage summary
\[
  C_r^\tau(\mathcal A_\tau)
  =
  \frac{1}{|\mathcal A_\tau|}
  \sum_{\zeta\in\mathcal A_\tau}
  \mathbf 1
  \bigl\{
    \tau(\zeta)\in I_r^\tau(\zeta)
  \bigr\}.
\]
Then
\[
  \operatorname{Cov}_\tau(\mathcal A_\tau)
  =
  \frac{1}{R}
  \sum_{r=1}^{R}
  C_r^\tau(\mathcal A_\tau),
\]
and its MCSE follows directly from the replicate-level formula above. This construction preserves arbitrary dependence among coverage indicators across evaluation points within the same fitted replicate.

For RMSE, let
\[
  M_r^\tau(\mathcal A_\tau)
  =
  \frac{1}{|\mathcal A_\tau|}
  \sum_{\zeta\in\mathcal A_\tau}
  \{d_r^\tau(\zeta)\}^2
\]
denote the replicate-wise mean squared error and let
\[
  \bar M^\tau(\mathcal A_\tau)
  =
  \frac{1}{R}
  \sum_{r=1}^{R}
  M_r^\tau(\mathcal A_\tau),
\]
so that
\[
  \widehat{\operatorname{RMSE}}_\tau(\mathcal A_\tau)
  =
  \sqrt{\bar M^\tau(\mathcal A_\tau)}.
\]
The replicate-level formula gives
\[
  \operatorname{MCSE}
  \left\{
    \bar M^\tau(\mathcal A_\tau)
  \right\}
  =
  \frac{1}{\sqrt{R}}
  \sqrt{
    \frac{1}{R-1}
    \sum_{r=1}^{R}
    \left\{
      M_r^\tau(\mathcal A_\tau)
      -
      \bar M^\tau(\mathcal A_\tau)
    \right\}^{2}
  },
\]
and first-order delta-method propagation through the square-root transformation gives
\[
  \operatorname{MCSE}
  \left\{
    \widehat{\operatorname{RMSE}}_\tau(\mathcal A_\tau)
  \right\}
  \approx
  \frac{
    \operatorname{MCSE}
    \left\{
      \bar M^\tau(\mathcal A_\tau)
    \right\}
  }{
    2\sqrt{\bar M^\tau(\mathcal A_\tau)}
  }.
\]

For the joint-to-single degradation ratio
\[
  \widehat D
  =
  \frac{
    \widehat{\operatorname{RMSE}}_{\mathrm{joint}}
  }{
    \widehat{\operatorname{RMSE}}_{\mathrm{single}}
  },
\]
the implemented first-order propagation sets the covariance between the two Monte Carlo RMSE estimates to zero, giving
\[
  \operatorname{MCSE}(\widehat D)
  \approx
  \widehat D
  \sqrt{
    \left\{
      \frac{
        \operatorname{MCSE}
        (\widehat{\operatorname{RMSE}}_{\mathrm{joint}})
      }{
        \widehat{\operatorname{RMSE}}_{\mathrm{joint}}
      }
    \right\}^{2}
    +
    \left\{
      \frac{
        \operatorname{MCSE}
        (\widehat{\operatorname{RMSE}}_{\mathrm{single}})
      }{
        \widehat{\operatorname{RMSE}}_{\mathrm{single}}
      }
    \right\}^{2}
  }.
\]
No covariance term is included in this ratio approximation.

Regional squared bias and variance depend on the complete collection of replicates through \(\bar d^\tau(\zeta)\), rather than being simple averages of fixed replicate-wise summaries. Their Monte Carlo uncertainty is therefore estimated by delete-one-replicate jackknifing. For
\[
  T
  \in
  \left\{
    \operatorname{Bias}_\tau^2(\mathcal A_\tau),
    \operatorname{Var}_\tau(\mathcal A_\tau)
  \right\},
\]
let \(T_{(-r)}\) denote the estimate obtained after omitting replicate \(r\), and define
\[
  \bar T_{(-\cdot)}
  =
  \frac{1}{R}
  \sum_{r=1}^{R}
  T_{(-r)}.
\]
The jackknife Monte Carlo standard error is
\[
  \operatorname{MCSE}_{\mathrm{JK}}(T)
  =
  \sqrt{
    \frac{R-1}{R}
    \sum_{r=1}^{R}
    \left\{
      T_{(-r)}
      -
      \bar T_{(-\cdot)}
    \right\}^{2}
  }.
\]
These calculations are applied separately to the cumulative and full-surface targets, \(\tau\in\{F,f\}\), and to each reported evaluation region \(\mathcal A_\tau\).

\section{Mixture Subnetwork Decomposition}
\label{supp:mixture}

\begin{figure}[!ht]
  \centering
  \includegraphics[width=\linewidth]{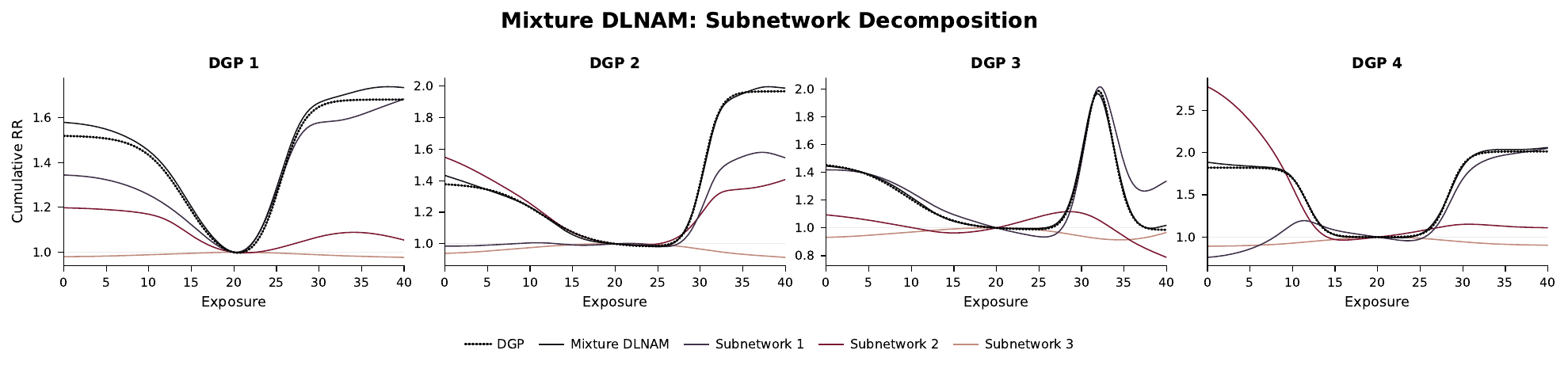}
  \caption{\textbf{Mixture DLNAM: Subnetwork Decomposition.} Cumulative relative-risk functions for the fitted Mixture DLNAM and its three weighted subnetwork contributions.}
  \label{fig:supp_mixture}
\end{figure}

Each Mixture DLNAM component is a learned linear combination of \(S\) separately initialized, jointly trained subnetworks,
\[
  f_m(\cdot,\cdot)
  =
  \sum_{s=1}^{S}
  \omega_{ms}\,
  \tilde f_{ms}(\cdot,\cdot).
\]
The ablation of Section~\ref{sec:results} removes this mixture construction and returns to the standard single-network formulation of Equation~\eqref{eq:dlnam_predictor_general}; although recovery degrades, the ablation does not show how the subnetworks combine within a fitted component. Figure~\ref{fig:supp_mixture} provides that decomposition for one independently initialized ensemble member fitted to one replicate of each DGP, without ensemble averaging.

The decomposition indicates that the mixture distributes functional complexity across weighted subnetwork contributions rather than requiring any single subnetwork to reproduce the complete exposure--response function. Individual contributions can emphasize different features of the response and combine through reinforcement or partial cancellation on the log-relative-risk scale. Together with the ablation results, this supports a representational role for the mixture construction beyond simple parameter expansion. The decomposition is illustrative rather than representative and does not imply that particular subnetwork roles are identifiable or stable across fits.

\section{Multivariate ExU Input Layers}
\label{supp:exu}

The scalar ExU of Equation~\eqref{eq:dlnam_exu} maps one input coordinate to \(d_j\) units, so components with \(p>1\) inputs require a multivariate extension; for a distributed-lag surface, \(z=(x,\ell)^\top\) with \(p=2\). We compare the reported coordinate-wise concatenation of Equation~\eqref{eq:dlnam_concat} with two unified alternatives that allow coordinate mixing within the input layer. Throughout, \(\phi\) acts elementwise, \(\odot\) denotes the Hadamard product, \(\mathbf 1_k\) is the all-ones vector of length \(k\), and exponentiation of vector- or matrix-valued quantities is elementwise; \(e^W\) does not denote a matrix exponential.

Components specified with linear inputs instead use the standard input layer, which mixes all coordinates in a single affine map,
\[
  h^{\mathrm{Lin}}(z)
  =
  \phi(Wz+b),
  \qquad
  W\in\mathbb{R}^{d\times p},\quad b\in\mathbb{R}^{d},
\]
with \(dp+d\) input-layer parameters and no unit-specific localization. It is not among the layers compared below, which concern the multivariate ExU parameterization used for exposure--lag surfaces.

The concatenation layer applies an independent scalar ExU to each coordinate and stacks the resulting outputs,
\[
  h^{\mathrm{C}}(z)
  =
  \bigl(
    h_1(z_1)^\top,\ldots,h_p(z_p)^\top
  \bigr)^\top
  \in\mathbb{R}^{d},
  \qquad
  h_j(z_j)
  =
  \phi\bigl(
    e^{w_j}\odot
    (z_j\mathbf 1_{d_j}-b_j)
  \bigr),
\]
where \(w_j,b_j\in\mathbb{R}^{d_j}\) and
\(d=\sum_{j=1}^{p}d_j\). For an exposure--lag surface, the width is split as
\(d_x=\lceil d/2\rceil\) and \(d_\ell=\lfloor d/2\rfloor\). Each first-layer unit therefore depends on only one coordinate, with exposure--lag interaction formed in subsequent hidden layers.

The two unified layers instead allow every first-layer unit to depend on all coordinates. With log-weights \(W\in\mathbb{R}^{p\times d}\), the unified local-bias layer assigns a separate location to each coordinate--unit pair through \(B\in\mathbb{R}^{p\times d}\),
\[
  h^{\mathrm{L}}(z)
  =
  \phi\Bigl(
    \bigl[
      e^W\odot
      (z\mathbf 1_d^\top-B)
    \bigr]^\top
    \mathbf 1_p
  \Bigr)
  \in\mathbb{R}^{d}.
\]
This retains the unit-specific localization of the scalar ExU and reduces exactly to Equation~\eqref{eq:dlnam_exu} when \(p=1\). The unified shared-bias layer instead uses one location per coordinate, \(b\in\mathbb{R}^{p}\), shared across units,
\[
  h^{\mathrm{S}}(z)
  =
  \phi\Bigl(
    (e^W)^\top(z-b)
  \Bigr)
  \in\mathbb{R}^{d}.
\]
When \(p=1\), all units share a single location and the scalar ExU is not recovered; the construction is included as a simpler unified multivariate parameterization with one shared location per coordinate.

The three layers are compared at matched output width \(d\), not matched parameter count. Their input-layer weights and biases number \(2d\), \(2pd\), and \(pd+p\) for concatenation, unified local-bias, and unified shared-bias, respectively. Input scaling and initialization settings otherwise match the reported configuration of Section~\ref{sec:estimators_fitting}.

\begin{figure}[!tbp]
  \centering
  \includegraphics[width=\linewidth]{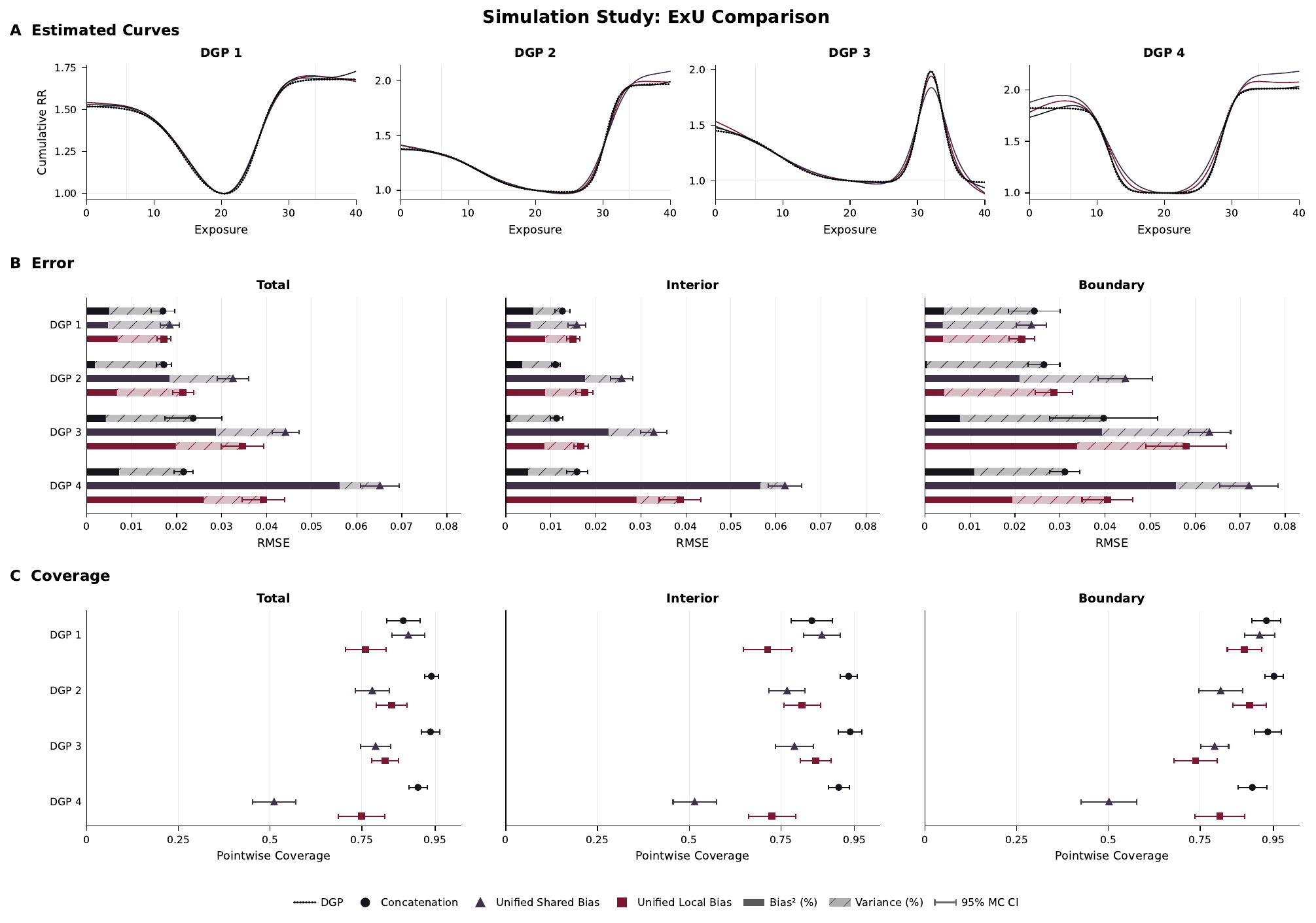}
  \caption{\textbf{Simulation Study: ExU Comparison.} Monte Carlo simulation of four DGPs, with \(R=50\) replicates per DGP. (\textbf{A}) Estimated cumulative relative-risk functions and data-generating functions, shown as pointwise Monte Carlo means across fitted replicates. (\textbf{B}) Cumulative log-relative-risk RMSE by method and exposure region, partitioned into squared bias and variance. (\textbf{C}) Empirical coverage of pointwise \(95\%\) intervals. Error bars denote 95\% Monte Carlo confidence intervals. Boundary denotes exposure values below the 5th or above the 95th percentile of the reference exposure distribution; interior denotes its complement. Associated RMSE and coverage results are reported in Table~\ref{tab:supp_exu}.}
  \label{fig:supp_exu}
\end{figure}

\begin{figure}[!tbp]
  \centering
  \includegraphics[width=\linewidth]{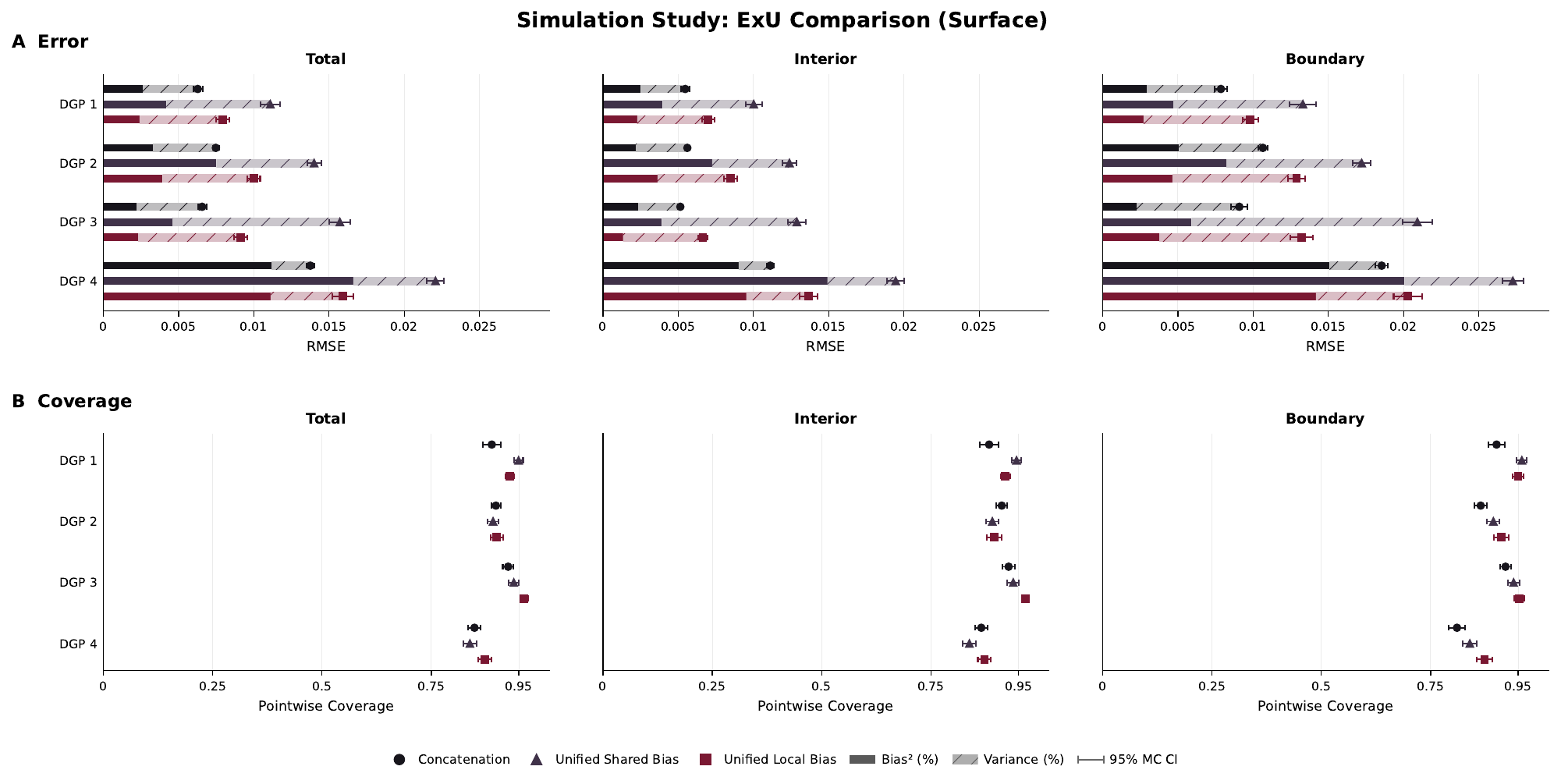}
  \caption{\textbf{Simulation Study: ExU Comparison (Surface).} Monte Carlo simulation of four DGPs, with \(R=50\) replicates per DGP. (\textbf{A}) Log-relative-risk RMSE by method and exposure region, partitioned into squared bias and variance. (\textbf{B}) Empirical coverage of pointwise \(95\%\) intervals. Error bars denote 95\% Monte Carlo confidence intervals. Boundary denotes exposure values below the 5th or above the 95th percentile of the reference exposure distribution; interior denotes its complement. All lags are included within each exposure region. Associated RMSE and coverage results are reported in Table~\ref{tab:supp_exu_surf}.}
  \label{fig:supp_exu_surf}
\end{figure}

Across DGPs 1--4 at \(R=50\) replicates, the reported concatenation layer gives the lowest error in every DGP for both the cumulative and full-surface targets (Figs.~\ref{fig:supp_exu} and~\ref{fig:supp_exu_surf}; Tables~\ref{tab:supp_exu} and~\ref{tab:supp_exu_surf}). The unified shared-bias layer, which removes unit-specific localization by sharing one location per coordinate across units, performs worst, giving the highest error on both the cumulative and full-surface targets and the lowest cumulative coverage. The unified local-bias layer retains unit-specific localization while allowing coordinate mixing in the input layer and gives intermediate cumulative error. These results indicate that the subsequent hidden layers are sufficient to represent the exposure--lag interactions present in the tested DGPs, while introducing multivariate coordinate mixing within the ExU input layer provides no recovery advantage. The comparison therefore supports the coordinate-wise concatenation used in the reported DLNAM against these two unified alternatives, without implying that it is universally optimal.

\section{Prospective Extensions}
\label{supp:extensions}

\subsection{Multitask DLNAM}

Let \(Y_t\in\mathcal Y^O\) denote the outcomes and define \(\mu_t=\mathbb E(Y_t\mid\mathcal H_t)\). For an invertible componentwise link function \(g\), the Multitask DLNAM predictor is
\begin{equation}
  g_o(\mu_{to})
  =
  \alpha_o
  +
  \sum_{m=1}^{M}\sum_{\ell\in\mathcal L_m}
  f_m^{\theta_{mo}^x}(x_{t-\ell,m},\ell)
  +
  \sum_{k=1}^{K}h_k^{\theta_{ko}^u}(u_{tk}),
  \label{eq:supp_multitask_predictor}
\end{equation}
where the outcome-specific components are
\begin{equation}
  f_m^{\theta_{mo}^x}(\cdot,\cdot)
  =
  \sum_{s=1}^{S_m^x}\omega_{mso}^{x}
  \tilde f_{ms}^{\vartheta_{ms}^x}(\cdot,\cdot),
  \qquad
  h_k^{\theta_{ko}^u}(\cdot)
  =
  \sum_{s=1}^{S_k^u}\omega_{kso}^{u}
  \tilde h_{ks}^{\vartheta_{ks}^u}(\cdot).
  \label{eq:supp_multitask_mixture}
\end{equation}
The subnetwork parameters \(\vartheta\) are shared, while the intercepts \(\alpha_o\) and mixing weights \(\omega_o=(\omega_o^x,\omega_o^u)\) are outcome-specific. This is the distributed-lag analogue of the multitask NAM construction \cite{agarwal2021neural}. For one outcome, Equations~\eqref{eq:supp_multitask_predictor} and~\eqref{eq:supp_multitask_mixture} reduce to the Mixture DLNAM. The shared representation induces cross-outcome regularization while limiting outcome-specific component flexibility to the mixing weights; it does not itself specify residual dependence.

\subsection{Hierarchical DLNAM}

For grouped outcomes \(Y_{tc}\in\mathcal Y\), define \(\mu_{tc}=\mathbb E(Y_{tc}\mid\mathcal H_t)\). For an invertible link function \(g\), the Hierarchical DLNAM predictor is
\begin{equation}
  g(\mu_{tc})
  =
  \alpha+\alpha_c
  +
  \sum_{m=1}^{M}\sum_{\ell\in\mathcal L_m}
  \bigl\{f_m^{\theta_m^x}(x_{t-\ell,m},\ell)
  +f_{mc}^{\theta_{mc}^x}(x_{t-\ell,m},\ell)\bigr\}
  +
  \sum_{k=1}^{K}\bigl\{h_k^{\theta_k^u}(u_{tk})
  +h_{kc}^{\theta_{kc}^u}(u_{tk})\bigr\},
  \label{eq:supp_hierarchical_predictor}
\end{equation}
where the first component in each brace is global and follows Equation~\eqref{eq:dlnam_mixture}, while the group-specific deviations are
\begin{equation}
  f_{mc}^{\theta_{mc}^x}(\cdot,\cdot)
  =
  \sum_{s=1}^{S_m^x}\omega_{msc}^{x}
  \tilde f_{msc}^{\vartheta_{msc}^x}(\cdot,\cdot),
  \qquad
  h_{kc}^{\theta_{kc}^u}(\cdot)
  =
  \sum_{s=1}^{S_k^u}\omega_{ksc}^{u}
  \tilde h_{ksc}^{\vartheta_{ksc}^u}(\cdot).
  \label{eq:supp_hierarchical_deviations}
\end{equation}
Partial pooling is induced by
\begin{equation}
  \omega_{msc}^{x}
  \sim
  \mathcal N\bigl(0,(\sigma_{ms}^{x})^2\bigr),
  \qquad
  \omega_{ksc}^{u}
  \sim
  \mathcal N\bigl(0,(\sigma_{ks}^{u})^2\bigr),
  \qquad
  \alpha_c
  \sim
  \mathcal N(0,\sigma_\alpha^2).
  \label{eq:supp_hierarchical_pooling}
\end{equation}
This is the global-plus-deviation construction of hierarchical DLNMs \cite{economou2024hierarchical}, expressed through global and group-specific neural subnetworks. Unlike the Multitask construction, the deviation subnetworks are not shared across groups. The Gaussian hierarchy shrinks their mixing weights toward zero and thereby pools each group toward the global components. Because a deviation subnetwork and its mixing weight enter only through their product, the pooling standard deviations are defined relative to the scale of the deviation subnetworks, which the weight-decay penalty fixes in fitting. As the group-level variances approach zero, Equation~\eqref{eq:supp_hierarchical_predictor} reduces to a common Mixture DLNAM. Nested or crossed groupings follow by adding the corresponding deviation terms and pooling distributions. Full uncertainty propagation for the group-specific deviations and pooling parameters would require extending the last-layer construction of Section~\ref{supp:laplace}, which conditions on fitted mixing weights and learned representations.

\section{Surface-Target Comparisons}
\label{supp:surface}

\begin{figure}[!tbp]
  \centering
  \includegraphics[width=\linewidth]{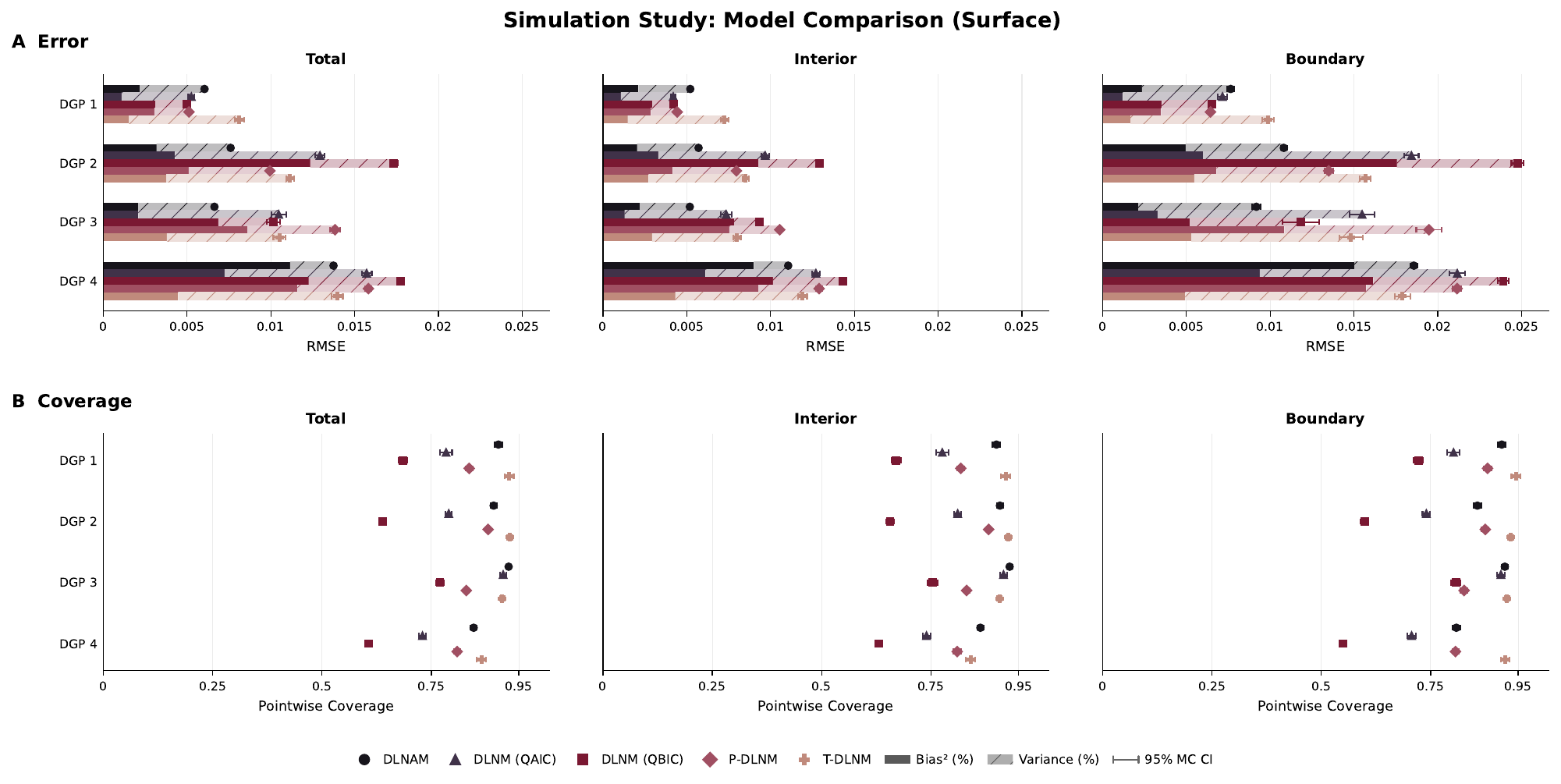}
  \caption{\textbf{Simulation Study: Model Comparison (Surface).} Monte Carlo simulation of four DGPs, with \(R=200\) replicates per DGP. (\textbf{A}) Log-relative-risk RMSE by method and exposure region, partitioned into squared bias and variance. (\textbf{B}) Empirical coverage of pointwise \(95\%\) intervals. Error bars denote 95\% Monte Carlo confidence intervals. Boundary denotes exposure values below the 5th or above the 95th percentile of the reference exposure distribution; interior denotes its complement. All lags are included within each exposure region. Associated numerical results are reported in Tables~\ref{tab:supp_mc_surf} and~\ref{tab:supp_mc_surf_bv}.}
  \label{fig:supp_mc_surf}
\end{figure}

\begin{figure}[!tbp]
  \centering
  \includegraphics[width=\linewidth]{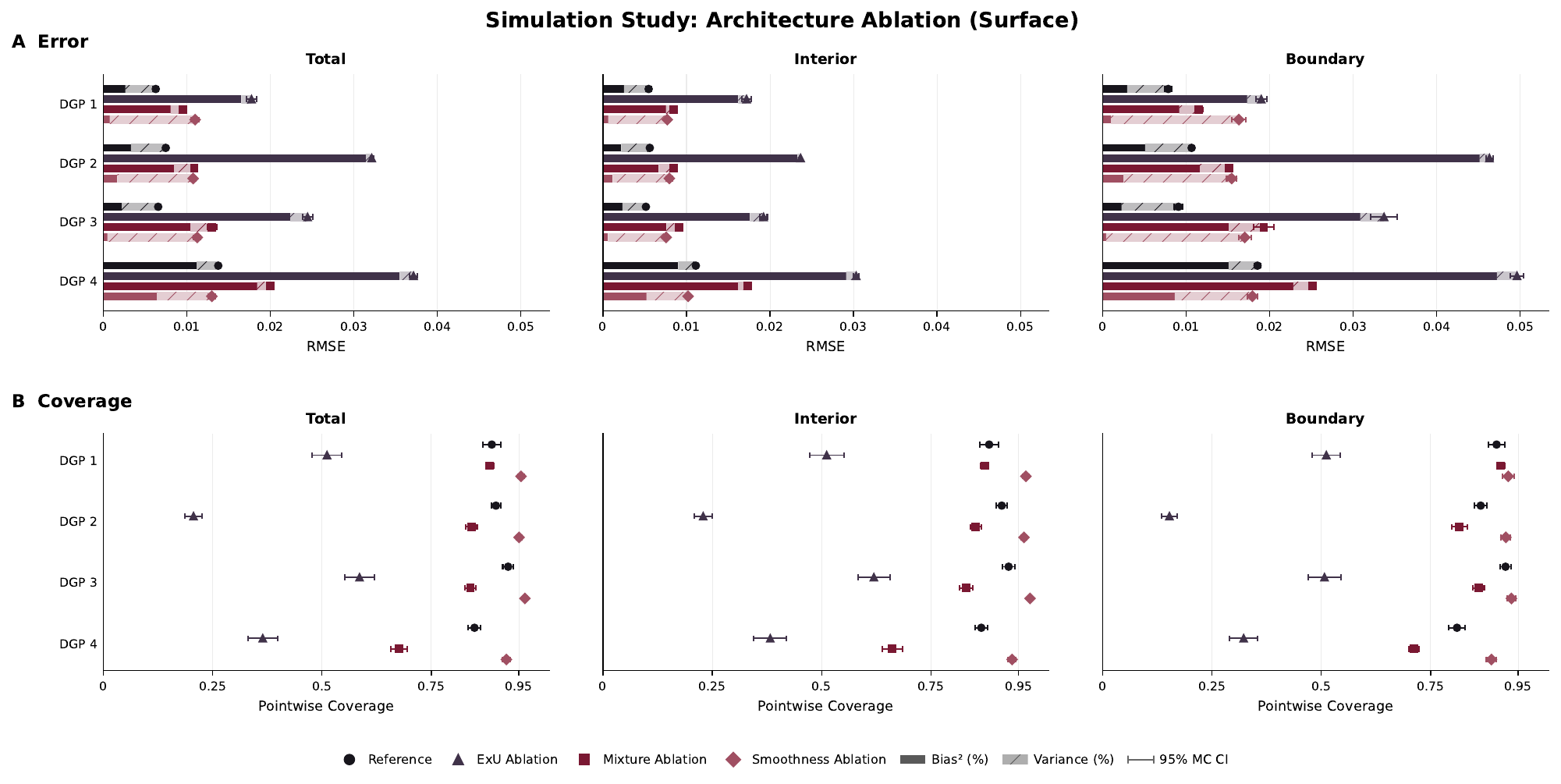}
  \caption{\textbf{Simulation Study: Architecture Ablation (Surface).} Monte Carlo simulation of four DGPs, with \(R=50\) replicates per DGP. (\textbf{A}) Log-relative-risk RMSE by method and exposure region, partitioned into squared bias and variance. (\textbf{B}) Empirical coverage of pointwise \(95\%\) intervals. Error bars denote 95\% Monte Carlo confidence intervals. Boundary denotes exposure values below the 5th or above the 95th percentile of the reference exposure distribution; interior denotes its complement. All lags are included within each exposure region. Ablations are applied one at a time and are not factorial. Associated numerical results are reported in Table~\ref{tab:supp_abl_surf}.}
  \label{fig:supp_abl_surf}
\end{figure}

The main text emphasizes the cumulative curve as the primary simulation estimand and applied summary. Here we repeat the model comparison and architecture ablation for the full exposure--lag surface, scoring every exposure-grid point at every lag as defined in Section~\ref{sec:metrics}. Figure~\ref{fig:supp_mc_surf} and Tables~\ref{tab:supp_mc_surf}, \ref{tab:supp_mc_surf_bv}, and~\ref{tab:supp_mc_surf_rel} give the estimator comparison; Fig.~\ref{fig:supp_abl_surf} and Table~\ref{tab:supp_abl_surf} give the ablation. As discussed in Section~\ref{sec:metrics}, surface scoring provides a complementary, finer-grained assessment of how the fitted effect is allocated across exposure and lag; it is not treated as a superior recovery target to the cumulative summary.

\section{Analysis Specifications}
\label{supp:fitting}

This section collects supplementary implementation details and the study-specific specifications used in the principal simulation and applications. Tables~\ref{tab:supp_fitting} and~\ref{tab:supp_comparators} summarize the study design, DLNAM fitting, comparator, adjustment, and computing specifications; settings shared across analyses are listed once, with study-specific departures shown separately.

For the criterion-selected cross-basis DLNMs, a candidate with \(k\) cross-basis coefficients and fitted dispersion \(\hat\varphi\) is scored by
\[
  \operatorname{QAIC}
  =
  -2\ell(\hat\mu)+2\hat\varphi k,
  \qquad
  \operatorname{QBIC}
  =
  -2\ell(\hat\mu)+\log(n)\,\hat\varphi k.
\]
The two criteria differ only in the penalty applied to cross-basis complexity.

The nonlinear T-DLNM implementation requires a two-dimensional fixed-effect design, so a single centered linear index was included as a fixed-effect regressor; it forms no part of the DGP. Fits that failed to initialize were re-attempted with a new seed, up to ten times. Of \(800\) treed fits in the principal model comparison, \(800\) completed, \(83\) required more than one attempt (at most \(7\)), and \(0\) were excluded.

\begin{table}[!tbp]
\centering
\footnotesize
\renewcommand{\arraystretch}{1.04}
\setlength{\tabcolsep}{4.5pt}
\begin{tabularx}{\textwidth}{>{\raggedright\arraybackslash}p{2.65cm} *{3}{>{\raggedright\arraybackslash}X}}
\toprule
\textbf{Setting} & \textbf{Simulation} & \textbf{Chicago} & \textbf{Malaria} \\
\midrule
\multicolumn{4}{l}{\textbf{A. Analysis Design}} \\
\addlinespace[0.2em]
Family / Link       & Poisson; Log & Quasi-Poisson; Log & Bernoulli; Logit \\
Dispersion          & Fixed: 1 & Pearson: $\hat\varphi=1.03$ & Fixed: 1 \\
Observations        & $n=5000$ & $n=4638$ & $n=348{,}565$ \\
Exposures / Lags    & 1; 0--14 & 1; 0--30 & 5; 1--6 \\
Reference           & $x^\star=20$ & Median & Median \\
\bottomrule
\end{tabularx}

\begin{tabularx}{\textwidth}{>{\raggedright\arraybackslash}p{2.90cm} >{\raggedright\arraybackslash}p{2.40cm} >{\raggedright\arraybackslash}X}
\toprule
\textbf{Setting} & \textbf{Applicability} & \textbf{Specification} \\
\midrule
\multicolumn{3}{l}{\textbf{B. DLNAM Architecture and Fitting}} \\
\addlinespace[0.2em]
Ensemble Members, Subnetworks & All & $B=3$; $S=3$ \\
Surface Component & All & ExU Input: 128; Hidden Layer: 128; Mish; Linear Output \\
Exposure Input & All & Min--Max Scaling; Concatenation ExU; $(\mu_w,\sigma_w)=(1.5,0.5)$ \\
Lag Input & All & Min--Max Scaling; Concatenation ExU; $(\mu_w,\sigma_w)=(2.5,0.5)$ \\
Trend Network & Chicago & ExU; Widths: 128--128--128; Mish; $(\mu_w,\sigma_w)=(4.5,0.5)$ \\
Covariate Networks & Chicago & Width: 32; Mish; Linear Input; Z-Score \\
 & Malaria & Month + Year; Width: 32; Mish; Linear Input; Z-Score \\
Grouping Encoding & Malaria & Cluster + Country; Learned Scalar per Level; Indexed Lookup \\
Last-Layer Parameters & All & Global Intercept + Target Final Layer \\
Mixing Weights & All & $\sigma_\omega=0.1$; Unconstrained \\
Optimizer & Simulation, Chicago, Malaria & AdamW; $\eta=8\cdot10^{-4}$; $\eta_{\min}=1\cdot10^{-4}$; Gradient Clip $=10$ \\
 & Joint Simulation & AdamW; $\eta=7\cdot10^{-4}$; $\eta_{\min}=1\cdot10^{-4}$; Gradient Clip $=10$ \\
Scheduler & All & Cosine Annealing; Step: Per Update \\
Epochs, Batching & Simulation & $E=2500$; Full Batch \\
 & Chicago & $E=5000$; Full Batch \\
 & Joint Simulation & $E=5000$; Full Batch \\
 & Malaria & $E=50$; Minibatch $1\%$ \\
Regularization & All & Weight Decay: $\lambda_w=1\cdot10^{-4}$; Dropout: 0; Subnetwork Dropout: 0; Output Penalty: 0 \\
 & Simulation, Malaria & Surface Roughness Penalty: 0 \\
 & Chicago & Surface Roughness Penalty: 0.003 \\
Uncertainty & All & Last-Layer Laplace + Between-Member Variation \\
\bottomrule
\end{tabularx}
\caption{\textbf{Study Design and DLNAM Specifications.} Analysis design and DLNAM fitting specifications for the principal simulation and applications; joint-exposure departures are shown where applicable.}
\label{tab:supp_fitting}
\end{table}

\begin{table}[p]
\centering
\footnotesize
\renewcommand{\arraystretch}{0.86}
\setlength{\tabcolsep}{4pt}
\begin{tabularx}{\textwidth}{>{\raggedright\arraybackslash}p{2.85cm} >{\raggedright\arraybackslash}p{2.25cm} >{\raggedright\arraybackslash}X}
\toprule
\textbf{Setting} & \textbf{Applicability} & \textbf{Specification} \\
\midrule
\multicolumn{3}{l}{\textbf{A. DLNM}} \\
\addlinespace[0.15em]
Exposure Margin & Simulation, Chicago & Natural Cubic Spline; Equally Spaced Knots \\
 & Malaria & Natural Cubic Spline; $3$ df; Default Knots \\
Lag Margin & Simulation, Chicago & Natural Cubic Spline; Log-Spaced Knots \\
 & Malaria & Natural Cubic Spline; $3$ df \\
Dimension Grid & Simulation & $(v_x,v_\ell)\in\{2,\ldots,10\}^2$ \\
 & Chicago & $(v_x,v_\ell)\in\{2,\ldots,5\}^2$ \\
Selection Criterion & Simulation, Chicago & QAIC; QBIC \\
Dispersion / Intervals & Simulation, Chicago & Pearson Dispersion; Pointwise Wald Intervals \\
\addlinespace[0.30em]
\multicolumn{3}{l}{\textbf{B. P-DLNM}} \\
\addlinespace[0.12em]
Marginal Bases & Simulation & P-Spline; $10\times10$ \\
 & Chicago & P-Spline; $5\times5$ \\
Smoothing & Simulation, Chicago & Second-Order Difference Penalties; REML \\
Intervals & Simulation, Chicago & Pointwise Wald Intervals \\
\addlinespace[0.30em]
\multicolumn{3}{l}{\textbf{C. T-DLNM}} \\
\addlinespace[0.12em]
Likelihood & Simulation, Chicago & Gaussian; Response: $\log(1+Y_t)$ \\
Trees, Exposure Splits & Simulation, Chicago & $20$ Trees; Package Priors; $30$ Exposure Splits \\
Sampler & Simulation, Chicago & Burn-In: $5{,}000$; Post-Burn: $15{,}000$; Thinning: $10$; Retry Cap: $10$ \\
Intervals & Simulation, Chicago & Posterior Credible Intervals \\
\addlinespace[0.30em]
\multicolumn{3}{l}{\textbf{D. Adjustment}} \\
\addlinespace[0.12em]
Trend, Seasonality & Chicago & Date Spline; $98$ df \\
 & Malaria & Month Spline; 4 df \\
Long-Term Trend & Malaria & Survey Year; Linear \\
Other Covariates & Chicago & Day of Week: Categorical (Monday Reference); Dew Point: Spline, 3 df; Ozone: Linear; $\mathrm{PM}_{10}$: Linear; Exposure Means: Lags 0--1 \\
Target / Controls & Malaria & Temperature: Precipitation; Precipitation: Temperature; Soil Moisture: Temperature + Precipitation; Evapotranspiration: Soil Moisture + Temperature + Precipitation; Specific Humidity: Soil Moisture + Temperature + Precipitation + Evapotranspiration \\
Multi-Exposure Representation & Malaria & DLNM: Target-Specific Fit; Controls: Scalar Lag Means; DLNAM: Target-Specific Fit; Controls: Full Lag Surfaces \\
Hierarchy & Malaria & DLNM: Random Intercepts, Estimated Variance (Cluster, Country); DLNAM: Level-Specific Effects, Fixed Ridge Penalty \\
\addlinespace[0.30em]
\multicolumn{3}{l}{\textbf{E. Software and Computing}} \\
\addlinespace[0.12em]
Software & Simulation & Python 3.10.12; PyTorch 2.5.0a0+e000cf0ad9.nv24.10; NumPy 1.24.4; pandas 2.2.2; R 4.6.1 \\
 & Applications, Scaling & Python 3.11.9; PyTorch 2.13.0+cu126; NumPy 2.4.6; pandas 3.0.3; R 4.5.0 \\
 & All & \texttt{dlnm} 2.4.10; \texttt{mgcv} 1.9.4; \texttt{dlmtree} 1.1.1; \texttt{glmmTMB} 1.1.14 \\
DLNAM Device & All & CUDA \\
Main MC Compute & Simulation & Alvis, Chalmers University of Technology \\
Workstation & Applications, Scaling & Intel Core i7-9700K (8 Logical Cores); 16~GB RAM; NVIDIA GeForce GTX 1070 (8~GB) \\
Scaling Benchmark & Computational Scaling & $n=5000$; $M=1,\ldots,4$; $R=3$; $2500$ Epochs \\
Timing Definition & Computational Scaling & Construction/Preprocessing + Fitting + Uncertainty + Cumulative Extraction; Excludes Data Generation + Startup; T-DLNM: Sum of Target-Specific Fits \\
\bottomrule
\end{tabularx}
\caption{\textbf{Comparator and Analysis Specifications.} Comparator, adjustment, software, and computing specifications for the principal simulation, applications, and computational-scaling benchmark.}
\label{tab:supp_comparators}
\end{table}

\clearpage

\begin{sidewaystable}[p]
\centering
\small
\setlength{\tabcolsep}{3pt}
\begin{tabular}{ll ccc ccc}
\toprule
 & & \multicolumn{3}{c}{RMSE (MCSE)} & \multicolumn{3}{c}{Coverage (MCSE)}\\
\cmidrule(lr){3-5}\cmidrule(lr){6-8}
DGP & Model & Total & Interior & Boundary & Total & Interior & Boundary\\
\midrule
DGP 1 & DLNAM & 0.0167 (0.0005) & 0.0130 (0.0004) & 0.0232 (0.0010) & 0.850 (0.012) & 0.815 (0.014) & 0.931 (0.009)\\
 & DLNM (QAIC) & 0.0225 (0.0007) & 0.0147 (0.0005) & 0.0346 (0.0015) & 0.795 (0.012) & 0.756 (0.016) & 0.888 (0.011)\\
 & DLNM (QBIC) & 0.0238 (0.0005) & 0.0191 (0.0005) & 0.0323 (0.0010) & 0.633 (0.012) & 0.563 (0.016) & 0.798 (0.012)\\
 & P-DLNM & 0.0256 (0.0005) & 0.0220 (0.0004) & 0.0326 (0.0011) & 0.654 (0.009) & 0.545 (0.012) & 0.909 (0.009)\\
 & T-DLNM & 0.0155 (0.0002) & 0.0153 (0.0002) & 0.0161 (0.0005) & 0.752 (0.007) & 0.661 (0.009) & 0.966 (0.007)\\
\addlinespace
DGP 2 & DLNAM & 0.0172 (0.0004) & 0.0120 (0.0003) & 0.0256 (0.0009) & 0.922 (0.007) & 0.913 (0.008) & 0.943 (0.008)\\
 & DLNM (QAIC) & 0.0287 (0.0007) & 0.0192 (0.0005) & 0.0435 (0.0013) & 0.801 (0.011) & 0.777 (0.013) & 0.858 (0.010)\\
 & DLNM (QBIC) & 0.0370 (0.0008) & 0.0269 (0.0005) & 0.0537 (0.0017) & 0.560 (0.012) & 0.510 (0.014) & 0.676 (0.013)\\
 & P-DLNM & 0.0440 (0.0008) & 0.0364 (0.0005) & 0.0579 (0.0020) & 0.490 (0.008) & 0.402 (0.010) & 0.696 (0.011)\\
 & T-DLNM & 0.0345 (0.0006) & 0.0250 (0.0003) & 0.0502 (0.0013) & 0.646 (0.008) & 0.615 (0.010) & 0.719 (0.009)\\
\addlinespace
DGP 3 & DLNAM & 0.0240 (0.0014) & 0.0114 (0.0003) & 0.0404 (0.0028) & 0.935 (0.006) & 0.935 (0.008) & 0.938 (0.008)\\
 & DLNM (QAIC) & 0.0514 (0.0016) & 0.0293 (0.0008) & 0.0826 (0.0032) & 0.642 (0.013) & 0.635 (0.016) & 0.656 (0.011)\\
 & DLNM (QBIC) & 0.0538 (0.0024) & 0.0306 (0.0010) & 0.0867 (0.0048) & 0.624 (0.014) & 0.614 (0.017) & 0.646 (0.012)\\
 & P-DLNM & 0.0981 (0.0016) & 0.0640 (0.0009) & 0.1504 (0.0036) & 0.294 (0.009) & 0.239 (0.010) & 0.422 (0.009)\\
 & T-DLNM & 0.0504 (0.0015) & 0.0337 (0.0006) & 0.0764 (0.0030) & 0.476 (0.009) & 0.445 (0.010) & 0.549 (0.010)\\
\addlinespace
DGP 4 & DLNAM & 0.0223 (0.0005) & 0.0157 (0.0004) & 0.0329 (0.0010) & 0.895 (0.007) & 0.894 (0.008) & 0.898 (0.010)\\
 & DLNM (QAIC) & 0.0363 (0.0006) & 0.0301 (0.0007) & 0.0478 (0.0011) & 0.533 (0.008) & 0.468 (0.008) & 0.685 (0.013)\\
 & DLNM (QBIC) & 0.0351 (0.0005) & 0.0285 (0.0004) & 0.0471 (0.0010) & 0.531 (0.007) & 0.463 (0.008) & 0.692 (0.011)\\
 & P-DLNM & 0.0399 (0.0012) & 0.0369 (0.0014) & 0.0463 (0.0012) & 0.485 (0.009) & 0.391 (0.010) & 0.704 (0.012)\\
 & T-DLNM & 0.0359 (0.0006) & 0.0290 (0.0004) & 0.0483 (0.0013) & 0.645 (0.008) & 0.581 (0.009) & 0.795 (0.012)\\
\addlinespace
\bottomrule
\end{tabular}
\caption{\textbf{Simulation Study: Model Comparison.} Cumulative log-relative-risk RMSE and pointwise 95\% interval coverage by DGP, method, and exposure region, with Monte Carlo standard errors in parentheses; $R=200$.}
\label{tab:supp_mc}
\end{sidewaystable}

\begin{sidewaystable}[p]
\centering
\small
\setlength{\tabcolsep}{3pt}
\begin{tabular}{ll ccc ccc}
\toprule
 & & \multicolumn{3}{c}{$10^{3}\times\mathrm{Bias}^2$ (MCSE)} & \multicolumn{3}{c}{$10^{3}\times\mathrm{Variance}$ (MCSE)}\\
\cmidrule(lr){3-5}\cmidrule(lr){6-8}
DGP & Model & Total & Interior & Boundary & Total & Interior & Boundary\\
\midrule
DGP 1 & DLNAM & 0.0769 (0.0082) & 0.0837 (0.0081) & 0.0609 (0.0163) & 0.2023 (0.0134) & 0.0845 (0.0055) & 0.4790 (0.0421)\\
 & DLNM (QAIC) & 0.0637 (0.0100) & 0.0658 (0.0084) & 0.0588 (0.0230) & 0.4444 (0.0321) & 0.1497 (0.0091) & 1.1368 (0.1008)\\
 & DLNM (QBIC) & 0.1657 (0.0156) & 0.1836 (0.0174) & 0.1238 (0.0277) & 0.4012 (0.0203) & 0.1802 (0.0087) & 0.9204 (0.0592)\\
 & P-DLNM & 0.2916 (0.0173) & 0.3271 (0.0186) & 0.2082 (0.0347) & 0.3648 (0.0196) & 0.1551 (0.0100) & 0.8576 (0.0585)\\
 & T-DLNM & 0.1403 (0.0043) & 0.1730 (0.0051) & 0.0634 (0.0105) & 0.1008 (0.0063) & 0.0607 (0.0045) & 0.1950 (0.0162)\\
\addlinespace
DGP 2 & DLNAM & 0.0291 (0.0040) & 0.0365 (0.0042) & 0.0117 (0.0060) & 0.2664 (0.0149) & 0.1063 (0.0055) & 0.6427 (0.0447)\\
 & DLNM (QAIC) & 0.1467 (0.0143) & 0.1261 (0.0095) & 0.1951 (0.0360) & 0.6766 (0.0400) & 0.2407 (0.0175) & 1.7009 (0.1188)\\
 & DLNM (QBIC) & 0.6632 (0.0485) & 0.4472 (0.0249) & 1.1707 (0.1331) & 0.7073 (0.0375) & 0.2774 (0.0140) & 1.7178 (0.1199)\\
 & P-DLNM & 1.2549 (0.0489) & 1.0649 (0.0374) & 1.7013 (0.1476) & 0.6778 (0.0381) & 0.2632 (0.0168) & 1.6521 (0.1172)\\
 & T-DLNM & 0.9377 (0.0302) & 0.5337 (0.0109) & 1.8869 (0.0996) & 0.2537 (0.0174) & 0.0933 (0.0063) & 0.6306 (0.0532)\\
\addlinespace
DGP 3 & DLNAM & 0.1029 (0.0208) & 0.0097 (0.0014) & 0.3219 (0.0698) & 0.4742 (0.0554) & 0.1198 (0.0067) & 1.3073 (0.1812)\\
 & DLNM (QAIC) & 1.1575 (0.0888) & 0.4120 (0.0302) & 2.9095 (0.2583) & 1.4815 (0.1202) & 0.4461 (0.0284) & 3.9148 (0.3881)\\
 & DLNM (QBIC) & 1.2833 (0.1040) & 0.4436 (0.0366) & 3.2567 (0.3071) & 1.6143 (0.1927) & 0.4917 (0.0338) & 4.2523 (0.6377)\\
 & P-DLNM & 6.5289 (0.2732) & 3.3630 (0.1293) & 13.9687 (0.8815) & 3.0937 (0.2279) & 0.7324 (0.0465) & 8.6428 (0.7251)\\
 & T-DLNM & 1.8178 (0.0928) & 0.9466 (0.0351) & 3.8650 (0.2569) & 0.7208 (0.0837) & 0.1916 (0.0164) & 1.9644 (0.2673)\\
\addlinespace
DGP 4 & DLNAM & 0.1859 (0.0164) & 0.0850 (0.0073) & 0.4228 (0.0485) & 0.3092 (0.0170) & 0.1600 (0.0109) & 0.6599 (0.0430)\\
 & DLNM (QAIC) & 0.5428 (0.0227) & 0.4653 (0.0215) & 0.7251 (0.0408) & 0.7744 (0.0374) & 0.4406 (0.0285) & 1.5588 (0.0972)\\
 & DLNM (QBIC) & 0.4869 (0.0141) & 0.4079 (0.0072) & 0.6726 (0.0401) & 0.7443 (0.0337) & 0.4031 (0.0236) & 1.5463 (0.0914)\\
 & P-DLNM & 0.5275 (0.0298) & 0.5991 (0.0439) & 0.3592 (0.0312) & 1.0675 (0.0776) & 0.7618 (0.0676) & 1.7860 (0.1207)\\
 & T-DLNM & 0.8138 (0.0327) & 0.6520 (0.0182) & 1.1938 (0.0949) & 0.4721 (0.0244) & 0.1889 (0.0148) & 1.1378 (0.0698)\\
\addlinespace
\bottomrule
\end{tabular}
\caption{\textbf{Simulation Study: Bias and Variance.} Squared bias and variance of cumulative log-relative-risk by DGP, method, and exposure region, with Monte Carlo standard errors in parentheses; $R=200$.}
\label{tab:supp_mc_bv}
\end{sidewaystable}

\begin{sidewaystable}[p]
\centering
\small
\setlength{\tabcolsep}{3pt}
\begin{tabular}{ll ccc ccc}
\toprule
 & & \multicolumn{3}{c}{Relative RMSE} & \multicolumn{3}{c}{Mean Interval Width}\\
\cmidrule(lr){3-5}\cmidrule(lr){6-8}
DGP & Model & Total & Interior & Boundary & Total & Interior & Boundary\\
\midrule
DGP 1 & DLNAM & 1.00 & 0.78 & 1.39 & 0.046 & 0.034 & 0.076\\
 & DLNM (QAIC) & 1.35 & 0.88 & 2.07 & 0.053 & 0.034 & 0.098\\
 & DLNM (QBIC) & 1.42 & 1.14 & 1.93 & 0.043 & 0.027 & 0.080\\
 & P-DLNM & 1.53 & 1.31 & 1.95 & 0.054 & 0.035 & 0.100\\
 & T-DLNM & 0.93 & 0.91 & 0.96 & 0.042 & 0.030 & 0.069\\
\addlinespace
DGP 2 & DLNAM & 1.00 & 0.70 & 1.49 & 0.053 & 0.038 & 0.088\\
 & DLNM (QAIC) & 1.67 & 1.11 & 2.53 & 0.068 & 0.046 & 0.120\\
 & DLNM (QBIC) & 2.15 & 1.57 & 3.13 & 0.051 & 0.033 & 0.093\\
 & P-DLNM & 2.56 & 2.12 & 3.37 & 0.059 & 0.038 & 0.107\\
 & T-DLNM & 2.01 & 1.46 & 2.92 & 0.054 & 0.036 & 0.094\\
\addlinespace
DGP 3 & DLNAM & 1.00 & 0.47 & 1.68 & 0.059 & 0.039 & 0.105\\
 & DLNM (QAIC) & 2.14 & 1.22 & 3.44 & 0.071 & 0.047 & 0.129\\
 & DLNM (QBIC) & 2.24 & 1.27 & 3.61 & 0.071 & 0.047 & 0.128\\
 & P-DLNM & 4.08 & 2.66 & 6.26 & 0.073 & 0.042 & 0.145\\
 & T-DLNM & 2.10 & 1.40 & 3.18 & 0.056 & 0.038 & 0.099\\
\addlinespace
DGP 4 & DLNAM & 1.00 & 0.70 & 1.48 & 0.062 & 0.048 & 0.095\\
 & DLNM (QAIC) & 1.63 & 1.35 & 2.15 & 0.062 & 0.042 & 0.107\\
 & DLNM (QBIC) & 1.58 & 1.28 & 2.12 & 0.060 & 0.041 & 0.104\\
 & P-DLNM & 1.79 & 1.66 & 2.08 & 0.058 & 0.038 & 0.105\\
 & T-DLNM & 1.61 & 1.30 & 2.17 & 0.066 & 0.045 & 0.114\\
\addlinespace
\bottomrule
\end{tabular}
\caption{\textbf{Simulation Study: Relative Error and Interval Width.} Relative RMSE and mean pointwise 95\% interval width by DGP, method, and exposure region; $R=200$. Relative RMSE uses the DLNAM total RMSE within each DGP as the common denominator. Interval width is on the log-relative-risk scale.}
\label{tab:supp_mc_rel}
\end{sidewaystable}

\begin{sidewaystable}[p]
\centering
\small
\setlength{\tabcolsep}{3pt}
\begin{tabular}{ll ccc ccc}
\toprule
 & & \multicolumn{3}{c}{Coverage -- Reported (MCSE)} & \multicolumn{3}{c}{Coverage -- Laplace (MCSE)}\\
\cmidrule(lr){3-5}\cmidrule(lr){6-8}
DGP & Model & Total & Interior & Boundary & Total & Interior & Boundary\\
\midrule
DGP 1 & DLNAM & 0.850 (0.012) & 0.815 (0.014) & 0.931 (0.009) & 0.816 (0.012) & 0.774 (0.015) & 0.917 (0.010)\\
 & DLNM (QAIC) & 0.795 (0.012) & 0.756 (0.016) & 0.888 (0.011) & -- & -- & --\\
 & DLNM (QBIC) & 0.633 (0.012) & 0.563 (0.016) & 0.798 (0.012) & -- & -- & --\\
 & P-DLNM & 0.654 (0.009) & 0.545 (0.012) & 0.909 (0.009) & -- & -- & --\\
 & T-DLNM & 0.752 (0.007) & 0.661 (0.009) & 0.966 (0.007) & -- & -- & --\\
\addlinespace
DGP 2 & DLNAM & 0.922 (0.007) & 0.913 (0.008) & 0.943 (0.008) & 0.892 (0.008) & 0.877 (0.010) & 0.925 (0.009)\\
 & DLNM (QAIC) & 0.801 (0.011) & 0.777 (0.013) & 0.858 (0.010) & -- & -- & --\\
 & DLNM (QBIC) & 0.560 (0.012) & 0.510 (0.014) & 0.676 (0.013) & -- & -- & --\\
 & P-DLNM & 0.490 (0.008) & 0.402 (0.010) & 0.696 (0.011) & -- & -- & --\\
 & T-DLNM & 0.646 (0.008) & 0.615 (0.010) & 0.719 (0.009) & -- & -- & --\\
\addlinespace
DGP 3 & DLNAM & 0.935 (0.006) & 0.935 (0.008) & 0.938 (0.008) & 0.903 (0.008) & 0.899 (0.009) & 0.913 (0.009)\\
 & DLNM (QAIC) & 0.642 (0.013) & 0.635 (0.016) & 0.656 (0.011) & -- & -- & --\\
 & DLNM (QBIC) & 0.624 (0.014) & 0.614 (0.017) & 0.646 (0.012) & -- & -- & --\\
 & P-DLNM & 0.294 (0.009) & 0.239 (0.010) & 0.422 (0.009) & -- & -- & --\\
 & T-DLNM & 0.476 (0.009) & 0.445 (0.010) & 0.549 (0.010) & -- & -- & --\\
\addlinespace
DGP 4 & DLNAM & 0.895 (0.007) & 0.894 (0.008) & 0.898 (0.010) & 0.806 (0.010) & 0.792 (0.011) & 0.837 (0.012)\\
 & DLNM (QAIC) & 0.533 (0.008) & 0.468 (0.008) & 0.685 (0.013) & -- & -- & --\\
 & DLNM (QBIC) & 0.531 (0.007) & 0.463 (0.008) & 0.692 (0.011) & -- & -- & --\\
 & P-DLNM & 0.485 (0.009) & 0.391 (0.010) & 0.704 (0.012) & -- & -- & --\\
 & T-DLNM & 0.645 (0.008) & 0.581 (0.009) & 0.795 (0.012) & -- & -- & --\\
\addlinespace
\bottomrule
\end{tabular}
\caption{\textbf{Simulation Study: DLNAM Interval Decomposition.} Empirical pointwise 95\% interval coverage for the reported interval and, for DLNAM, the conditional last-layer Laplace interval, by DGP, method, and exposure region, with Monte Carlo standard errors in parentheses; $R=200$.}
\label{tab:supp_cov_decomp}
\end{sidewaystable}

\begin{sidewaystable}[p]
\centering
\small
\setlength{\tabcolsep}{3pt}
\begin{tabular}{ll ccc ccc}
\toprule
 & & \multicolumn{3}{c}{RMSE (MCSE)} & \multicolumn{3}{c}{Coverage (MCSE)}\\
\cmidrule(lr){3-5}\cmidrule(lr){6-8}
DGP & Encoder & Total & Interior & Boundary & Total & Interior & Boundary\\
\midrule
DGP 1 & Concatenation & 0.0169 (0.0013) & 0.0126 (0.0008) & 0.0243 (0.0029) & 0.863 (0.023) & 0.834 (0.029) & 0.931 (0.020)\\
 & Unified Shared Bias & 0.0185 (0.0011) & 0.0158 (0.0010) & 0.0237 (0.0017) & 0.877 (0.022) & 0.861 (0.026) & 0.912 (0.021)\\
 & Unified Local Bias & 0.0172 (0.0008) & 0.0150 (0.0008) & 0.0215 (0.0014) & 0.760 (0.028) & 0.713 (0.033) & 0.871 (0.024)\\
\addlinespace
DGP 2 & Concatenation & 0.0172 (0.0009) & 0.0111 (0.0005) & 0.0264 (0.0018) & 0.940 (0.009) & 0.935 (0.012) & 0.951 (0.013)\\
 & Unified Shared Bias & 0.0325 (0.0018) & 0.0257 (0.0012) & 0.0445 (0.0031) & 0.778 (0.024) & 0.767 (0.025) & 0.806 (0.030)\\
 & Unified Local Bias & 0.0214 (0.0012) & 0.0175 (0.0010) & 0.0286 (0.0021) & 0.831 (0.021) & 0.808 (0.026) & 0.885 (0.023)\\
\addlinespace
DGP 3 & Concatenation & 0.0237 (0.0032) & 0.0113 (0.0007) & 0.0397 (0.0061) & 0.937 (0.013) & 0.939 (0.016) & 0.934 (0.019)\\
 & Unified Shared Bias & 0.0442 (0.0015) & 0.0329 (0.0015) & 0.0632 (0.0024) & 0.788 (0.021) & 0.787 (0.026) & 0.790 (0.019)\\
 & Unified Local Bias & 0.0346 (0.0024) & 0.0167 (0.0008) & 0.0580 (0.0046) & 0.813 (0.019) & 0.845 (0.021) & 0.738 (0.030)\\
\addlinespace
DGP 4 & Concatenation & 0.0215 (0.0011) & 0.0158 (0.0012) & 0.0311 (0.0017) & 0.903 (0.013) & 0.907 (0.015) & 0.892 (0.020)\\
 & Unified Shared Bias & 0.0651 (0.0022) & 0.0620 (0.0019) & 0.0720 (0.0033) & 0.511 (0.030) & 0.515 (0.030) & 0.502 (0.039)\\
 & Unified Local Bias & 0.0392 (0.0024) & 0.0387 (0.0024) & 0.0405 (0.0029) & 0.749 (0.032) & 0.726 (0.033) & 0.804 (0.035)\\
\addlinespace
\bottomrule
\end{tabular}
\caption{\textbf{Simulation Study: ExU Comparison.} Cumulative log-relative-risk RMSE and pointwise 95\% interval coverage by DGP, method, and exposure region, with Monte Carlo standard errors in parentheses; $R=50$.}
\label{tab:supp_exu}
\end{sidewaystable}

\begin{sidewaystable}[p]
\centering
\small
\setlength{\tabcolsep}{3pt}
\begin{tabular}{ll ccc ccc}
\toprule
 & & \multicolumn{3}{c}{RMSE (MCSE)} & \multicolumn{3}{c}{Coverage (MCSE)}\\
\cmidrule(lr){3-5}\cmidrule(lr){6-8}
DGP & Configuration & Total & Interior & Boundary & Total & Interior & Boundary\\
\midrule
DGP 1 & Reference & 0.0169 (0.0013) & 0.0126 (0.0008) & 0.0243 (0.0029) & 0.863 (0.023) & 0.834 (0.029) & 0.931 (0.020)\\
 & ExU Ablation & 0.0570 (0.0034) & 0.0584 (0.0045) & 0.0535 (0.0027) & 0.729 (0.028) & 0.697 (0.030) & 0.805 (0.032)\\
 & Mixture Ablation & 0.0252 (0.0011) & 0.0181 (0.0006) & 0.0368 (0.0020) & 0.779 (0.021) & 0.738 (0.021) & 0.875 (0.028)\\
 & Smoothness Ablation & 0.0502 (0.0028) & 0.0231 (0.0011) & 0.0848 (0.0055) & 0.944 (0.008) & 0.967 (0.009) & 0.891 (0.013)\\
\addlinespace
DGP 2 & Reference & 0.0172 (0.0009) & 0.0111 (0.0005) & 0.0264 (0.0018) & 0.940 (0.009) & 0.935 (0.012) & 0.951 (0.013)\\
 & ExU Ablation & 0.1402 (0.0021) & 0.0726 (0.0008) & 0.2313 (0.0047) & 0.114 (0.018) & 0.129 (0.021) & 0.080 (0.016)\\
 & Mixture Ablation & 0.0342 (0.0009) & 0.0280 (0.0005) & 0.0454 (0.0018) & 0.595 (0.019) & 0.543 (0.017) & 0.717 (0.031)\\
 & Smoothness Ablation & 0.0429 (0.0022) & 0.0228 (0.0008) & 0.0703 (0.0042) & 0.963 (0.005) & 0.981 (0.004) & 0.922 (0.012)\\
\addlinespace
DGP 3 & Reference & 0.0237 (0.0032) & 0.0113 (0.0007) & 0.0397 (0.0061) & 0.937 (0.013) & 0.939 (0.016) & 0.934 (0.019)\\
 & ExU Ablation & 0.1869 (0.0051) & 0.1031 (0.0012) & 0.3033 (0.0104) & 0.468 (0.022) & 0.496 (0.025) & 0.403 (0.024)\\
 & Mixture Ablation & 0.0911 (0.0042) & 0.0453 (0.0009) & 0.1515 (0.0085) & 0.512 (0.020) & 0.508 (0.021) & 0.522 (0.032)\\
 & Smoothness Ablation & 0.0458 (0.0023) & 0.0233 (0.0008) & 0.0759 (0.0046) & 0.966 (0.004) & 0.981 (0.003) & 0.930 (0.010)\\
\addlinespace
DGP 4 & Reference & 0.0215 (0.0011) & 0.0158 (0.0012) & 0.0311 (0.0017) & 0.903 (0.013) & 0.907 (0.015) & 0.892 (0.020)\\
 & ExU Ablation & 0.1540 (0.0046) & 0.0954 (0.0021) & 0.2409 (0.0095) & 0.414 (0.031) & 0.525 (0.041) & 0.152 (0.019)\\
 & Mixture Ablation & 0.0918 (0.0014) & 0.0856 (0.0013) & 0.1050 (0.0023) & 0.265 (0.012) & 0.264 (0.011) & 0.268 (0.020)\\
 & Smoothness Ablation & 0.0364 (0.0014) & 0.0234 (0.0014) & 0.0561 (0.0026) & 0.965 (0.007) & 0.974 (0.009) & 0.943 (0.011)\\
\addlinespace
\bottomrule
\end{tabular}
\caption{\textbf{Simulation Study: Architecture Ablation.} Cumulative log-relative-risk RMSE and pointwise 95\% interval coverage by DGP, method, and exposure region, with Monte Carlo standard errors in parentheses; $R=50$.}
\label{tab:supp_abl}
\end{sidewaystable}

\begin{sidewaystable}[p]
\centering
\small
\setlength{\tabcolsep}{3pt}
\begin{tabular}{ll ccc ccc}
\toprule
 & & \multicolumn{3}{c}{RMSE (MCSE)} & \multicolumn{3}{c}{Coverage (MCSE)}\\
\cmidrule(lr){3-5}\cmidrule(lr){6-8}
DGP & Model & Total & Interior & Boundary & Total & Interior & Boundary\\
\midrule
DGP 1 & DLNAM & 0.0060 (0.0001) & 0.0052 (0.0001) & 0.0076 (0.0001) & 0.903 (0.004) & 0.900 (0.004) & 0.912 (0.004)\\
 & DLNM (QAIC) & 0.0053 (0.0001) & 0.0042 (0.0001) & 0.0072 (0.0001) & 0.784 (0.007) & 0.776 (0.007) & 0.802 (0.007)\\
 & DLNM (QBIC) & 0.0050 (0.0000) & 0.0042 (0.0000) & 0.0065 (0.0001) & 0.686 (0.005) & 0.670 (0.006) & 0.722 (0.005)\\
 & P-DLNM & 0.0051 (0.0000) & 0.0044 (0.0000) & 0.0064 (0.0001) & 0.837 (0.004) & 0.818 (0.004) & 0.880 (0.005)\\
 & T-DLNM & 0.0081 (0.0001) & 0.0072 (0.0001) & 0.0099 (0.0002) & 0.928 (0.005) & 0.921 (0.005) & 0.945 (0.005)\\
\addlinespace
DGP 2 & DLNAM & 0.0076 (0.0001) & 0.0057 (0.0001) & 0.0108 (0.0001) & 0.892 (0.003) & 0.908 (0.004) & 0.857 (0.004)\\
 & DLNM (QAIC) & 0.0129 (0.0001) & 0.0097 (0.0001) & 0.0184 (0.0002) & 0.790 (0.004) & 0.811 (0.004) & 0.740 (0.004)\\
 & DLNM (QBIC) & 0.0173 (0.0001) & 0.0129 (0.0001) & 0.0248 (0.0002) & 0.639 (0.004) & 0.656 (0.005) & 0.599 (0.005)\\
 & P-DLNM & 0.0099 (0.0001) & 0.0080 (0.0001) & 0.0135 (0.0001) & 0.880 (0.003) & 0.882 (0.003) & 0.875 (0.004)\\
 & T-DLNM & 0.0111 (0.0001) & 0.0085 (0.0001) & 0.0157 (0.0002) & 0.929 (0.003) & 0.927 (0.003) & 0.933 (0.003)\\
\addlinespace
DGP 3 & DLNAM & 0.0066 (0.0001) & 0.0052 (0.0000) & 0.0092 (0.0001) & 0.927 (0.003) & 0.930 (0.003) & 0.920 (0.003)\\
 & DLNM (QAIC) & 0.0105 (0.0002) & 0.0074 (0.0002) & 0.0155 (0.0004) & 0.914 (0.003) & 0.916 (0.004) & 0.911 (0.004)\\
 & DLNM (QBIC) & 0.0101 (0.0002) & 0.0093 (0.0000) & 0.0118 (0.0006) & 0.770 (0.005) & 0.754 (0.006) & 0.807 (0.005)\\
 & P-DLNM & 0.0138 (0.0002) & 0.0105 (0.0001) & 0.0195 (0.0004) & 0.830 (0.003) & 0.831 (0.003) & 0.827 (0.004)\\
 & T-DLNM & 0.0105 (0.0002) & 0.0080 (0.0001) & 0.0148 (0.0004) & 0.912 (0.003) & 0.907 (0.003) & 0.925 (0.003)\\
\addlinespace
DGP 4 & DLNAM & 0.0137 (0.0001) & 0.0111 (0.0001) & 0.0186 (0.0001) & 0.847 (0.003) & 0.863 (0.003) & 0.809 (0.005)\\
 & DLNM (QAIC) & 0.0157 (0.0002) & 0.0127 (0.0001) & 0.0212 (0.0002) & 0.730 (0.004) & 0.740 (0.005) & 0.706 (0.005)\\
 & DLNM (QBIC) & 0.0177 (0.0001) & 0.0143 (0.0001) & 0.0239 (0.0002) & 0.607 (0.004) & 0.631 (0.004) & 0.549 (0.004)\\
 & P-DLNM & 0.0158 (0.0001) & 0.0129 (0.0001) & 0.0212 (0.0001) & 0.809 (0.004) & 0.810 (0.004) & 0.807 (0.004)\\
 & T-DLNM & 0.0140 (0.0002) & 0.0119 (0.0001) & 0.0179 (0.0002) & 0.864 (0.005) & 0.840 (0.006) & 0.921 (0.005)\\
\addlinespace
\bottomrule
\end{tabular}
\caption{\textbf{Simulation Study: Model Comparison (Surface).} Log-relative-risk RMSE and pointwise 95\% interval coverage by DGP, method, and exposure region, with Monte Carlo standard errors in parentheses; $R=200$.}
\label{tab:supp_mc_surf}
\end{sidewaystable}

\begin{sidewaystable}[p]
\centering
\small
\setlength{\tabcolsep}{3pt}
\begin{tabular}{ll ccc ccc}
\toprule
 & & \multicolumn{3}{c}{$10^{3}\times\mathrm{Bias}^2$ (MCSE)} & \multicolumn{3}{c}{$10^{3}\times\mathrm{Variance}$ (MCSE)}\\
\cmidrule(lr){3-5}\cmidrule(lr){6-8}
DGP & Model & Total & Interior & Boundary & Total & Interior & Boundary\\
\midrule
DGP 1 & DLNAM & 0.0131 (0.0008) & 0.0110 (0.0006) & 0.0179 (0.0011) & 0.0234 (0.0008) & 0.0161 (0.0006) & 0.0405 (0.0014)\\
 & DLNM (QAIC) & 0.0057 (0.0008) & 0.0045 (0.0006) & 0.0086 (0.0013) & 0.0219 (0.0014) & 0.0131 (0.0010) & 0.0426 (0.0027)\\
 & DLNM (QBIC) & 0.0155 (0.0001) & 0.0123 (0.0001) & 0.0230 (0.0002) & 0.0095 (0.0003) & 0.0053 (0.0002) & 0.0194 (0.0007)\\
 & P-DLNM & 0.0156 (0.0005) & 0.0127 (0.0004) & 0.0224 (0.0008) & 0.0105 (0.0003) & 0.0070 (0.0003) & 0.0190 (0.0007)\\
 & T-DLNM & 0.0124 (0.0014) & 0.0107 (0.0011) & 0.0164 (0.0021) & 0.0536 (0.0024) & 0.0418 (0.0019) & 0.0812 (0.0037)\\
\addlinespace
DGP 2 & DLNAM & 0.0241 (0.0006) & 0.0116 (0.0003) & 0.0536 (0.0014) & 0.0336 (0.0010) & 0.0209 (0.0006) & 0.0633 (0.0019)\\
 & DLNM (QAIC) & 0.0553 (0.0026) & 0.0320 (0.0015) & 0.1103 (0.0055) & 0.1118 (0.0047) & 0.0618 (0.0029) & 0.2293 (0.0104)\\
 & DLNM (QBIC) & 0.2139 (0.0092) & 0.1199 (0.0051) & 0.4348 (0.0193) & 0.0864 (0.0068) & 0.0474 (0.0036) & 0.1780 (0.0149)\\
 & P-DLNM & 0.0506 (0.0013) & 0.0332 (0.0008) & 0.0917 (0.0029) & 0.0482 (0.0012) & 0.0303 (0.0008) & 0.0902 (0.0027)\\
 & T-DLNM & 0.0419 (0.0017) & 0.0232 (0.0009) & 0.0858 (0.0038) & 0.0824 (0.0028) & 0.0494 (0.0020) & 0.1601 (0.0053)\\
\addlinespace
DGP 3 & DLNAM & 0.0138 (0.0004) & 0.0114 (0.0002) & 0.0195 (0.0010) & 0.0301 (0.0008) & 0.0155 (0.0005) & 0.0646 (0.0021)\\
 & DLNM (QAIC) & 0.0218 (0.0015) & 0.0095 (0.0007) & 0.0508 (0.0044) & 0.0876 (0.0044) & 0.0445 (0.0024) & 0.1889 (0.0103)\\
 & DLNM (QBIC) & 0.0696 (0.0020) & 0.0731 (0.0017) & 0.0616 (0.0045) & 0.0334 (0.0035) & 0.0141 (0.0016) & 0.0785 (0.0104)\\
 & P-DLNM & 0.1188 (0.0038) & 0.0796 (0.0021) & 0.2111 (0.0116) & 0.0724 (0.0034) & 0.0317 (0.0010) & 0.1681 (0.0104)\\
 & T-DLNM & 0.0399 (0.0017) & 0.0235 (0.0007) & 0.0785 (0.0048) & 0.0706 (0.0030) & 0.0406 (0.0017) & 0.1414 (0.0075)\\
\addlinespace
DGP 4 & DLNAM & 0.1530 (0.0020) & 0.0995 (0.0013) & 0.2789 (0.0038) & 0.0356 (0.0010) & 0.0227 (0.0006) & 0.0661 (0.0020)\\
 & DLNM (QAIC) & 0.1137 (0.0016) & 0.0777 (0.0010) & 0.1983 (0.0033) & 0.1332 (0.0046) & 0.0837 (0.0031) & 0.2496 (0.0097)\\
 & DLNM (QBIC) & 0.2171 (0.0067) & 0.1452 (0.0042) & 0.3861 (0.0126) & 0.0971 (0.0065) & 0.0595 (0.0039) & 0.1854 (0.0134)\\
 & P-DLNM & 0.1832 (0.0032) & 0.1196 (0.0020) & 0.3327 (0.0061) & 0.0670 (0.0018) & 0.0467 (0.0014) & 0.1147 (0.0032)\\
 & T-DLNM & 0.0623 (0.0045) & 0.0513 (0.0031) & 0.0883 (0.0081) & 0.1325 (0.0045) & 0.0903 (0.0032) & 0.2318 (0.0081)\\
\addlinespace
\bottomrule
\end{tabular}
\caption{\textbf{Simulation Study: Bias and Variance (Surface).} Squared bias and variance of log-relative-risk by DGP, method, and exposure region, with Monte Carlo standard errors in parentheses; $R=200$.}
\label{tab:supp_mc_surf_bv}
\end{sidewaystable}

\begin{sidewaystable}[p]
\centering
\small
\setlength{\tabcolsep}{3pt}
\begin{tabular}{ll ccc ccc}
\toprule
 & & \multicolumn{3}{c}{Relative RMSE} & \multicolumn{3}{c}{Mean Interval Width}\\
\cmidrule(lr){3-5}\cmidrule(lr){6-8}
DGP & Model & Total & Interior & Boundary & Total & Interior & Boundary\\
\midrule
DGP 1 & DLNAM & 1.00 & 0.86 & 1.27 & 0.017 & 0.014 & 0.025\\
 & DLNM (QAIC) & 0.87 & 0.69 & 1.18 & 0.011 & 0.008 & 0.018\\
 & DLNM (QBIC) & 0.83 & 0.70 & 1.08 & 0.009 & 0.007 & 0.014\\
 & P-DLNM & 0.85 & 0.73 & 1.07 & 0.013 & 0.010 & 0.019\\
 & T-DLNM & 1.34 & 1.20 & 1.64 & 0.023 & 0.019 & 0.033\\
\addlinespace
DGP 2 & DLNAM & 1.00 & 0.75 & 1.42 & 0.020 & 0.016 & 0.031\\
 & DLNM (QAIC) & 1.70 & 1.27 & 2.43 & 0.028 & 0.021 & 0.042\\
 & DLNM (QBIC) & 2.28 & 1.70 & 3.26 & 0.020 & 0.016 & 0.031\\
 & P-DLNM & 1.31 & 1.05 & 1.78 & 0.026 & 0.021 & 0.038\\
 & T-DLNM & 1.47 & 1.12 & 2.06 & 0.033 & 0.025 & 0.052\\
\addlinespace
DGP 3 & DLNAM & 1.00 & 0.78 & 1.38 & 0.019 & 0.014 & 0.030\\
 & DLNM (QAIC) & 1.58 & 1.11 & 2.34 & 0.024 & 0.018 & 0.037\\
 & DLNM (QBIC) & 1.53 & 1.41 & 1.79 & 0.012 & 0.009 & 0.020\\
 & P-DLNM & 2.09 & 1.59 & 2.94 & 0.023 & 0.018 & 0.034\\
 & T-DLNM & 1.59 & 1.21 & 2.24 & 0.021 & 0.016 & 0.033\\
\addlinespace
DGP 4 & DLNAM & 1.00 & 0.80 & 1.35 & 0.022 & 0.017 & 0.032\\
 & DLNM (QAIC) & 1.14 & 0.93 & 1.54 & 0.030 & 0.024 & 0.045\\
 & DLNM (QBIC) & 1.29 & 1.04 & 1.74 & 0.021 & 0.017 & 0.032\\
 & P-DLNM & 1.15 & 0.94 & 1.54 & 0.029 & 0.024 & 0.042\\
 & T-DLNM & 1.02 & 0.87 & 1.30 & 0.035 & 0.027 & 0.053\\
\addlinespace
\bottomrule
\end{tabular}
\caption{\textbf{Simulation Study: Relative Error and Interval Width (Surface).} Relative RMSE and mean pointwise 95\% interval width by DGP, method, and exposure region; $R=200$. Relative RMSE uses the DLNAM total RMSE within each DGP as the common denominator. Interval width is on the log-relative-risk scale.}
\label{tab:supp_mc_surf_rel}
\end{sidewaystable}

\begin{sidewaystable}[p]
\centering
\small
\setlength{\tabcolsep}{3pt}
\begin{tabular}{ll ccc ccc}
\toprule
 & & \multicolumn{3}{c}{RMSE (MCSE)} & \multicolumn{3}{c}{Coverage (MCSE)}\\
\cmidrule(lr){3-5}\cmidrule(lr){6-8}
DGP & Encoder & Total & Interior & Boundary & Total & Interior & Boundary\\
\midrule
DGP 1 & Concatenation & 0.0063 (0.0002) & 0.0055 (0.0001) & 0.0079 (0.0002) & 0.888 (0.010) & 0.883 (0.011) & 0.901 (0.010)\\
 & Unified Shared Bias & 0.0111 (0.0003) & 0.0100 (0.0003) & 0.0133 (0.0004) & 0.949 (0.005) & 0.946 (0.006) & 0.958 (0.006)\\
 & Unified Local Bias & 0.0079 (0.0002) & 0.0070 (0.0002) & 0.0098 (0.0003) & 0.929 (0.005) & 0.920 (0.005) & 0.950 (0.006)\\
\addlinespace
DGP 2 & Concatenation & 0.0075 (0.0001) & 0.0056 (0.0001) & 0.0107 (0.0002) & 0.898 (0.006) & 0.912 (0.007) & 0.864 (0.008)\\
 & Unified Shared Bias & 0.0140 (0.0002) & 0.0124 (0.0002) & 0.0172 (0.0003) & 0.891 (0.007) & 0.890 (0.008) & 0.893 (0.007)\\
 & Unified Local Bias & 0.0100 (0.0002) & 0.0085 (0.0002) & 0.0129 (0.0003) & 0.900 (0.007) & 0.894 (0.009) & 0.912 (0.009)\\
\addlinespace
DGP 3 & Concatenation & 0.0066 (0.0001) & 0.0051 (0.0001) & 0.0091 (0.0003) & 0.926 (0.006) & 0.927 (0.007) & 0.921 (0.007)\\
 & Unified Shared Bias & 0.0157 (0.0004) & 0.0129 (0.0003) & 0.0209 (0.0005) & 0.938 (0.006) & 0.938 (0.007) & 0.940 (0.007)\\
 & Unified Local Bias & 0.0091 (0.0002) & 0.0066 (0.0001) & 0.0132 (0.0004) & 0.962 (0.004) & 0.966 (0.004) & 0.952 (0.006)\\
\addlinespace
DGP 4 & Concatenation & 0.0138 (0.0001) & 0.0111 (0.0001) & 0.0185 (0.0002) & 0.849 (0.007) & 0.865 (0.007) & 0.810 (0.009)\\
 & Unified Shared Bias & 0.0221 (0.0003) & 0.0195 (0.0003) & 0.0273 (0.0004) & 0.838 (0.007) & 0.838 (0.008) & 0.839 (0.008)\\
 & Unified Local Bias & 0.0159 (0.0004) & 0.0137 (0.0003) & 0.0203 (0.0005) & 0.872 (0.007) & 0.872 (0.008) & 0.873 (0.009)\\
\addlinespace
\bottomrule
\end{tabular}
\caption{\textbf{Simulation Study: ExU Comparison (Surface).} Log-relative-risk RMSE and pointwise 95\% interval coverage by DGP, method, and exposure region, with Monte Carlo standard errors in parentheses; $R=50$.}
\label{tab:supp_exu_surf}
\end{sidewaystable}

\begin{sidewaystable}[p]
\centering
\small
\setlength{\tabcolsep}{3pt}
\begin{tabular}{ll ccc ccc}
\toprule
 & & \multicolumn{3}{c}{RMSE (MCSE)} & \multicolumn{3}{c}{Coverage (MCSE)}\\
\cmidrule(lr){3-5}\cmidrule(lr){6-8}
DGP & Configuration & Total & Interior & Boundary & Total & Interior & Boundary\\
\midrule
DGP 1 & Reference & 0.0063 (0.0002) & 0.0055 (0.0001) & 0.0079 (0.0002) & 0.888 (0.010) & 0.883 (0.011) & 0.901 (0.010)\\
 & ExU Ablation & 0.0177 (0.0003) & 0.0172 (0.0003) & 0.0190 (0.0003) & 0.511 (0.017) & 0.511 (0.020) & 0.512 (0.016)\\
 & Mixture Ablation & 0.0095 (0.0002) & 0.0085 (0.0001) & 0.0115 (0.0002) & 0.884 (0.004) & 0.872 (0.005) & 0.910 (0.004)\\
 & Smoothness Ablation & 0.0110 (0.0002) & 0.0077 (0.0002) & 0.0163 (0.0004) & 0.955 (0.004) & 0.967 (0.003) & 0.927 (0.007)\\
\addlinespace
DGP 2 & Reference & 0.0075 (0.0001) & 0.0056 (0.0001) & 0.0107 (0.0002) & 0.898 (0.006) & 0.912 (0.007) & 0.864 (0.008)\\
 & ExU Ablation & 0.0321 (0.0001) & 0.0237 (0.0000) & 0.0463 (0.0002) & 0.206 (0.010) & 0.229 (0.010) & 0.153 (0.009)\\
 & Mixture Ablation & 0.0109 (0.0002) & 0.0084 (0.0001) & 0.0151 (0.0002) & 0.841 (0.007) & 0.852 (0.007) & 0.816 (0.009)\\
 & Smoothness Ablation & 0.0108 (0.0002) & 0.0079 (0.0002) & 0.0155 (0.0003) & 0.950 (0.003) & 0.963 (0.003) & 0.922 (0.005)\\
\addlinespace
DGP 3 & Reference & 0.0066 (0.0001) & 0.0051 (0.0001) & 0.0091 (0.0003) & 0.926 (0.006) & 0.927 (0.007) & 0.921 (0.007)\\
 & ExU Ablation & 0.0245 (0.0003) & 0.0192 (0.0002) & 0.0337 (0.0008) & 0.586 (0.017) & 0.619 (0.019) & 0.507 (0.019)\\
 & Mixture Ablation & 0.0130 (0.0003) & 0.0091 (0.0001) & 0.0193 (0.0006) & 0.839 (0.006) & 0.830 (0.007) & 0.860 (0.006)\\
 & Smoothness Ablation & 0.0113 (0.0002) & 0.0076 (0.0001) & 0.0170 (0.0004) & 0.964 (0.003) & 0.976 (0.002) & 0.935 (0.005)\\
\addlinespace
DGP 4 & Reference & 0.0138 (0.0001) & 0.0111 (0.0001) & 0.0185 (0.0002) & 0.849 (0.007) & 0.865 (0.007) & 0.810 (0.009)\\
 & ExU Ablation & 0.0372 (0.0002) & 0.0303 (0.0002) & 0.0496 (0.0004) & 0.365 (0.017) & 0.382 (0.019) & 0.322 (0.016)\\
 & Mixture Ablation & 0.0200 (0.0002) & 0.0173 (0.0002) & 0.0252 (0.0002) & 0.676 (0.009) & 0.661 (0.012) & 0.712 (0.006)\\
 & Smoothness Ablation & 0.0130 (0.0002) & 0.0102 (0.0002) & 0.0179 (0.0003) & 0.921 (0.004) & 0.935 (0.005) & 0.889 (0.006)\\
\addlinespace
\bottomrule
\end{tabular}
\caption{\textbf{Simulation Study: Architecture Ablation (Surface).} Log-relative-risk RMSE and pointwise 95\% interval coverage by DGP, method, and exposure region, with Monte Carlo standard errors in parentheses; $R=50$.}
\label{tab:supp_abl_surf}
\end{sidewaystable}

\begin{sidewaystable}[p]
\centering
\small
\setlength{\tabcolsep}{3pt}
\begin{tabular}{ll ccc ccc}
\toprule
 & & \multicolumn{3}{c}{RMSE (MCSE)} & \multicolumn{3}{c}{Coverage (MCSE)}\\
\cmidrule(lr){3-5}\cmidrule(lr){6-8}
Component & Estimator & Total & Interior & Boundary & Total & Interior & Boundary\\
\midrule
DGP 1 & DLNAM & 0.0167 (0.0009) & 0.0137 (0.0006) & 0.0215 (0.0017) & 0.854 (0.020) & 0.813 (0.021) & 0.938 (0.023)\\
 & DLNM (QAIC) & 0.0264 (0.0015) & 0.0182 (0.0009) & 0.0383 (0.0028) & 0.793 (0.023) & 0.748 (0.028) & 0.887 (0.024)\\
 & DLNM (QBIC) & 0.0320 (0.0021) & 0.0227 (0.0014) & 0.0457 (0.0038) & 0.639 (0.026) & 0.588 (0.030) & 0.746 (0.032)\\
 & P-DLNM & 0.0330 (0.0018) & 0.0265 (0.0012) & 0.0436 (0.0035) & 0.617 (0.023) & 0.525 (0.025) & 0.809 (0.029)\\
 & T-DLNM & 0.0512 (0.0029) & 0.0404 (0.0021) & 0.0684 (0.0052) & 0.607 (0.024) & 0.551 (0.024) & 0.725 (0.037)\\
\addlinespace
DGP 2 & DLNAM & 0.0173 (0.0007) & 0.0143 (0.0005) & 0.0223 (0.0014) & 0.890 (0.013) & 0.858 (0.016) & 0.956 (0.014)\\
 & DLNM (QAIC) & 0.0304 (0.0014) & 0.0211 (0.0011) & 0.0439 (0.0025) & 0.778 (0.019) & 0.760 (0.025) & 0.815 (0.021)\\
 & DLNM (QBIC) & 0.0405 (0.0015) & 0.0291 (0.0013) & 0.0575 (0.0028) & 0.582 (0.024) & 0.541 (0.029) & 0.668 (0.027)\\
 & P-DLNM & 0.0474 (0.0015) & 0.0380 (0.0013) & 0.0626 (0.0031) & 0.507 (0.021) & 0.431 (0.023) & 0.666 (0.028)\\
 & T-DLNM & 0.0596 (0.0029) & 0.0403 (0.0014) & 0.0871 (0.0059) & 0.629 (0.020) & 0.600 (0.023) & 0.690 (0.033)\\
\addlinespace
DGP 3 & DLNAM & 0.0241 (0.0014) & 0.0136 (0.0005) & 0.0382 (0.0029) & 0.888 (0.015) & 0.891 (0.018) & 0.880 (0.020)\\
 & DLNM (QAIC) & 0.0459 (0.0024) & 0.0288 (0.0021) & 0.0700 (0.0039) & 0.646 (0.025) & 0.644 (0.031) & 0.650 (0.024)\\
 & DLNM (QBIC) & 0.0471 (0.0025) & 0.0293 (0.0021) & 0.0721 (0.0042) & 0.639 (0.025) & 0.635 (0.029) & 0.648 (0.025)\\
 & P-DLNM & 0.0971 (0.0024) & 0.0693 (0.0019) & 0.1398 (0.0054) & 0.273 (0.015) & 0.213 (0.016) & 0.403 (0.020)\\
 & T-DLNM & 0.0716 (0.0027) & 0.0502 (0.0013) & 0.1040 (0.0056) & 0.483 (0.020) & 0.434 (0.022) & 0.591 (0.027)\\
\addlinespace
DGP 4 & DLNAM & 0.0224 (0.0008) & 0.0170 (0.0008) & 0.0308 (0.0016) & 0.847 (0.017) & 0.824 (0.021) & 0.894 (0.015)\\
 & DLNM (QAIC) & 0.0368 (0.0013) & 0.0297 (0.0008) & 0.0483 (0.0024) & 0.506 (0.018) & 0.439 (0.016) & 0.645 (0.028)\\
 & DLNM (QBIC) & 0.0377 (0.0013) & 0.0301 (0.0010) & 0.0500 (0.0022) & 0.494 (0.018) & 0.430 (0.016) & 0.626 (0.029)\\
 & P-DLNM & 0.0463 (0.0019) & 0.0394 (0.0017) & 0.0582 (0.0029) & 0.440 (0.017) & 0.361 (0.016) & 0.606 (0.027)\\
 & T-DLNM & 0.0722 (0.0026) & 0.0643 (0.0027) & 0.0864 (0.0040) & 0.428 (0.020) & 0.365 (0.023) & 0.560 (0.027)\\
\addlinespace
Null Exposure & DLNAM & 0.0124 (0.0008) & 0.0092 (0.0006) & 0.0177 (0.0014) & 0.955 (0.012) & 0.964 (0.013) & 0.935 (0.022)\\
 & DLNM (QAIC) & 0.0167 (0.0016) & 0.0114 (0.0011) & 0.0249 (0.0032) & 0.897 (0.023) & 0.879 (0.031) & 0.937 (0.019)\\
 & DLNM (QBIC) & 0.0123 (0.0008) & 0.0079 (0.0006) & 0.0188 (0.0012) & 0.928 (0.029) & 0.930 (0.031) & 0.924 (0.028)\\
 & P-DLNM & 0.0171 (0.0013) & 0.0119 (0.0010) & 0.0253 (0.0019) & 0.841 (0.042) & 0.850 (0.043) & 0.821 (0.042)\\
 & T-DLNM & 0.0603 (0.0044) & 0.0459 (0.0024) & 0.0845 (0.0089) & 0.549 (0.035) & 0.507 (0.036) & 0.644 (0.045)\\
\addlinespace
\bottomrule
\end{tabular}
\caption{\textbf{Simulation Study: Model Comparison (Joint).} Cumulative log-relative-risk RMSE and pointwise 95\% interval coverage by component, method, and exposure region, with Monte Carlo standard errors in parentheses; $R=50$.}
\label{tab:supp_joint}
\end{sidewaystable}

\end{document}